\documentclass[10pt]{article} % For LaTeX2e
\usepackage[preprint]{tmlr}

\makeatletter
\let\addcontentsline\@addcontentsline
\makeatother
\usepackage{titletoc}

\usepackage[hyphens]{url}  % DO NOT CHANGE THIS
\usepackage{graphicx} % DO NOT CHANGE THIS
\usepackage{natbib}  % DO NOT CHANGE THIS AND DO NOT ADD ANY OPTIONS TO IT
\usepackage{caption} % DO NOT CHANGE THIS AND DO NOT ADD ANY OPTIONS TO IT
\usepackage[hidelinks,colorlinks=true,linkcolor=blue,citecolor=blue]{hyperref}
\usepackage{amsmath}
\usepackage{amssymb}
\usepackage{amsthm}
\usepackage{mathtools}
\usepackage{bm}
\usepackage{booktabs}
\usepackage{multirow}
\usepackage{xcolor}
\usepackage{colortbl}
\usepackage{framed}
\usepackage{algorithm}
\usepackage{algpseudocode}
\usepackage{wrapfig}
\usepackage{pifont}
\usepackage{fontawesome5}
\usepackage{tikz}
\usepackage[most]{tcolorbox}
\usetikzlibrary{arrows.meta,positioning,fit,backgrounds,calc,shapes.geometric,decorations.pathreplacing}
\graphicspath{{figures/}{../}}   

\newcommand{\exbad}[1]{\textcolor{red!72!black}{#1}}
\newcommand{\exmask}{\textcolor{gray!40}{\rule[-0.05em]{0.75em}{0.7em}}}
\newcommand{\exmore}{\textcolor{gray}{\ldots}}
\tcbset{excard/.style={enhanced, boxrule=0.5pt, arc=1.2mm, left=3pt, right=3pt, top=2pt, bottom=3pt, toptitle=1pt, bottomtitle=1pt, fonttitle=\sffamily\scriptsize\bfseries, fontupper=\scriptsize\linespread{1.15}\selectfont\raggedright}}
\newtcolorbox{mdlmcard}[2]{excard, colback=gray!4, colframe=gray!55!black, colbacktitle=gray!55!black, coltitle=white, title={#1\hspace{0.8em}\hfill\mdseries #2}}
\newtcolorbox{srhcard}[2]{excard, colback=teal!4, colframe=teal!65!black, colbacktitle=teal!65!black, coltitle=white, title={#1\hspace{0.8em}\hfill\mdseries #2}}

{%
 \MakeFramed{\advance\hsize-\width\FrameRestore}%
 \noindent\small{\color{blue!45!black}\textbf{#1}}\par\nobreak\smallskip\noindent\small}%
{\endMakeFramed}

\newcommand{\method}{\textsc{Domino}}
\newcommand{\bx}{\boldsymbol{x}}
\newcommand{\bX}{\boldsymbol{X}}
\newcommand{\bz}{\boldsymbol{z}}

\newcommand{\xt}{\bz_t}

\newcommand{\Mask}{\texttt{[MASK]}}
\newcommand{\I}{\mathrm{I}}
\newcommand{\TC}{\mathrm{TC}}
\newcommand{\HH}{\mathrm{H}}
\newcommand{\pms}[1]{\ensuremath{_{\scriptscriptstyle\pm #1}}}
\newcommand{\E}{\mathbb{E}}
\newcommand{\KL}{\mathrm{KL}}
\newcommand{\Sset}{\mathcal{S}}

\newcommand{\Fam}{\mathcal{F}}

\DeclareMathOperator*{\softmax}{softmax}

\newtheorem{theorem}{Theorem}
\newtheorem{proposition}{Proposition}
\newtheorem{lemma}{Lemma}
\newtheorem{corollary}{Corollary}
\newtheorem{definition}{Definition}
\newtheoremstyle{uprightremark}{\topsep}{\topsep}{\normalfont}{}{\bfseries}{.}{.5em}{}
\theoremstyle{uprightremark}
\newtheorem{remark}{Remark}
\theoremstyle{plain}
\definecolor{shadecolor}{rgb}{0.91, 0.95, 0.93}
\let\origtheorem\theorem     \let\origendtheorem\endtheorem
\renewenvironment{theorem}{\begin{snugshade}\origtheorem}{\origendtheorem\end{snugshade}}
\let\origproposition\proposition \let\origendproposition\endproposition
\renewenvironment{proposition}{\begin{snugshade}\origproposition}{\origendproposition\end{snugshade}}
\let\origlemma\lemma         \let\origendlemma\endlemma
\renewenvironment{lemma}{\begin{snugshade}\origlemma}{\origendlemma\end{snugshade}}
\let\origcorollary\corollary \let\origendcorollary\endcorollary
\renewenvironment{corollary}{\begin{snugshade}\origcorollary}{\origendcorollary\end{snugshade}}

\definecolor{mlnavy}{HTML}{1D3557}
\definecolor{mlteal}{HTML}{2A9D8F}
\definecolor{mlpurple}{HTML}{7209B7}
\definecolor{mlamber}{HTML}{E9A03B}
\definecolor{mlgray}{HTML}{5F6368}
\definecolor{mlmask}{HTML}{ECEEF1}
\definecolor{mlmaskedge}{HTML}{C4C8CF}
\definecolor{mlgrad}{HTML}{C44536}
\definecolor{mlpanel}{HTML}{F5F6F8}
\newcommand{\algfwd}[1]{{\color{mlgray}#1}}
\newcommand{\algnet}[1]{{\color{mlnavy}#1}}
\newcommand{\alghead}[1]{{\color{mlteal!85!black}#1}}

\newcommand{\algobj}[1]{{\color{mlgrad}#1}}

\def\mlmaskkey{m}
\providecommand{\mltokrow}[3]{%
  \foreach \mlc/\mlw [count=\mlj from 0] in {#3} {%
    \ifx\mlc\mlmaskkey\node[msk] (#1-\mlj) at ($(#2)+(\mlj*0.88,0)$) {[M]};%
    \else\node[tok=\mlc] (#1-\mlj) at ($(#2)+(\mlj*0.88,0)$) {\mlw};\fi}%
}%

\title{Domino: Joint Unmasking via First-order Markov Chain for Few-Step Masked Diffusion}

\author{
\begin{center}
  Mansoor Ahmed~\textsuperscript{1,2\thanks{Address correspondence to: \quad \texttt{mahmed76@gsu.edu}, \quad \texttt{hvenkateswara@gsu.edu} }}  ,
  Lilia Chebbah~\textsuperscript{1},
  Jenny Dinh\textsuperscript{1},
  Murray Patterson \textsuperscript{1}, \\
  \textbf{Hemanth Venkateswara}\textsuperscript{1\tiny{*}} \hfill \\[8pt]
  \normalfont\normalsize
  \textsuperscript{1}\textit{Georgia State University, Atlanta, GA, USA} \\
  \textsuperscript{2}\textit{Georgia Institute of Technology, Atlanta, GA, USA} \\
  \end{center}
}

\def\month{MM}  % Insert correct month for camera-ready version
\def\year{YYYY} % Insert correct year for camera-ready version
\def\openreview{\url{https://openreview.net/forum?id=XXXX}} % Insert correct link to OpenReview for camera-ready version

\begin{document}

% \linenumbers 

\maketitle

\begin{abstract}
Masked diffusion models generate a sequence by revealing masked positions over many denoising steps. A few-step sampler has to reveal many positions at once, but the standard reverse kernel draws them \textit{independently} and primarily ignores the conditional dependence between tokens revealed in the same step. We propose a novel framework called \method{}, which models this dependence with a first-order Markov chain over the revealed positions, given the current state. The first revealed position keeps the posterior of a pre-trained masked diffusion model (MDM), and each subsequent position follows a transition given the token of its predecessor. Notably, the resulting step is normalized by construction, contains the factorized step in MDMs as a special case, and has a likelihood in closed form. We implement the \method{} transitions via a lightweight module on the hidden states of a pre-trained MDM denoiser with only 2.6$\%$ parameter overhead and train by keeping the MDM denoiser frozen. Our extensive evaluations across text, molecules, and images show that \method{} lowers the few-step generative perplexity of MDMs by a factor of 1.7 to 3.8 on One Billion Word (LM1B), WikiText-103, and OpenWebText at nearly unchanged sample entropy. It also raises the validity of few-step molecule samples by 1.5$\times$.
The source code is available at: \url{https://github.com/mansoor181/domino.git}
\end{abstract}

% =============================================================================
\section{Introduction}
% =============================================================================

\begin{wrapfigure}{r}{0.45\linewidth}
\vspace{-0.3in}
\centering
\includegraphics[width=\linewidth]{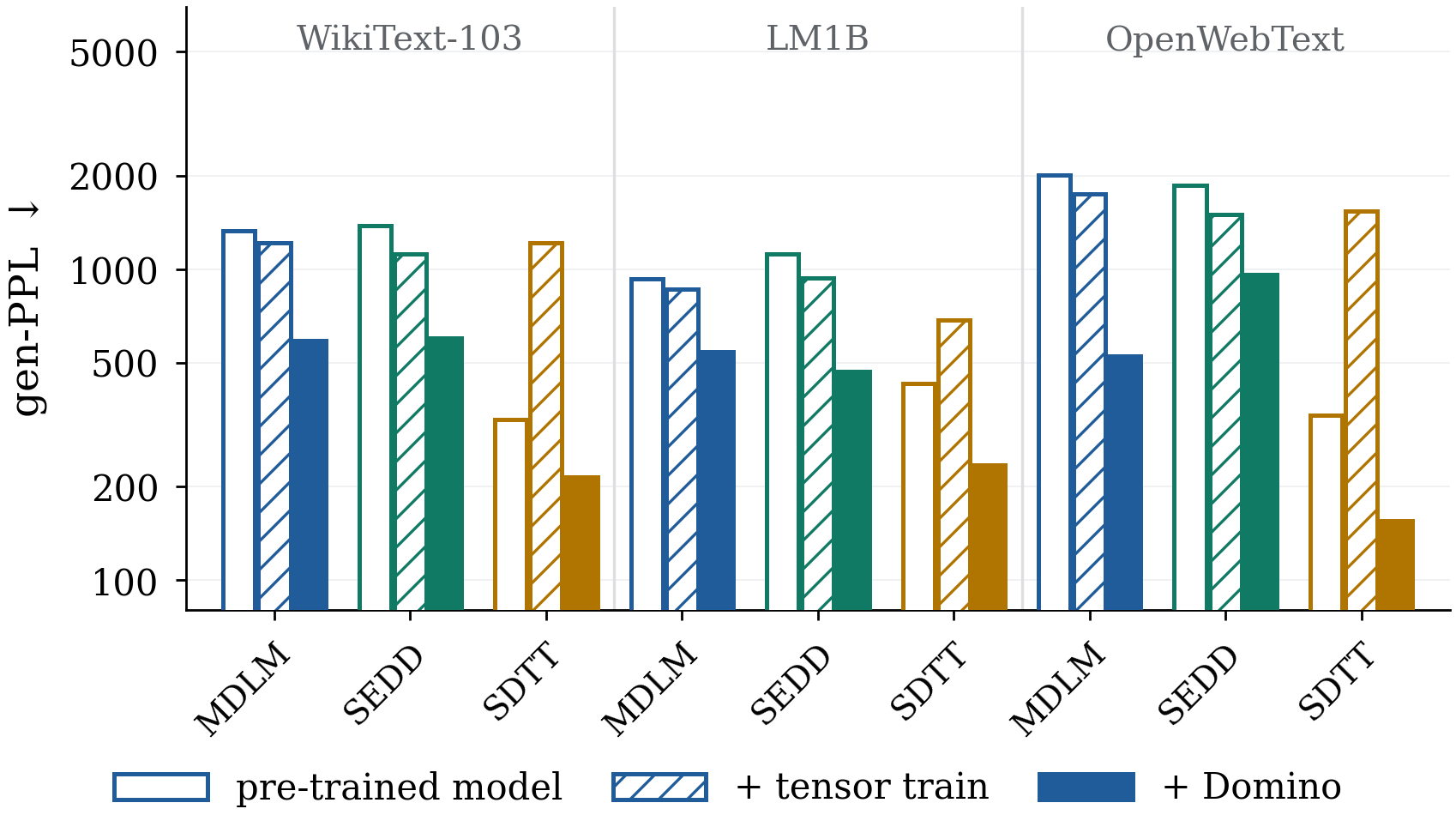}
\caption{\textbf{Few-step generation at four function evaluations.} \method{} achieves lower generative perplexity compared to the pre-trained masked diffusion models (MDMs) and TTD on WikiText-103, LM1B, and OWT datasets.}
\label{fig:intro}
\end{wrapfigure}

Discrete diffusion and flow models have emerged as a strong alternative to autoregressive models for generating sequences of discrete tokens, such as text and molecules \citep{austin2021d3pm,lou2023sedd,campbell2024dfm,gat2024dfm}. Their key promise is parallel generation, which simultaneously produces many tokens per network evaluation. Masked diffusion models (MDM) are the most common instance in which the forward process replaces tokens with an absorbing mask symbol \citep{sahoo2024mdlm,shi2024simplified,ou2024your}, and have recently scaled to large language models with competitive quality \citep{nie2025llada}.

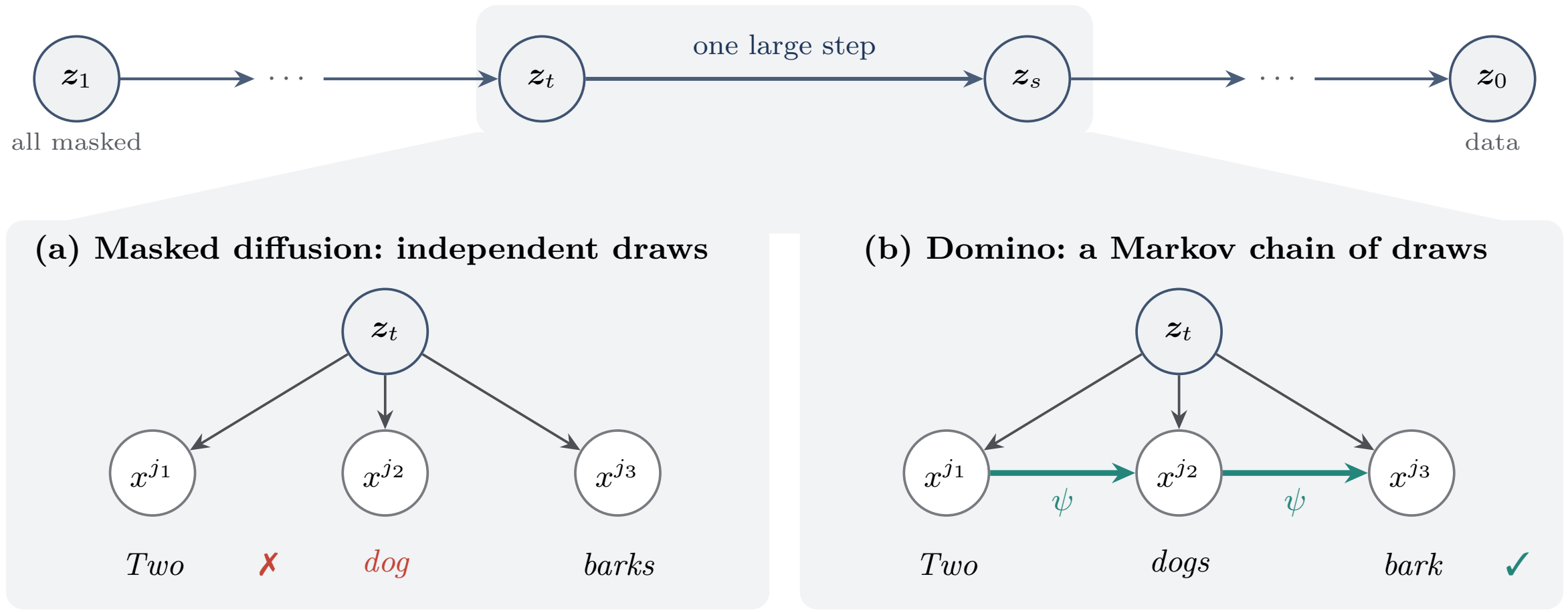
\begin{figure}[ht!]
\centering
\begin{tikzpicture}[
  x=1cm, y=1cm, font=\small,
  zn/.style={circle, draw=mlnavy!85, fill=mlnavy!7, line width=0.7pt, minimum size=8.4mm, inner sep=0pt, font=\small},
  xn/.style={circle, draw=mlgray!85, fill=white, line width=0.7pt, minimum size=8.4mm, inner sep=0pt, font=\small},
  dep/.style={-{Stealth[length=1.9mm, width=1.6mm]}, mlgray!80!black, line width=0.7pt},
  tstep/.style={-{Stealth[length=2mm, width=1.7mm]}, mlnavy!80, line width=0.8pt},
  chain/.style={-{Stealth[length=2.3mm, width=2mm]}, mlteal!85!black, line width=1.6pt},
  panel/.style={fill=mlnavy!6, rounded corners=5pt},
  ptitle/.style={anchor=west, font=\small\bfseries},
  word/.style={font=\small\itshape, inner sep=1pt},
  lab/.style={font=\scriptsize, text=mlgray!95!black},
]
% ================================================================== top: the denoising chain over steps
\node[zn] (z1) at (1.0,5.25) {$\bz_1$};
\node[font=\small, text=mlgray] (d1) at (3.1,5.25) {$\cdots$};
\node[zn] (zt) at (5.6,5.25) {$\bz_t$};
\node[zn] (zs) at (10.4,5.25) {$\bz_s$};
\node[font=\small, text=mlgray] (d2) at (12.9,5.25) {$\cdots$};
\node[zn] (z0) at (15.0,5.25) {$\bz_0$};
\draw[tstep] (z1) -- (d1); \draw[tstep] (d1) -- (zt); \draw[tstep] (zs) -- (d2); \draw[tstep] (d2) -- (z0);
\draw[tstep, line width=1.1pt] (zt) -- node[above=1.5pt, font=\footnotesize, text=mlnavy] {one large step} (zs);
\node[lab] at (1.0,4.63) {all masked};
\node[lab] at (15.0,4.63) {data};
% zoom from the step to the two panels below
\begin{scope}[on background layer]
\fill[mlnavy!6, rounded corners=6pt] (4.95,4.68) rectangle (11.05,5.98);
\fill[mlnavy!6] (4.95,4.72) -- (11.05,4.72) -- (15.7,3.72) -- (0.3,3.72) -- cycle;
\end{scope}
% ================================================================== (a) independent draws
\fill[panel] (0.3,0) rectangle (7.85,3.85);
\node[ptitle] at (0.45,3.55) {(a) Masked diffusion: independent draws};
\node[zn] (a0) at (4.05,2.75) {$\bz_t$};
\foreach \k/\x in {1/1.75, 2/4.05, 3/6.35} \node[xn] (a\k) at (\x,1.35) {$x^{j_\k}$};
\foreach \k in {1,2,3} \draw[dep] (a0) -- (a\k);
\node[word] at (1.75,0.45) {Two}; \node[word, text=mlgrad] at (4.05,0.45) {dog}; \node[word] at (6.35,0.45) {barks};
\node[text=mlgrad, font=\small] at (2.9,0.45) {\ding{55}};
% ================================================================== (b) Domino, a Markov chain over the revealed positions
\begin{scope}[xshift=7.85cm]
\fill[panel] (0.3,0) rectangle (7.85,3.85);
\node[ptitle] at (0.8,3.55) {(b) \method{}: a Markov chain of draws};
\node[zn] (b0) at (4.05,2.75) {$\bz_t$};
\foreach \k/\x in {1/1.75, 2/4.05, 3/6.35} \node[xn] (b\k) at (\x,1.35) {$x^{j_\k}$};
\foreach \k in {1,2,3} \draw[dep] (b0) -- (b\k);
\draw[chain] (b1) -- node[below, font=\small, text=mlteal!85!black] {$\psi$} (b2);
\draw[chain] (b2) -- node[below, font=\small, text=mlteal!85!black] {$\psi$} (b3);
\node[word] at (1.75,0.45) {Two}; \node[word] at (4.05,0.45) {dogs}; \node[word] at (6.35,0.45) {bark};
\node[text=mlteal!85!black, font=\small] at (7.4,0.45) {\ding{51}};
\end{scope}
\end{tikzpicture}
\caption{\textbf{A Markov chain within a denoising step.} One large step from $\bz_t$ to $\bz_s$ reveals three positions, which are present in the data as \emph{A dog barks} or \emph{Two dogs bark}. (a) Masked diffusion models (MDM) draw the tokens independently given $\bz_t$ and can mix the two sentences. (b) \method{} draws the tokens along a Markov chain through a pairwise potential $\psi$, given $\bz_t$ and its predecessor, and produces coherent sequences.}
\label{fig:concept}
\end{figure}

However, the sample quality degrades sharply in practice when a model generates sequences in only a few steps \citep{deschenaux2025sdtt,hayakawa2024di4c,sahoo2025duo}. This happens because of the form of the reverse step, where a few-step sampler must reveal many tokens in a denoising step. Given the current state $\bz_t$, an MDM draws the tokens of the masked positions $S$ from the posterior $P_i(\cdot \mid \bz_t)$ for each position $i$ \textit{independently},
$p(\bx^{S} \mid \bz_t, S) \;=\; \prod_{i \in S} P_i(x^{i} \mid \bz_t).$
We call this reverse kernel the \textit{product kernel}. However, the tokens revealed in the same step are rarely independent in the data. For instance, a subject and its verb should be coherent, and an opening bracket needs a closing bracket. The product kernel ignores such dependence and can combine tokens that do not fit together, as shown in Figure~\ref{fig:concept}(a). The smallest error is thus the total correlation (TC) of the step \citep{csiszar1975idivergence,watanabe1960information,zhang2026order}, which grows with the number of positions revealed at once. The error is therefore largest in the few-step regime, affecting faster parallel generation.

A joint distribution over the revealed tokens is therefore needed to remove this error. However, the full joint over $n$ positions has $V^{n}$ entries for a vocabulary of size $V$ and cannot be represented explicitly. To mitigate this, four broad lines of work have been proposed. They either couple the positions of a step through a shared latent variable, learned by distillation from a teacher \citep{hayakawa2024di4c} or under a variational bound \citep{zhang2025vmd}, or capture the dependence from a second model at inference, such as an autoregressive copula \citep{liu2024copula} or an energy model \citep{xu2025edlm}. Other methods reintroduce autoregressive generation across blocks \citep{arriola2025blockdiff} or across positions in a flexible order \citep{sahoo2025esolm,du2026a3,arriola2026setdiffusion}, at the cost of additional sequential passes. Unmasking approaches instead reveal together only positions that are close to independent, which limits the number of positions per step \citep{kim2025train,peng2025p2,luxembourg2025dus,benhamu2025eb,azangulov2025punt}. Closest to our work is tensor-train decomposition (TTD; \citealp{kim2026tensortrain}) that represents the joint of the masked positions as a low-rank tensor and fine-tunes a pre-trained model.

We build on two main observations. First, a pre-trained MDM already captures much of the dependence between the revealed tokens, since it predicts a token well once the related tokens are revealed. However, its product kernel cannot use this knowledge within a single step. Second, this dependence is largely local and of low order, and a few interactions between nearby positions are likely to recover most of it. To the best of our knowledge, the existing work has not explored a cheap way to pass the information between the tokens of one step inside the pre-trained MDM by exploiting its learned marginals.
This motivates our central question:
\begin{center}
\begin{minipage}{0.82\textwidth}
\begin{center}
\emph{Can we restore the dependence between the tokens revealed in one step of a pre-trained MDM, without retraining the model and at the sampling cost of a factorized step?}
\end{center}
\end{minipage}
\end{center}

In this work, we propose a novel framework called \method{} that draws the unmasked tokens in a step along a first-order Markov chain over their positions, given the current state (Figure~\ref{fig:concept}(b)). The first revealed position keeps the posterior of the pre-trained denoiser, and each subsequent position follows a transition given the token of its predecessor. A lightweight module computes these transitions from the hidden states of the pre-trained MDM denoiser. The product kernel is thus the special case in which the transitions are independent of the predecessor, and each step remains a normalized distribution with a likelihood in closed form. Our main contributions are as follows:

\begin{itemize}
    \item \textbf{Novel framework for joint unmasking in MDMs via a Markov chain.} We model the tokens revealed in a denoising step as a first-order Markov chain over their positions. The resulting step is normalized with a likelihood in closed form and the product kernel of MDMs as a special case.
    \item \textbf{A lightweight \method{} module on a frozen denoiser.} We compute the transitions of the Markov chain from the hidden states of a pre-trained MDM with a lightweight module. \method{} is trained without fine-tuning the denoiser, needs one denoiser pass per step with a parameter overhead of only about 2.6$\%$, and can be integrated into distilled and from-scratch MDMs.
    \item \textbf{Analysis of the few-step error.} We show that \method{} is trained on a bound on the likelihood of the few-step sampler, and that the chain removes the information shared by consecutive revealed positions from the error of each step. 
\end{itemize}

Our extensive experiments on text, molecules, and images show that \method{} significantly improves the few-step generation of pre-trained MDMs. At four steps, it lowers the generative perplexity of the masked diffusion language model (MDLM; \citealp{sahoo2024mdlm}) by a factor of 1.7 to 3.8 on three text corpora, with significant improvements compared to TTD (Figure~\ref{fig:intro}).

% =============================================================================
\section{Preliminaries and background}
\label{sec:prelim}
% =============================================================================

\subsection{Masked diffusion models (MDM)}
\label{sec:setup}

We follow the formulation of MDLM \citep{sahoo2024mdlm}. A clean sequence $\bx = (x^{1}, \dots, x^{L})$ of length $L$ has its tokens in a vocabulary $\Sset$ of size $V$, while corrupted sequences additionally have a mask token $\Mask \notin \Sset$. Superscripts index positions and subscripts index time. For a partially masked sequence $\bz$, $\mathcal{M}(\bz)$ is the set of its masked positions, and $\bx^{S}$ denotes the tokens of $\bx$ at a set $S$ of positions. The notations used throughout this paper are listed in App. Table~\ref{tab:notation}.

\textbf{Forward masking process.} For $t \in [0, 1]$, the forward process masks each token of $\bx$ independently,
\begin{equation}
q(\bz_t \mid \bx) \;=\; \prod_{i=1}^{L} \mathrm{Cat}\big(z_t^{i};\; \alpha_t\, \mathbf{e}_{x^{i}} + (1 - \alpha_t)\, \mathbf{e}_{\Mask}\big),
\label{eq:forward}
\end{equation}
where $\mathrm{Cat}(\cdot)$ is a categorical distribution over the vocabulary, $\mathbf{e}_{v}$ is the one-hot vector of token $v$, and the schedule $\alpha_t$ decreases from $\alpha_0 = 1$ to $\alpha_1 = 0$ \citep{sahoo2024mdlm,shi2024simplified}. Each token is kept with probability $\alpha_t$ and masked otherwise. The state $\bz_0$ is the clean sequence, and $\bz_1$ is fully masked.

\textbf{Reverse unmasking process.} For $s < t$, the posterior $q(\bz_s \mid \bz_t, \bx)$ of the forward process keeps each unmasked position of $\bz_t$ and unmasks each masked position independently with probability
\begin{equation}
r_{s,t} \;=\; \frac{\alpha_s - \alpha_t}{1 - \alpha_t},
\label{eq:rate}
\end{equation}
in which case the position takes its clean token. A masked diffusion model replaces the unknown clean token with the prediction of a denoiser. The denoiser maps $\bz_t$ to a hidden state $h_i \in \mathbb{R}^{D}$ at each position, and a linear output layer turns this state into logits $\ell_i \in \mathbb{R}^{V}$ and the posterior $P_i(\cdot \mid \bz_t) = \softmax(\ell_i)$ over the clean token at position $i$. The reverse step of the model factorizes over positions,
\begin{equation}
p_\theta(\bz_s \mid \bz_t) \;=\; \prod_{i=1}^{L} p_\theta(z_s^{i} \mid \bz_t),
\label{eq:reverse}
\end{equation}
where an unmasked position keeps its token, and a masked position stays masked with probability $1 - r_{s,t}$ and otherwise takes a token drawn from $P_i(\cdot \mid \bz_t)$ \citep{sahoo2024mdlm}.

\textbf{Objective and sampling.} The MDLM denoiser is trained on the negative evidence lower bound (NELBO) in continuous time,
\begin{equation}
\mathcal{L}_{\infty}(\bx) \;=\; \int_0^1 \frac{-\alpha_t'}{1 - \alpha_t}\; \E_{\bz_t \sim q(\cdot \mid \bx)}\Big[\sum_{i \in \mathcal{M}(\bz_t)} -\log P_i(x^{i} \mid \bz_t)\Big]\, dt,
\label{eq:nelbo}
\end{equation}
a cross-entropy at the masked positions, weighted over time, that bounds the negative log-likelihood of the model in the limit of infinitely many steps. Sampling starts from the fully masked sequence $\bz_1$ and applies the reverse step along a decreasing schedule $1 = t_0 > t_1 > \dots > t_K = 0$ with one network call per step. The number of steps $K$ is thus the number of function evaluations (NFEs).

\textbf{Reveal sets and content kernels.} A reverse step from $t$ to $s$ can be split into two draws. It first draws the set $S \subseteq \mathcal{M}(\bz_t)$ of positions to be unmasked (the \emph{reveal set}), and then draws their tokens. Since each masked position is unmasked independently with probability $r_{s,t}$, the reveal set follows the \emph{reveal distribution}
\begin{equation}
\pi_{s,t}(S \mid \bz_t) \;=\; r_{s,t}^{\,|S|}\, \big(1 - r_{s,t}\big)^{|\mathcal{M}(\bz_t)| - |S|},
\label{eq:reveal}
\end{equation}
which depends on $\bz_t$ only through its mask pattern. For example, with $\alpha_t = 1 - t$ and the equally spaced schedule $t_k = 1 - k/K$, the step from $t_k$ to $t_{k+1}$ unmasks each masked position with probability $1/(K - k)$ and reveals $L/K$ positions in expectation. We call the distribution of the revealed tokens given $\bz_t$ and $S$ the \emph{content kernel}. For the model of Eq.~\ref{eq:reverse}, it is the \emph{product kernel}
\begin{equation}
p(\bx^{S} \mid \bz_t, S) \;=\; \prod_{i \in S} P_i(x^{i} \mid \bz_t),
\label{eq:factorized}
\end{equation}
and we write $\Fam_{\mathrm{prod}}$ for the family of such kernels. MDLM and the absorbing-state variant of score entropy discrete diffusion (SEDD; \citealp{lou2023sedd}) are canonical models with this kernel.

\subsection{Factorization error}
\label{sec:ceiling}

To quantify the error of the product kernel, we fix a state $\bz_t$ and reveal set $S$, and let $q = q(\bx^{S} \mid \bz_t)$ be the data's distribution over the tokens revealed in a step, with marginal $q_i$ at position $i$. A product kernel can at best match these marginals, and its smallest error is the \emph{total correlation} (TC) of the transition \citep{watanabe1960information},
\begin{equation}
\TC(q) \;=\; \KL\Big(q \,\Big\|\, \prod_{i \in S} q_i\Big) \;=\; \sum_{i \in S} \HH(q_i) - \HH(q).
\label{eq:tc}
\end{equation}
Here $\KL$ is the Kullback-Leibler (KL) divergence and $\HH$ is the Shannon entropy. TC is zero if and only if the revealed tokens are independent given the state, and it thus measures the conditional dependence between them. In the toy example of Figure~\ref{fig:concept}, the data complete the three revealed positions as \emph{A dog barks} or \emph{Two dogs bark}, each with probability $\tfrac12$. Each position is then uniform over its two words, and $\TC(q) = 3 \log 2 - \log 2 = 2 \log 2$. This is the error of the product kernel in Figure~\ref{fig:concept}(a), which spreads its mass over all eight combinations. A product kernel cannot remove this error because the error is a property of the data's transition determined by the step size.

% =============================================================================
\section{\method{}}
\label{sec:method}
% =============================================================================

\begin{figure}[ht!]
\centering
\begin{tikzpicture}[
  x=1cm, y=1cm, font=\small,
  tok/.style={draw=mlgray!55, fill=white, rounded corners=2pt, minimum height=5.6mm, minimum width=12.5mm, inner sep=2pt, font=\small},
  old/.style={tok, text=mlgray},
  new/.style={tok, draw=mlteal!80!black, fill=mlteal!14, line width=0.6pt},
  msk/.style={draw=mlmaskedge, fill=mlmask, rounded corners=2pt, minimum height=5.6mm, minimum width=12.5mm, inner sep=2pt, text=mlgray, font=\scriptsize\ttfamily},
  rev/.style={msk, draw=mlteal!80!black, dashed, line width=0.6pt},
  stok/.style={tok, minimum width=9.4mm, minimum height=4.8mm, font=\footnotesize},
  smsk/.style={msk, minimum width=9.4mm, minimum height=4.8mm, font=\tiny\ttfamily},
  srev/.style={smsk, draw=mlteal!80!black, dashed, line width=0.5pt},
  panel/.style={fill=mlpanel, rounded corners=5pt},
  ptitle/.style={anchor=west, font=\small\bfseries, text=black},
  lab/.style={font=\scriptsize, text=mlgray!95!black, inner sep=1pt},
  flow/.style={-{Stealth[length=1.8mm]}, mlgray!80!black, semithick},
  tflow/.style={-{Stealth[length=1.8mm]}, mlteal!80!black, semithick},
  cond/.style={draw=mlteal!80!black, fill=white, rounded corners=2pt, minimum width=13mm, minimum height=7.5mm, align=center, font=\scriptsize, inner sep=1.5pt},
  gnode/.style={draw=mlteal!80!black, fill=mlteal!18, rounded corners=2pt, minimum width=7mm, minimum height=4mm, font=\scriptsize, inner sep=1pt},
]
% ================================================================ glyph helpers
% logits: four bars in a small frame; hidden state: four stacked squares; potential matrix: a 3x3 grid
\def\mllogits#1#2#3{%
  \begin{scope}[opacity=#3]
  \draw[mlnavy!50, fill=white, rounded corners=1pt] (#1-0.2,#2-0.17) rectangle (#1+0.2,#2+0.17);
  \foreach \k/\h in {0/0.08, 1/0.22, 2/0.12, 3/0.05} \fill[mlnavy!70] (#1-0.15+\k*0.085,#2-0.13) rectangle ++(0.055,\h);
  \end{scope}}
\def\mlhid#1#2#3{%
  \begin{scope}[opacity=#3]
  \foreach \k/\c in {0/35, 1/70, 2/20, 3/55} \draw[mlnavy!60, fill=mlnavy!\c] (#1-0.06,#2-0.2+\k*0.1) rectangle ++(0.12,0.1);
  \end{scope}}
\def\mlmat#1#2{%
  \foreach \r/\cc/\v in {0/0/60, 0/1/15, 0/2/35, 1/0/20, 1/1/70, 1/2/10, 2/0/40, 2/1/25, 2/2/55} \draw[mlteal!70!black, line width=0.2pt, fill=mlteal!\v] (#1-0.165+\cc*0.11,#2+0.055-\r*0.11) rectangle ++(0.11,0.11);}
% ================================================================ panel (a): one step
\fill[panel] (0,0) rectangle (10.0,6.62);
\node[ptitle] at (2.7,6.34) {(a) Sampling with \method{}};
\foreach \i/\x in {1/1.75, 2/3.55, 3/5.35, 4/7.15, 5/8.95} \coordinate (c\i) at (\x,0);
% input state and reveal set
\node[lab, anchor=west, font=\small] at (9.62,0.88) {$\bz_t$};
\node[old] (t1) at (c1 |- 0,0.88) {The};
\node[rev] (t2) at (c2 |- 0,0.88) {[M]};
\node[rev] (t3) at (c3 |- 0,0.88) {[M]};
\node[rev] (t4) at (c4 |- 0,0.88) {[M]};
\node[msk] (t5) at (c5 |- 0,0.88) {[M]};
\draw[decorate, decoration={brace, mirror, amplitude=3pt}, mlteal!80!black] (2.9,0.5) -- (7.8,0.5);
\node[lab, text=mlteal!80!black] at (5.35,0.22) {reveal set $S$ of the step, filled along a chain in index order};
% frozen trunk
\draw[mlnavy, fill=mlnavy!7, rounded corners=3pt, line width=0.6pt] (1.05,1.33) rectangle (9.6,1.93);
\node[font=\small, text=mlnavy] at (5.35,1.63) {\faSnowflake\ \ pre-trained MDM denoiser, one pass per step};
\foreach \i in {1,...,5} \draw[flow] (t\i.north) -- ++(0,0.17);
% per-position outputs of the trunk: logits and hidden state
\foreach \i/\x/\o in {1/1.75/0.3, 2/3.55/1, 3/5.35/1, 4/7.15/1, 5/8.95/0.3} {
  \mllogits{\x-0.25}{2.36}{\o}
  \mlhid{\x+0.25}{2.36}{\o}
}
\node[lab, anchor=east] at (1.25,2.36) {$\ell_i$};
\node[lab, anchor=west] at (2.25,2.36) {$h_i$};
% the head: pairs of hidden states and their gap produce the potential matrix of an edge
\draw[mlteal!80!black, fill=mlteal!6, rounded corners=3pt, line width=0.6pt] (1.05,2.74) rectangle (9.6,3.98);
\node[font=\small, text=mlteal!80!black, anchor=west] at (1.15,3.72) {\faFire\ \ module};
\node[lab, anchor=west, align=left] at (1.15,3.13) {pair of states\\$+$ gap $\to$ potential};
\foreach \a/\b/\m in {2/3/4.75, 3/4/6.55} {
  \node[gnode] (g\a) at (\m,3.03) {$g_\theta$};
  \draw[mlteal!80!black, line width=0.5pt] ($(c\a)+(0.25,2.58)$) .. controls ++(0,0.3) and ++(-0.4,0) .. (g\a.west);
  \draw[mlteal!80!black, line width=0.5pt] ($(c\b)+(0.25,2.58)$) .. controls ++(0,0.3) and ++(0.4,0) .. (g\a.east);
  \draw[mlteal!80!black, line width=0.5pt] (g\a.north) -- ++(0,0.2);
  \mlmat{\m}{3.62}
  \node[lab, anchor=east, text=mlteal!80!black] at (\m-0.2,3.62) {$A_{\a\b}$};
}
% conditionals along the chain
\node[cond] (p2) at (c2 |- 0,4.66) {{\color{mlgray}softmax}\\$\ell_2$};
\node[cond] (p3) at (c3 |- 0,4.66) {{\color{mlgray}softmax}\\$\ell_3+\psi_{23}$};
\node[cond] (p4) at (c4 |- 0,4.66) {{\color{mlgray}softmax}\\$\ell_4+\psi_{34}$};
\node[lab, anchor=north] at (p2.south) {};
\draw[flow] ($(c2)+(-0.25,2.55)$) -- ($(c2)+(-0.25,4.28)$);
\draw[flow] ($(c3)+(-0.25,2.55)$) -- ($(c3)+(-0.25,4.28)$);
\draw[flow] ($(c4)+(-0.25,2.55)$) -- ($(c4)+(-0.25,4.28)$);
\draw[tflow] (4.75,3.8) -- (4.75,4.28);
\draw[tflow] (6.55,3.8) -- (6.55,4.28);
\node[lab, text=mlteal!80!black, anchor=east] at (4.72,4.1) {$\psi_{23}$};
\node[lab, text=mlteal!80!black, anchor=east] at (6.52,4.1) {$\psi_{34}$};
% output state
\node[lab, anchor=west, font=\small] at (9.62,5.68) {$\bz_s$};
\node[old] (s1) at (c1 |- 0,5.68) {The};
\node[new] (s2) at (c2 |- 0,5.68) {two};
\node[new] (s3) at (c3 |- 0,5.68) {dogs};
\node[new] (s4) at (c4 |- 0,5.68) {bark};
\node[msk] (s5) at (c5 |- 0,5.68) {[M]};
\foreach \i in {2,3,4} \draw[flow] (p\i.north) -- (s\i.south);
% the sampled parent token selects the row of the potential used by the next revealed position
\draw[tflow, densely dashed] (s2.south east) .. controls ++(0.3,-0.28) and ++(-0.25,0.3) .. (p3.north west);
\draw[tflow, densely dashed] (s3.south east) .. controls ++(0.3,-0.28) and ++(-0.25,0.3) .. (p4.north west);
\draw[mlteal!80!black, densely dashed, semithick] (1.12,4.8) -- (1.52,4.8);
\node[lab, text=mlteal!80!black, anchor=west, align=left] at (1.55,4.68) {sampled\\predecessor};
% next step
\draw[flow, rounded corners=4pt] (s1.west) -- ++(-0.62,0) |- node[lab, pos=0.25, rotate=90, above, inner sep=2pt] {next state, $K$ steps in total} (t1.west);
% ================================================================ panel (b): training
\fill[panel] (10.35,0) rectangle (16,6.62);
\node[ptitle] at (10.5,6.34) {(b) Training \method{} module};
\node[lab, anchor=west] at (12,5.98) {clean sequence $\bx$};
\foreach \w [count=\k from 0] in {The, two, dogs, bark, loudly} \node[stok] (x\k) at (10.97+\k*1.04,5.58) {\w};
\node[draw=mlgray!70, fill=white, rounded corners=2pt, align=center, font=\scriptsize, inner sep=3pt] (fwd) at (13.05,4.72) {forward process: mask each token,\\then draw the reveal set $S$ of a step};
\draw[flow] (13.05,5.32) -- (fwd.north);
\draw[flow] (fwd.south) -- ++(0,-0.2);
\foreach \k in {1,2,3} \node[srev] at (10.97+\k*1.04,3.72) {[M]};
\node[stok, text=mlgray] at (10.97,3.72) {The};
\node[smsk] at (15.13,3.72) {[M]};
\node[lab, anchor=north] at (13.05,3.42) {$\bz_t$ and $S$ enter the networks of (a)};
\draw[flow] (13.05,3.0) -- (13.05,2.72);
\node[draw=mlgrad, fill=mlgrad!7, rounded corners=3pt, align=center, font=\scriptsize, inner sep=3pt] (loss) at (13.05,2.2) {chain likelihood at the true tokens\\[2pt]$\mathcal{L}=-\sum_{m\ge 2}\log P^{\theta}_{j_m\mid j_{m-1}}\big(x^{j_m}\mid x^{j_{m-1}},\bz_t\big)$};
\draw[-{Stealth[length=1.8mm]}, mlgrad, densely dashed, semithick, rounded corners=4pt] (loss.west) -- (9.87,2.2) |- (9.62,3.36);
\node[font=\scriptsize, text=mlgrad, anchor=north, align=center] at (13.05,1.64) {the predecessor is the true token,\\and only the module gets a gradient};
\node[lab, align=left, anchor=west] at (10.6,0.55) {\faSnowflake\ \ frozen: denoiser and logits\\[1pt]\faFire\ \ trained: $W_u$, $E$, gap table, $g_\theta$};
\end{tikzpicture}
\caption{\textbf{Sampling (a) and training (b) with \method{}.} (a) The module maps the hidden states of consecutive revealed positions to a pairwise potential $\psi_{ij}$, and the position after the first is drawn given its sampled predecessor. (b) We only train the \method{} module on the likelihood of the chain at the true tokens.}
\label{fig:method}
\end{figure}
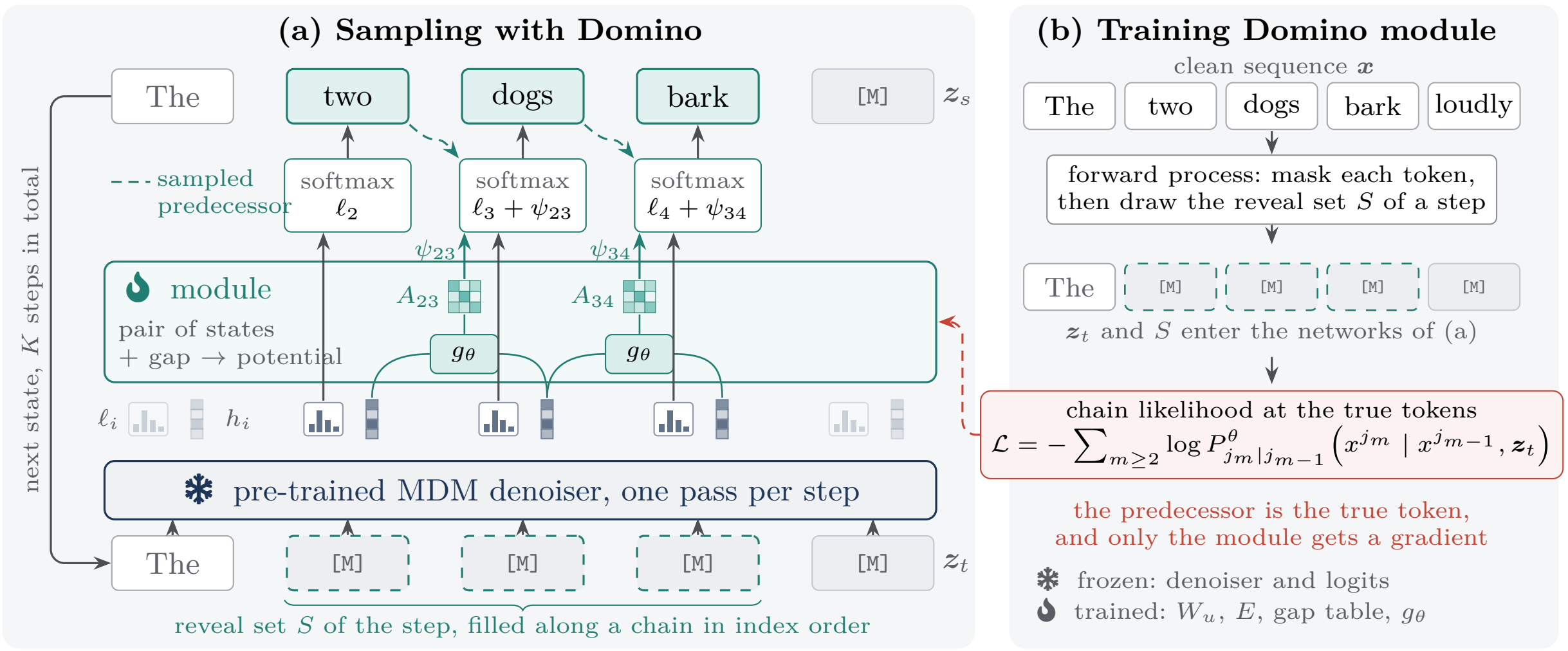

In this section, we describe the formulation and implementation of \method{}.
We first re-formulate the product kernel with a first-order Markov chain over the revealed positions to capture pairwise relationships in the data (Section~\ref{sec:kernel}). We then implement \method{} as a lightweight module integrated with pre-trained MDMs in Section~\ref{sec:potentials} and train with the objective in Section~\ref{sec:objective}. Figure~\ref{fig:method} illustrates the sampling and training.

\subsection{Markov chain over the revealed positions}
\label{sec:kernel}

Let $S \subseteq \mathcal{M}(\xt)$ be the reveal set of a step, drawn from the reveal distribution of Eq.~\ref{eq:reveal}, and let $j_1 < \dots < j_n$ be its positions in index order. \method{} draws the tokens of $S$ along this \emph{chain}, the first position from the denoiser's posterior and each further one from a \emph{transition} given the token of its predecessor. The resulting content kernel is
\begin{equation}
p^{\theta}(\bx^{S} \mid \xt, S) \;=\; P_{j_1}(x^{j_1} \mid \xt) \prod_{m=2}^{n} P^{\theta}_{j_m \mid j_{m-1}}\big(x^{j_m} \,\big|\, x^{j_{m-1}}, \xt\big),
\label{eq:kernel}
\end{equation}
where each transition $P^{\theta}_{j \mid i}(\cdot \mid a, \xt)$ is a categorical distribution over the vocabulary, one for each token $a$ at the predecessor $i$. Eq.~\ref{eq:kernel} is a first-order Markov chain over the revealed positions, conditioned on the state. Given $\xt$ and the token at $j_{m-1}$, the token at $j_m$ is independent of the tokens revealed before it, while all of them can depend on the visible tokens of $\xt$. The transitions vary with the positions and with the state, and the chain is therefore inhomogeneous. We write $C(S) = \{(j_{m-1}, j_m) : 2 \le m \le n\}$ for the consecutive pairs of the chain. As a product of conditionals, the kernel sums to one over $\Sset^{S}$, and its logarithm is a sum of $n$ terms. A chain whose transitions ignore the predecessor is the product kernel, and the Markov property is thus a weaker assumption than the conditional independence of the product kernel. The kernel reduces to the pre-trained model when a step reveals only one position.

\subsection{Parameterization of the transitions}
\label{sec:potentials}

One of our main contributions is to parameterize the transitions with a well-engineered lightweight module on the hidden states of the MDM denoiser. The module projects the hidden state $h_i$ at position $i$ to a \emph{pair factor} $u_i = W_u h_i \in \mathbb{R}^{d_p}$. For a consecutive pair $(i, j) \in C(S)$, the transition into position $j$ is
\begin{equation}
P^{\theta}_{j \mid i}(b \mid a, \xt) \;=\; \softmax_{b}\!\big(\ell_j(b) + \psi_{ij}(a, b)\big),
\qquad
\psi_{ij}(a, b) \;=\; E_{a}^{\top} A_{ij} E_{b},
\qquad
A_{ij} \;=\; g_\theta\big(u_i, u_j, \mathrm{emb}(j - i)\big),
\label{eq:potential}
\end{equation}
where $E \in \mathbb{R}^{V \times d_p}$ is a trained projection of the pre-trained input embedding, $g_\theta$ is a small network that returns a $d_p \times d_p$ matrix, and $\mathrm{emb}(j - i)$ embeds the signed gap between the two positions. Intuitively, $\psi_{ij}(a, b)$ shifts the logit of token $b$ at position $j$ by how well it combines with token $a$ at position $i$ in the current context, and the softmax renormalizes the result. As a $V \times V$ matrix, the potential has rank at most $d_p$, which sets the capacity of the module. We initialize the output layer of $g_\theta$ at zero. Each transition then starts at the denoiser's marginal, and \method{} initially coincides with the pre-trained model.

\subsection{Training objective}
\label{sec:objective}

We train the \method{} module by maximum likelihood on the transitions of the data over one step of a $K$-step schedule, while the denoiser stays frozen. Given a clean sequence $\bx$ and a step from $t$ to $s < t$, we draw $\xt$ from the forward process of Eq.~\ref{eq:forward} and the reveal set $S$ from Eq.~\ref{eq:reveal}. The loss is the negative log-likelihood of the true tokens of $S$ under the chain, with each transition evaluated at the true token of its predecessor,
\begin{equation}
\mathcal{L}(\theta) \;=\; -\E_{(t,s) \sim \nu}\, \E_{\bx, \xt, S}\Big[\sum_{m=2}^{n} \log P^{\theta}_{j_m \mid j_{m-1}}\big(x^{j_m} \,\big|\, x^{j_{m-1}}, \xt\big)\Big],
\label{eq:loss}
\end{equation}
where $\nu$ draws a step of the schedule with probability $1 - \eta$ and a step of random size with probability $\eta$. The random steps expose the module to the step sizes of other schedules. The first factor of Eq.~\ref{eq:kernel} and the reveal distribution do not depend on $\theta$ and drop out of the loss. The loss is computed by taking one forward pass of the denoiser and a softmax per revealed position, and updates only the \method{} module. On the steps of a $K$-step schedule, the loss thus equals the $K$-step bound of Proposition~\ref{prop:kbound} divided by $K$, up to a term that does not depend on $\theta$, and training minimizes a bound on the likelihood of the few-step sampler. The steps of random size act as an augmentation over other schedules and leave the optimal transitions unchanged (Remark~\ref{rem:objective}). Algorithms~\ref{alg:train} and~\ref{alg:sample} present the pseudocode for \method{} training and sampling.

\begin{figure}[t!]
\centering
\begin{minipage}[t]{0.505\textwidth}
\begin{algorithm}[H]
\footnotesize
\caption{\method{} training.}
\label{alg:train}
\begin{algorithmic}[1]
\Require denoiser $\phi$; module $\theta = (W_u, E, \mathrm{emb}, g_\theta)$; steps $K$; rate $\eta$
\Repeat
  \State $\bx \sim \mathrm{data}$
  \State \algfwd{$(t,s) \gets (t_k, t_{k+1})$ with $k \sim \mathcal{U}\{0,\dots,K-1\}$ w.p. $1-\eta$,}
  \Statex \algfwd{\hspace{1.05em} a step of random size w.p. $\eta$}
  \State \algfwd{$\xt \gets$ mask each $x^{i}$ independently w.p. $1-\alpha_t$}
  \State \algfwd{$S \gets$ each $i \in \mathcal{M}(\xt)$ independently w.p. $r_{s,t}$}
  \State \algnet{$(\ell, h) \gets \phi(\xt, t)$} \Comment{one pass, no gradient}
  \State \alghead{$u_i \gets W_u h_i$ for $i \in S$} \Comment{pair factors}
  \State $(j_1, \dots, j_n) \gets$ positions of $S$ in index order
  \State \algobj{$\mathcal{L} \gets 0$}
  \For{$m = 2, \dots, n$}
    \State \alghead{$A \gets g_\theta\big(u_{j_{m-1}}, u_{j_m}, \mathrm{emb}(j_m - j_{m-1})\big)$}
    \State \alghead{$\psi \gets E_{x^{j_{m-1}}}^{\top} A\, E$} \Comment{true predecessor}
    \State \algobj{$\mathcal{L} \gets \mathcal{L} - \log \softmax(\ell_{j_m} + \psi)[x^{j_m}]$}
  \EndFor
  \State \algobj{$\theta \gets \mathrm{OptimizerStep}(\theta, \mathcal{L})$} \Comment{module only}
\Until{convergence}
\end{algorithmic}
\end{algorithm}
\end{minipage}\hfill
\begin{minipage}[t]{0.475\textwidth}
\begin{algorithm}[H]
\footnotesize
\caption{\method{} sampling.}
\label{alg:sample}
\begin{algorithmic}[1]
\Require denoiser $\phi$; module $\theta$; steps $K$; length $L$
\State $\bz \gets [\Mask]^{L}$
\For{$k = 0, \dots, K-1$}
  \State $t \gets 1 - k/K$;\; $s \gets 1 - (k+1)/K$
  \State \algfwd{$S \gets$ each $i \in \mathcal{M}(\bz)$ independently w.p. $r_{s,t}$}
  \State \algnet{$(\ell, h) \gets \phi(\bz, t)$} \Comment{one pass}
  \State \alghead{$u_i \gets W_u h_i$ for $i \in S$}
  \State $(j_1, \dots, j_n) \gets$ positions of $S$ in index order
  \State $z^{j_1} \sim \softmax(\ell_{j_1})$ \Comment{denoiser marginal}
  \For{$m = 2, \dots, n$}
    \State \alghead{$A \gets g_\theta\big(u_{j_{m-1}}, u_{j_m}, \mathrm{emb}(j_m - j_{m-1})\big)$}
    \State \alghead{$\psi \gets E_{z^{j_{m-1}}}^{\top} A\, E$} \Comment{sampled predecessor}
    \State $z^{j_m} \sim \softmax(\ell_{j_m} + \psi)$ \Comment{Eq.~\ref{eq:potential}}
  \EndFor
\EndFor
\State \Return $\bz$
\end{algorithmic}
\end{algorithm}
\end{minipage}
\end{figure}

\section{Theoretical analysis}
\label{sec:theory}
 
In this section, we analyze the few-step error of \method{}. We first bound the likelihood of a $K$-step sampler and show that \method{} is trained on this bound (Section~\ref{sec:kbound}). We then derive the error of the Markov chain in a denoising step and its gain over the product kernel (Section~\ref{sec:steperror}). The proofs are provided in App.~\ref{app:proofs}.
 
\subsection{Likelihood bound of a few-step sampler}
\label{sec:kbound}
 
A $K$-step sampler starts from the fully masked sequence and, at each step $k = 0, \dots, K-1$, draws a reveal set $S_k$ from Eq.~\ref{eq:reveal} and then the revealed tokens from a content kernel $p^{\theta}(\cdot \mid \bz_{t_k}, S_k)$. We write $p_K$ for the distribution of its samples and $q$ for the data distribution. The continuous-time bound of Eq.~\ref{eq:nelbo} describes the limit $K \to \infty$. To bound $p_K$, we couple the forward process across times by masking each position $i$ of a clean sequence $\bx$ at an independent random time $U_i$ with $\Pr(U_i \le t) = 1 - \alpha_t$, so that $\bz_t$ follows Eq.~\ref{eq:forward} at each $t$. Let $S_k$ contain the positions with $t_{k+1} < U_i \le t_k$, which are masked at $t_k$ and visible at $t_{k+1}$, and let $\bz_{t_k}$ be the state that shows the tokens revealed before step $k$.
 
\begin{proposition}[$K$-step bound]
\label{prop:kbound}
For each content kernel $p^{\theta}$ and sequence $\bx$,
\begin{equation}
-\log p_K(\bx) \;\le\; \mathcal{L}_K(\theta; \bx) \;=\; \E\Big[\sum_{k=0}^{K-1} -\log p^{\theta}\big(\bx^{S_k} \,\big|\, \bz_{t_k}, S_k\big)\Big],
\label{eq:kbound}
\end{equation}
where the expectation is over the reveal sets of the forward process. Moreover, in expectation over the data,
\begin{equation}
\E_{\bx \sim q}\big[\mathcal{L}_K(\theta; \bx)\big] \;=\; \HH(\bX) + \sum_{k=0}^{K-1} \E\Big[\KL\big(q_k \,\big\|\, p^{\theta}(\cdot \mid \bz_{t_k}, S_k)\big)\Big],
\label{eq:kdecomp}
\end{equation}
where $q_k = q(\bx^{S_k} \mid \bz_{t_k})$ is the data's distribution over the reveal set of step $k$.
\end{proposition}
 
The bound $\mathcal{L}_K$ scores the true tokens of each step under the content kernel, and Eq.~\ref{eq:kdecomp} shows that the error of a few-step sampler accumulates over its steps. In particular, $\KL(q \,\|\, p_K)$ is at most the sum of the per-step divergences in Eq.~\ref{eq:kdecomp}. The likelihood $p_K(\bx)$ itself sums over the possible sequences of reveal sets and is intractable in general, while $\mathcal{L}_K$ is an expectation that we can estimate by sampling reveal sets. For the product kernel, $\mathcal{L}_K$ is the discrete-time bound of masked diffusion \citep{sahoo2024mdlm}, and each divergence is at least TC of the step (Lemma~\ref{lem:product}). This excess vanishes as the steps shrink, at a rate of order $1/K$ (Remark~\ref{rem:ctelbo}), in line with the analysis of \citet{zhao2026adaptation}, and the form of the kernel therefore matters when the number of steps is small. For \method{}, the first factor of Eq.~\ref{eq:kernel} does not depend on $\theta$. Hence, the loss of Eq.~\ref{eq:loss} on the steps of a schedule equals $\mathcal{L}_K / K$ up to a constant, and training \method{} minimizes a bound on the likelihood of the few-step sampler (Remark~\ref{rem:objective}).
 
\subsection{Error and gain of the Markov chain}
\label{sec:steperror}
 
We now quantify the amount of per-step error in Eq.~\ref{eq:kdecomp} removed by the chain formulation. As in Section~\ref{sec:ceiling}, we fix a state $\bz_t$ and a reveal set $S$, let $q = q(\bx^{S} \mid \bz_t)$, and write $\I_q(X^{i}; X^{j})$ for the mutual information between two revealed positions under $q$.
 
\begin{proposition}[Error of the chain]
\label{prop:chowliu}
The kernel $p^{\theta}$ of Eq.~\ref{eq:kernel} has the error
\begin{equation}
\KL\big(q \,\big\|\, p^{\theta}\big) \;=\; \rho(q) + \KL\big(q_{j_1} \,\big\|\, P_{j_1}\big) + \varepsilon_\theta(q), \qquad \rho(q) \;=\; \TC(q) - \sum_{(i,j) \in C(S)} \I_q(X^{i}; X^{j}),
\label{eq:steperror}
\end{equation}
where $\varepsilon_\theta(q) \ge 0$ is the expected divergence between the data's pairwise conditionals $q(x^{j_m} \mid x^{j_{m-1}})$ and the transitions of Eq.~\ref{eq:potential}, which vanishes if and only if the two coincide. Consequently, the gain of \method{} over the pre-trained product kernel is
\begin{equation}
\KL\Big(q \,\Big\|\, \prod_{i \in S} P_i\Big) - \KL\big(q \,\big\|\, p^{\theta}\big) \;=\; \sum_{(i,j) \in C(S)} \I_q(X^{i}; X^{j}) \;+\; \sum_{i \in S \setminus \{j_1\}} \KL\big(q_i \,\big\|\, P_i\big) \;-\; \varepsilon_\theta(q).
\label{eq:gain}
\end{equation}
\end{proposition}
 
Eq.~\ref{eq:steperror} separates three sources of error. The first term $\rho(q)$ follows from the dependence trees of \citet{chow1968approximating} and is the error of the best chain kernel. The chain removes the information that consecutive revealed positions share from TC, and $\rho(q)$ vanishes if and only if $q$ is itself a first-order Markov chain over $S$ in index order. In the example of Figure~\ref{fig:concept}, each of the two consecutive pairs shares $\log 2$ nats, and the chain removes the whole total correlation of $2 \log 2$. More generally, $\rho(q)$ vanishes at each state and reveal set when the data form a first-order Markov chain along the positions, since neither fixing the visible tokens nor summing out the masked positions outside $S$ breaks the Markov property. The second term is the error of the frozen denoiser at the first position, and the third term is the only one that depends on $\theta$ and is what the loss of Eq.~\ref{eq:loss} minimizes.
 
Eq.~\ref{eq:gain} shows that the gain of \method{} has two sources. The first is the information between consecutive revealed positions, which is the dependence ignored by the product kernel. The second is a correction of the denoiser's marginals at the positions after the first position, since each transition is fitted to the data. Both sources are attained with unrestricted transitions. The module shares its parameters across states and pairs and uses potentials of rank at most $d_p$, so its transition error $\varepsilon_\theta(q)$ is positive in general and lowers the gain. Intuitively, the optimal transition shifts the logits of the denoiser by the pointwise mutual information of the two tokens given the state, which \method{} approximates with a potential of rank at most $d_p$ (Remark~\ref{rem:pmi}).

\section{Experiments}
\label{sec:experiments}
% =============================================================================

\begin{figure}[t!]
\centering
\includegraphics[width=\textwidth]{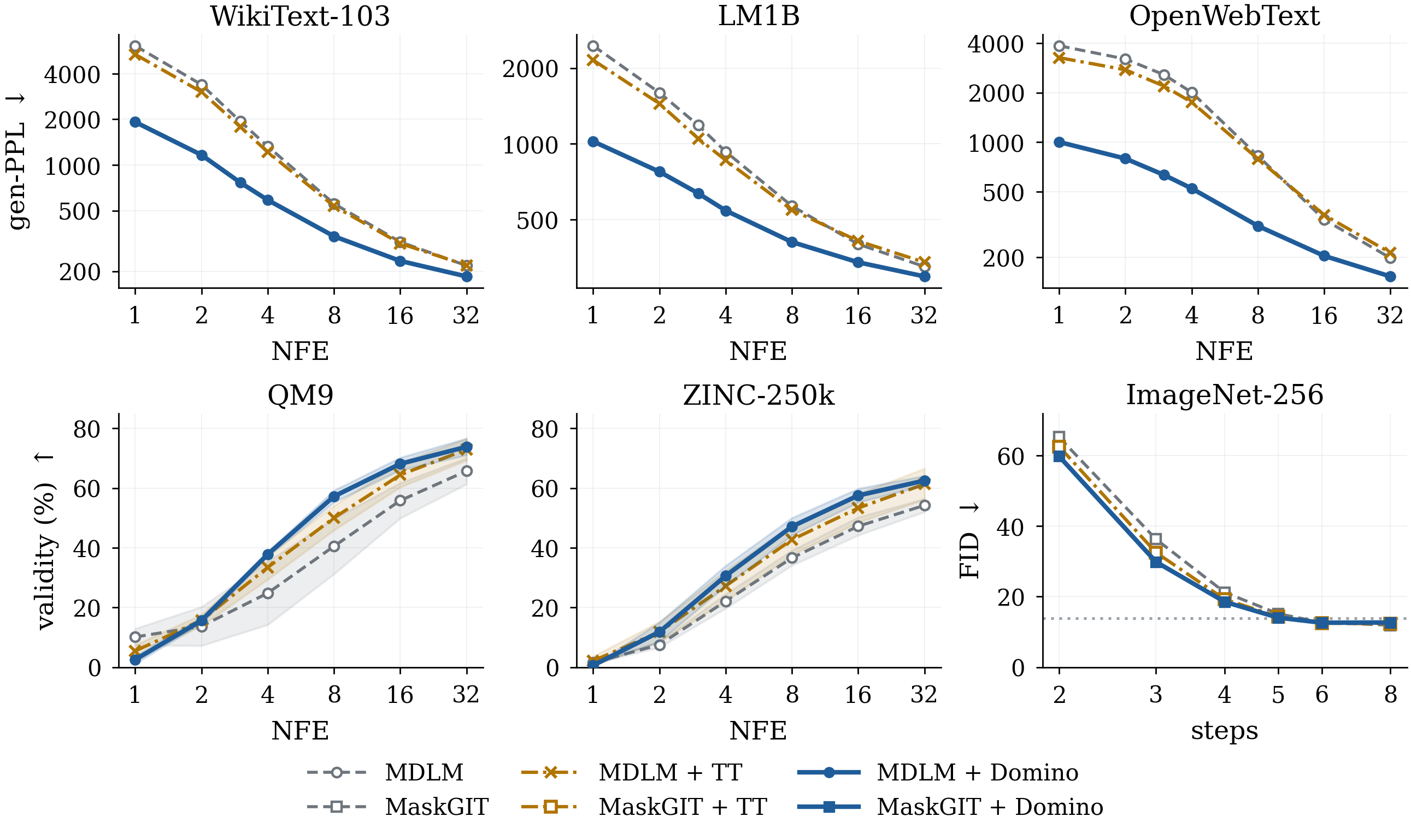}
\caption{\textbf{Few-step generation on six datasets.} Generative perplexity ($\downarrow$) on text, validity ($\uparrow$) on molecules, and Fr\'echet inception distance (FID, $\downarrow$) on ImageNet-256 across NFEs for MDLM (MaskGIT on images), with TTD and with \method{}. }
\label{fig:main}
\end{figure}

\subsection{Setup}
\label{sec:protocol}

\textbf{Dataset.} We integrate \method{} with different MDMs and evaluate on the WikiText-103 \citep{merity2016pointer} and One Billion Word benchmark (LM1B; \citealp{chelba2013lm1b}) at length $L = 128$, and OpenWebText (OWT; \citep{gokaslan2019owt}) at length 1,024. For the molecule datasets QM9 \citep{ramakrishnan2014qm9} and ZINC-250k \citep{irwin2012zinc}, we follow the training and evaluation setup in \citet{park2025pairflow} to generate strings in the simplified molecular-input line-entry system (SMILES). For images, we use the official checkpoint of the masked generative image transformer (MaskGIT; \citealp{chang2022maskgit}) for ImageNet at $256 \times 256$. Further experimental details on the data, training, and evaluation are provided in App.~\ref{app:protocol}.

\textbf{Model training.} Our primary baseline is MDLM as the product kernel. We also compare with absorbing-state SEDD and few-step SDTT students distilled from MDLM. The closest method to ours is TTD of \citet{kim2026tensortrain}, which represents the joint distribution of the masked positions as a product of low-rank tensors. We train it on the same frozen MDMs and number of steps at rank 2, which roughly matches the parameter size of \method{} for fair comparison\footnote{The authors in tensor-train (TT; \citealp{kim2026tensortrain}) originally use rank 8, which has a parameter overhead of roughly 39M.}. Finally, we compare with DUO \citep{sahoo2025duo}, a uniform-state diffusion model whose forward process replaces tokens with random tokens instead of a mask. We train DUO from scratch on WikiText-103 and LM1B, while we use its official checkpoint and the distilled student on OpenWebText. \method{} does not apply to such uniform-state models, since they can change any token at any step and do not reveal a set of positions.

\textbf{Metrics.} We report the generative perplexity (gen-PPL) under GPT-2 Large \citep{radford2019gpt2} for text evaluation. Since samplers may lower the perplexity by narrowing the outputs, we also report the average entropy of the per-position marginal. We also report MAUVE \citep{pillutla2021mauve}, which measures the closeness of the distribution of generated text to that of held-out text. For molecules, we report the percentage of valid samples, uniqueness and novelty of the valid samples, and Fr\'echet ChemNet distance (FCD; \citealp{preuer2018fcd}). Finally, we report the Fr\'echet inception distance (FID) and Inception Score (IS) for images.

\subsection{Language modeling}
\label{sec:headline}

\method{} lowers the few-step generative perplexity of the pre-trained MDLM by a factor of 2.3 on WikiText-103 and 1.7 on LM1B at 4 NFEs, as shown in Table~\ref{tab:main} and Figure~\ref{fig:main}. The per-position entropy stays close to the base model and held-out text (Table~\ref{tab:entropy}). Moreover, MAUVE rises from about 0.05 to about 0.3 on both corpora at 4 NFEs (App. Table~\ref{tab:textquality}). Similarly, it lowers the 4-step perplexity on SEDD, where the gains narrow as the number of steps grows, since each step then reveals fewer positions. 
\method{} lowers the 4-step perplexity by a factor of 3.8 on OpenWebText. SDTT students attain a low perplexity by narrowing their samples, with a per-position entropy well below that of held-out text on all three corpora. However, integrating \method{} with the SDTT student restores this entropy. Its entropy still stays above that of the student at a temperature of 0.9, and lowers the 4-step perplexity of the student by a factor of 1.5 to 2.2. At matched entropy, \method{} integrated with SDTT also has a lower perplexity than the distilled student of DUO on OpenWebText.
The performance gap is comparatively smaller against the distillation baselines because they have already been trained with consistency objectives to capture token dependencies.
App.~\ref{app:samples} shows samples generated by MDLM and \method{} drawn under shared reveal schedules (Figure~\ref{fig:samples}).

\begin{table}[h!]
\centering
\scriptsize
\setlength{\tabcolsep}{1.5pt}
\caption{\textbf{Few-step generation on text.} Generative perplexity (gen-PPL) across NFEs and the per-position entropy $\HH$ at 32 NFEs. \method{} and tensor-train decomposition (TTD) are integrated into the pre-trained masked diffusion models (MDM) where shaded rows show \method{} (\textit{ours}). Within each dataset, bold and underline mark the best and second-best gen-PPL in each column and the entropies closest to held-out data. }
\label{tab:main}
\resizebox{\textwidth}{!}{%
\begin{tabular}{@{}l *{8}{r} >{\hspace{4pt}}r*{7}{r} >{\hspace{4pt}}r*{7}{r}@{}}
\toprule
& \multicolumn{8}{c}{\textbf{WikiText-103} ($L = 128$)} & \multicolumn{8}{c}{\textbf{LM1B} ($L = 128$)} & \multicolumn{8}{c}{\textbf{OpenWebText} ($L = 1024$)} \\
\cmidrule(lr){2-9}\cmidrule(lr){10-17}\cmidrule(l){18-25}
& \multicolumn{7}{c}{gen-PPL $\downarrow$ at NFE} & & \multicolumn{7}{c}{gen-PPL $\downarrow$ at NFE} & & \multicolumn{7}{c}{gen-PPL $\downarrow$ at NFE} & \\
\cmidrule(lr){2-8}\cmidrule(lr){10-16}\cmidrule(lr){18-24}
model & $1$ & $2$ & $3$ & $4$ & $8$ & $16$ & $32$ & $\HH$ & $1$ & $2$ & $3$ & $4$ & $8$ & $16$ & $32$ & $\HH$ & $1$ & $2$ & $3$ & $4$ & $8$ & $16$ & $32$ & $\HH$ \\
\midrule
MDLM & 6095 & 3384 & 1940 & 1328 & 557 & 312 & 218 & 5.33 & 2457 & 1598 & 1186 & 931 & 566 & 399 & 325 & \underline{5.23} & 3850 & 3198 & 2565 & 2007 & 826 & 339 & 198 & 5.84 \\
\quad + TTD & 5349 & 3029 & 1788 & 1218 & 537 & 305 & 219 & 5.33 & 2154 & 1446 & 1050 & 861 & 546 & 411 & 338 & 5.24 & 3259 & 2751 & 2189 & 1749 & 792 & 361 & 213 & 5.82 \\
\rowcolor{gray!12}
\quad + \method{} & 1920 & 1166 & 768 & 589 & 340 & 234 & 185 & \underline{5.32} & 1023 & 775 & 636 & 543 & 407 & 338 & 297 & \underline{5.23} & 1002 & 795 & 632 & 524 & 310 & 204 & 153 & 5.81 \\
\midrule
SEDD & 4003 & 4052 & 2146 & 1382 & 638 & 362 & 247 & 5.40 & 3235 & 3253 & 1596 & 1117 & 485 & 303 & 225 & 5.35  & 3673 & 3188 & 2465 & 1856 & 723 & 310 & 188 & 5.84 \\
\quad + TTD & 5800 & 2799 & 1618 & 1122 & 530 & 305 & 226 & 5.35 & 4551 & 2290 & 1324 & 934 & 472 & 315 & 250 & 5.37  & 3154 & 2648 & 1979 & 1496 & 651 & 313 & 198 & 5.83 \\
\rowcolor{gray!12}
\quad + \method{} & 1852 & 1096 & 783 & 601 & 356 & 253 & 205 & 5.39 & 1445 & 811 & 574 & 470 & 329 & 262 & 229 & 5.37  & 2204 & 1717 & 1271 & 964 & 466 & 259 & 180 & 5.86 \\
\midrule
SDTT & \underline{1581} & \underline{817} & \underline{482} & \underline{328} & \underline{157} & \underline{94} & \underline{73} & 5.08 & 1353 & 849 & 569 & 428 & \underline{245} & \underline{159} & \underline{123} & 4.91 & 1256 & 886 & 540 & 339 & 110 & \textbf{57.5} & \textbf{40.1} & 5.52 \\
\quad + TTD & 5819 & 3030 & 1720 & 1212 & 560 & 342 & 254 & \underline{5.32} & 2121 & 1294 & 877 & 685 & 403 & 281 & 224 & \textbf{5.19} & 2765 & 2585 & 2029 & 1540 & 665 & 322 & 204 & \underline{5.79} \\
\rowcolor{gray!12}
\quad + \method{} & 1698 & 832 & 532 & 402 & 229 & 151 & 113 & 5.24 & \underline{985} & \underline{637} & \underline{481} & \underline{391} & 271 & 217 & 176 & 5.10 & \underline{938} & \underline{712} & 513 & 394 & 213 & 134 & 102 & \textbf{5.76} \\
\rowcolor{gray!12}
\quad + \method{}, $\tau{=}0.9$ & \textbf{835} & \textbf{439} & \textbf{286} & \textbf{214} & \textbf{127} & \textbf{88} & \textbf{68} & 5.10 & \textbf{547} & \textbf{374} & \textbf{281} & \textbf{235} & \textbf{165} & \textbf{133} & \textbf{109} & 4.90 & \textbf{449} & \textbf{282} & \textbf{199} & \textbf{155} & \textbf{89.4} & \underline{60.9} & \underline{47.2} & 5.52 \\
\midrule
DUO & 6528 & 1905 & 853 & 514 & 225 & 153 & 130 & \textbf{5.28} & 2457 & 1313 & 830 & 624 & 370 & 281 & 240 & 5.15 & 3768 & 1536 & 779 & 485 & 191 & 118 & 94.2 & 5.69 \\
DUO, distilled & \multicolumn{8}{c}{--} & \multicolumn{8}{c}{--} & 5597 & 905 & \underline{396} & \underline{242} & \underline{109} & 74.2 & 61.6 & 5.69 \\
\midrule
\textit{held-out data} & \multicolumn{7}{c}{} & 5.29 & \multicolumn{7}{c}{} & 5.21 & \multicolumn{7}{c}{} & 5.77 \\
\bottomrule
\end{tabular}}
\end{table}

\subsection{Molecule generation}
\label{sec:mol}

Molecular strings couple their positions through hard constraints. For instance, a SMILES string parses only if its ring-closure digits pair up and its branch parentheses balance, and two positions revealed in the same step can make the whole string invalid. As shown in Table~\ref{tab:mol}, \method{} raises the validity of MDLM from 2 NFEs onwards by 13 points at 4 NFEs on QM9 and by 4 to 11 points on ZINC-250k, and it lowers the FCD on both datasets. The distilled SDTT student also raises the validity, but it lowers the diversity of the samples and raises the FCD two- to threefold. \method{} integrated with the student lowers the FCD and leaves its validity nearly unchanged. \method{} integrated with SEDD raises validity at all evaluated steps on QM9, whereas it raises validity up to 8 NFEs and lowers it at 16 and 32 NFEs on ZINC-250k (App.~\ref{app:molextra}).

\begin{table}[h!]
\centering\footnotesize
\setlength{\tabcolsep}{4.3pt}
\caption{\textbf{Few-step generation on molecules.} Validity across NFEs, while uniqueness, novelty, and Fr\'echet ChemNet distance (FCD) are reported at 32 NFEs. \method{} and TTD are integrated into the base pre-trained MDM. The metrics are averaged over three training runs with the standard deviation as a subscript.}
\label{tab:mol}
\resizebox{\textwidth}{!}{%
\begin{tabular}{@{}l rrrrrr rrr@{}}
\toprule
& \multicolumn{6}{c}{validity (\%) $\uparrow$ at NFE} & \multicolumn{3}{c}{at 32 NFEs} \\
\cmidrule(lr){2-7}\cmidrule(lr){8-10}
model & $1$ & $2$ & $4$ & $8$ & $16$ & $32$ & unique $\uparrow$ & novel $\uparrow$ & FCD $\downarrow$ \\
\midrule
\multicolumn{10}{@{}l}{\textit{\textbf{QM9}}} \\
MDLM & \textbf{10.1}\pms{2.8} & 13.7\pms{6.5} & 24.8\pms{10.6} & 40.5\pms{9.3} & 55.8\pms{5.9} & 65.7\pms{4.4} & 79.7\pms{7.9} & 22.3\pms{3.8} & 1.36\pms{0.56} \\
MDLM + TTD& 5.4\pms{2.0} & \underline{15.7}\pms{1.9} & 33.4\pms{4.0} & 50.0\pms{4.1} & 64.4\pms{4.1} & 72.9\pms{3.6} & 75.8\pms{2.7} & 16.8\pms{1.6} & 1.52\pms{0.21} \\
\rowcolor{gray!12}
MDLM + \method{} (ours) & 2.5\pms{1.3} & 15.6\pms{1.0} & 37.8\pms{0.8} & 57.2\pms{2.0} & 68.1\pms{2.1} & 73.8\pms{2.8} & 80.1\pms{7.5} & 20.6\pms{3.5} & 1.18\pms{0.55} \\
SEDD & 7.1\pms{2.4} & 12.0\pms{9.6} & 23.7\pms{12.9} & 40.7\pms{9.2} & 56.6\pms{5.5} & 67.6\pms{3.4} & \underline{87.0}\pms{6.3} & \textbf{25.1}\pms{3.1} & \underline{0.91}\pms{0.33} \\
SEDD + TTD& 3.5\pms{0.0} & 13.7\pms{0.1} & 34.4\pms{0.5} & 54.6\pms{0.6} & 68.3\pms{0.5} & 78.1\pms{0.4} & 72.5\pms{0.5} & 14.9\pms{0.1} & 1.79\pms{0.06} \\
\rowcolor{gray!12}
SEDD + \method{} (ours) & \underline{8.6}\pms{3.8} & \textbf{17.7}\pms{2.4} & 31.0\pms{6.4} & 46.4\pms{7.2} & 60.7\pms{5.2} & 70.9\pms{3.4} & \textbf{87.6}\pms{6.3} & \underline{25.0}\pms{3.2} & \textbf{0.81}\pms{0.34} \\
SDTT & 0.0\pms{0.1} & 11.4\pms{0.4} & 38.0\pms{1.1} & 58.8\pms{1.0} & 72.7\pms{0.8} & 81.3\pms{0.7} & 59.7\pms{1.6} & 4.9\pms{0.6} & 3.90\pms{0.32} \\
SDTT + TTD& 3.5\pms{0.2} & 13.1\pms{0.7} & 31.9\pms{0.9} & 50.2\pms{0.7} & 65.0\pms{0.5} & 73.7\pms{0.4} & 74.1\pms{0.6} & 15.2\pms{0.4} & 1.62\pms{0.07} \\
\rowcolor{gray!12}
SDTT + \method{} (ours) & 0.0\pms{0.0} & 12.3\pms{0.1} & \underline{38.3}\pms{0.6} & \underline{60.2}\pms{1.0} & \underline{74.8}\pms{0.6} & \underline{82.6}\pms{0.8} & 60.7\pms{1.5} & 5.2\pms{0.4} & 3.62\pms{0.23} \\
\rowcolor{gray!12}
SDTT + \method{}, $\tau = 0.9$ (ours) & 0.0\pms{0.0} & 12.3\pms{0.0} & \textbf{38.8}\pms{0.6} & \textbf{61.3}\pms{0.9} & \textbf{75.5}\pms{0.6} & \textbf{83.2}\pms{0.7} & 59.4\pms{1.0} & 4.3\pms{0.6} & 3.87\pms{0.25} \\
\midrule
\multicolumn{10}{@{}l}{\textit{\textbf{ZINC-250k}}} \\
MDLM & 1.5\pms{0.2} & 7.5\pms{0.9} & 22.1\pms{2.4} & 36.6\pms{2.5} & 47.2\pms{3.0} & 54.2\pms{2.2} & 86.9\pms{1.8} & 86.8\pms{1.9} & 6.35\pms{0.68} \\
MDLM + TTD& \underline{2.1}\pms{1.5} & 11.8\pms{3.4} & 27.2\pms{5.1} & 42.7\pms{5.1} & 53.3\pms{4.8} & 61.3\pms{5.2} & 84.1\pms{0.8} & 83.9\pms{0.8} & 7.67\pms{0.78} \\
\rowcolor{gray!12}
MDLM + \method{} (ours) & 0.6\pms{0.0} & 11.8\pms{3.2} & 30.6\pms{3.4} & 47.1\pms{2.9} & 57.5\pms{2.3} & 62.5\pms{1.6} & 89.0\pms{2.2} & 88.8\pms{2.2} & \textbf{5.42}\pms{0.74} \\
SEDD & 1.1\pms{0.1} & 5.5\pms{1.2} & 16.1\pms{0.6} & 30.7\pms{0.7} & 41.3\pms{1.4} & 50.6\pms{1.1} & \textbf{90.1}\pms{1.3} & \textbf{90.0}\pms{1.3} & 5.54\pms{0.20} \\
SEDD + TTD& 1.0\pms{0.2} & 8.9\pms{0.9} & 25.4\pms{0.7} & 41.3\pms{0.7} & 53.2\pms{0.4} & 63.9\pms{0.1} & 86.4\pms{0.2} & 86.3\pms{0.2} & 8.15\pms{0.22} \\
\rowcolor{gray!12}
SEDD + \method{} (ours) & 1.7\pms{0.9} & 8.9\pms{1.3} & 22.8\pms{1.5} & 32.4\pms{4.7} & 40.1\pms{6.5} & 47.5\pms{6.0} & \underline{89.2}\pms{1.8} & \underline{89.0}\pms{1.8} & \underline{5.47}\pms{0.85} \\
SDTT & \textbf{2.5}\pms{3.0} & 18.1\pms{4.3} & 45.8\pms{5.5} & \underline{64.5}\pms{4.5} & \underline{78.1}\pms{3.0} & \textbf{86.0}\pms{1.9} & 75.8\pms{1.8} & 75.5\pms{1.8} & 14.04\pms{0.44} \\
SDTT + TTD& 1.7\pms{1.2} & 10.4\pms{4.1} & 23.4\pms{5.6} & 36.7\pms{5.2} & 45.9\pms{5.6} & 54.0\pms{5.2} & 86.8\pms{1.7} & 86.7\pms{1.7} & 6.51\pms{0.95} \\
\rowcolor{gray!12}
SDTT + \method{} (ours) & 0.0\pms{0.0} & \textbf{20.2}\pms{3.3} & \underline{46.3}\pms{6.0} & 64.1\pms{4.8} & 77.0\pms{3.0} & 84.1\pms{1.9} & 76.2\pms{1.8} & 75.9\pms{1.7} & 12.33\pms{0.25} \\
\rowcolor{gray!12}
SDTT + \method{}, $\tau = 0.9$ (ours) & 0.0\pms{0.0} & \underline{20.0}\pms{3.7} & \textbf{46.8}\pms{6.3} & \textbf{65.3}\pms{4.8} & \textbf{78.6}\pms{3.0} & \underline{85.9}\pms{1.7} & 75.2\pms{1.2} & 75.0\pms{1.2} & 12.67\pms{0.38} \\
\bottomrule
\end{tabular}}
\end{table}

Similarly, \method{} integrated with MDLM achieves higher validity than TTD~\citep{kim2026tensortrain} from 4 NFEs onwards (37.8$\%$ against 33.4$\%$ at 4 NFEs on QM9) and a lower FCD on both datasets. On SEDD, TTD has the higher validity from 4 NFEs on, but it lowers the uniqueness and novelty of the samples and roughly doubles the FCD on QM9 relative to \method{}. Our analysis shows that this higher validity comes from smaller molecules, whereas the two kernels are equally valid at equal string length (App.~\ref{app:molextra}). On the SDTT students, TTD achieves both lower validity and FCD. Its validity is nearly the same on all three pre-trained models of a dataset, which suggests that it learns the distribution of each position anew from the data. \method{} instead keeps the marginals of the pre-trained model, which helps in improving the few-step validity of a distilled student.
Figure~\ref{fig:molsamples} in App.~\ref{app:samples} shows sample molecules generated by MDLM and \method{}.

\vspace{-0.1in}

\subsection{Image generation}
\label{sec:image}

We also integrate \method{} into the official checkpoint of MaskGIT for ImageNet \citep{chang2022maskgit}, where the tokenizer encodes an image as 256 tokens on a $16 \times 16$ lattice. Unlike our text and molecule experiments, MaskGIT does not reveal random positions. Instead, it draws a candidate token at each masked position and keeps the most confident candidates. There is a risk that redrawing these kept positions from the chain would discard the confidence that selected them. We therefore keep MaskGIT's ranking and candidates, and let \method{} change a candidate only through the maximal coupling of the two kernels as used in speculative decoding \citep{leviathan2023speculative}. A candidate $c$ survives with probability $\min\{1, \frac{p(c)}{q(c)}\}$ along the chain over the kept positions, where $q$ and $p$ are the conditionals of the product kernel and \method{}, respectively.

\begin{table}[h!]
\centering
\footnotesize
\setlength{\tabcolsep}{4pt}
\caption{\textbf{Few-step generation on ImageNet-256.} FID on 10,000 samples across steps, while the Inception Score (IS) is shown at 4 steps. }
\label{tab:image}
\begin{tabular}{@{}l rrrrrr r@{}}
\toprule
& \multicolumn{6}{c}{FID $\downarrow$ at steps} & IS $\uparrow$ \\
\cmidrule(lr){2-7}
model & $2$ & $3$ & $4$ & $5$ & $6$ & $8$ & at 4 \\
\midrule
tokenizer reconstructions & \multicolumn{6}{c}{13.92} & -- \\
\midrule
\multicolumn{8}{@{}l}{\textit{Guidance 3, temperature 1}} \\
MaskGIT & 65.24 & 36.29 & 21.21 & 15.17 & 12.66 & \textbf{11.91} & 85.4 \\
MaskGIT + TTD& 62.60 & \underline{32.52} & \underline{19.25} & \underline{14.34} & \textbf{12.48} & \underline{12.30} & \textbf{94.3} \\
\rowcolor{gray!12}
% MaskGIT + \method{} (ours) & 61.49 & 34.24 & 21.61 & 16.26 & 13.96 & 12.79 & 80.8 \\
\rowcolor{gray!12}
MaskGIT + \method{} (ours) & \textbf{59.83} & \textbf{29.87} & \textbf{18.53} & \textbf{14.05} & \underline{12.61} & 12.67 & \underline{90.7} \\
\bottomrule
\end{tabular}
\end{table}

\method{} achieves lower FID at 2 to 5 steps with higher Inception Score as shown in Table~\ref{tab:image}, while MaskGIT catches up from 6 steps onwards under MaskGIT's sampler at a classifier-free guidance weight of 3. MaskGIT and \method{} lie within 0.3 FID of each other at a sharper temperature, and TTD trails by 8 FID at 2 steps. Overall, \method{} achieves lower FID at all evaluated steps under a random reveal order. TTD also lowers FID under guidance by less than \method{} at 2 to 5 steps and slightly more at 6 and 8 steps.

\subsection{Ablations}
\label{sec:analysis}

We vary one design choice at a time in Table~\ref{tab:ablation} to isolate their contributions in \method{}. The perplexity still decreases as the rank of the potential grows, but with the tradeoff of more parameter overhead. Hence, we propose the rank $d_p = 128$ as a lightweight module for practical purposes. Similarly, the \method{} module loses about a third of its perplexity without the gap embedding, and replacing the projected input embedding with a free token table also raises the perplexity. 
We also analyze the language consistency using the subject-verb items of the Benchmark of Linguistic Minimal Pairs (BLiMP; \citealp{warstadt2020blimp}) and found that \method{} learns the underlying grammatical relationships better than MDLM (App.~\ref{app:consistency}).

We also employ several existing unmasking approaches in \method{} that decide which masked positions to reveal in a step. Instead of the random unmasking in the \method{} sampler, these samplers choose the reveal set so that dependent positions are less likely to be unmasked together. They reveal the positions with the most confident or lowest-entropy predictions \citep{kim2025train,benhamu2025eb}, or spread the positions of a step over the sequence \citep{luxembourg2025dus}. 
Under the evaluated unmasking strategies, \method{} lowers the 4-step perplexity on WikiText-103 by a factor of 1.1 to 3.5 (App.~\ref{app:rules}). This suggests that \method{} complements a careful choice of the reveal set to further capture dependencies within a step. 

\begin{table}[h!]
\centering
\footnotesize
\setlength{\tabcolsep}{5pt}
\caption{\textbf{Ablations on WikiText-103.} Generative perplexity ($\downarrow$) at 4 and 32 NFEs, varying one choice at a time. The \method{} modules are trained for 25k steps, while the rank-64 reference is trained for 50k.}
\label{tab:ablation}
\begin{tabular}{@{}l l rr r@{}}
\toprule
kernel & variation & gen-PPL@4 & gen-PPL@32 & parameters \\
\midrule
MDLM & no module & 1304 & 214 & 0 \\
\midrule
\multicolumn{5}{@{}l}{\textit{\textbf{rank of the potential}}} \\
\method{} chain & $d_p = 32$ & 683 & 188 & 0.3M \\
\method{} chain & $d_p = 64$, 50k steps & 643 & 189 & 1.2M \\
\method{} chain & $d_p = 128$ (proposed) & \underline{589} & \underline{185} & 4.5M \\
\method{} chain & $d_p = 256$ & \textbf{564} & \textbf{182} & 17.4M \\
\midrule
\multicolumn{5}{@{}l}{\textit{\textbf{design and training,} $d_p = 64$}} \\
\method{} chain & no gap embedding & 850 & \underline{198} & 1.2M \\
\method{} chain & free token table for $E$ & \underline{677} & \textbf{192} & 4.4M \\
\bottomrule
\end{tabular}
\end{table}

\subsection{Computational cost}
\label{app:complexity}

We report the comparison of model parameters, training time, and sampling time in our WikiText-103 experiments on one NVIDIA H100 GPU in Table~\ref{tab:cost}. 
\method{} adds 4.5M parameters to a denoiser of 170M, while TTD~\citep{kim2026tensortrain} with rank 2 adds 3.5M new parameters. It takes 1.3 H100 hours (about 6$\%$ of the training time of MDLM) to train \method{}, while a quarter of the MDLM training time for TTD, whose loss passes several widened copies of the hidden state of each masked position through the output layer of the denoiser. During sampling, the sequential transitions of \method{} add 58$\%$ to the time of MDLM at 4 NFEs and 6$\%$ at 32 NFEs, where the denoiser passes dominate, while TTD makes sampling five to six times slower than MDLM.

\begin{table}[h!]
\centering
\footnotesize
\setlength{\tabcolsep}{5pt}
\caption{\textbf{Computational cost comparison on WikiText-103}. Trainable parameters, training steps and time, and sampling time per sequence at a batch size of 32. }
\label{tab:cost}
\begin{tabular}{@{}l rrr rr@{}}
\toprule
& & \multicolumn{2}{c}{training} & \multicolumn{2}{c}{sampling (ms per sequence)} \\
\cmidrule(lr){3-4}\cmidrule(lr){5-6}
model & trainable parameters & steps & H100 hours & 4 NFEs & 32 NFEs \\
\midrule
MDLM & 170.2M & 200k & 22.5 & 2.0 & 15.9 \\
SEDD & 169.6M & 200k & 23.4 & 2.5 & 20.1 \\
SDTT (distilled from MDLM) & 170.2M & 80k & 5.6 & 2.0 & 15.6 \\
DUO & 169.6M & 200k & 28.1 & 7.1 & 32.4 \\
MDLM + TTD& 3.5M & 25k & 5.3 & 12.5 & 83.5 \\
\rowcolor{gray!12}
MDLM + \method{} (ours) & 4.5M & 25k & 1.3 & 3.2 & 17.0 \\
\bottomrule
\end{tabular}
\end{table}

% =============================================================================
\section{Related work}
\label{sec:related}
% =============================================================================

\textbf{Acceleration of discrete diffusion.} Several lines of work accelerate discrete diffusion and flow models \citep{austin2021d3pm,lou2023sedd,campbell2024dfm,gat2024dfm}, of which masked diffusion is the standard instance \citep{sahoo2024mdlm,shi2024simplified,ou2024your}. They distill the sampling chain of a teacher \citep{deschenaux2025sdtt,hayakawa2024di4c,sahoo2025duo}, rectify the coupling between source and target \citep{yoo2026redi,park2025pairflow}, train with consistency objectives \citep{amin2026cdlm}, or learn flow maps over a continuous relaxation \citep{potaptchik2026dfm,lee2026fmlm}. The error of a coarse sampler has been analyzed through TC \citep{dmitriev2026sampling,zhao2026adaptation,zhang2026order}. Apart from distillation through dimensional correlations (Di4C; \citealp{hayakawa2024di4c}) and the flow maps, which carry the dependence in a latent variable or a continuous state, these methods keep the content kernel factorized. All of them train or retrain a model, whereas we change the form of the kernel on a pre-trained model without retraining it. 

\textbf{Dependence within a step.} A shared latent makes a step a mixture of factorized kernels, distilled from a teacher in Di4C and trained under a variational bound in variational masked diffusion \citep{zhang2025vmd}. Discrete copula diffusion \citep{liu2024copula} and energy-based diffusion language models \citep{xu2025edlm} correct a factorized step with a second model, while TTD adds a low-rank tensor-train joint over the masked positions and fine-tunes the pre-trained model \citep{kim2026tensortrain}. Several methods fill the reveal set from left to right with an extra network evaluated once per revealed token, a layer trained on a frozen model \citep{bansal2025adjust}, a separate small autoregressive model \citep{hersche2026codila}, or a causal block trained into the model \citep{campbell2026selfspec}, and autoregressive hybrids generate positions sequentially across blocks or in a flexible order \citep{arriola2025blockdiff,sahoo2025esolm,du2026a3,arriola2026setdiffusion}. \method{} instead models the dependence with a Markov chain whose transitions add one softmax per revealed position, with potentials computed from the same network pass.

\textbf{Choosing the reveal set.} A separate line of work keeps the kernel factorized and skips revealing dependent positions together, through confidence and planned orders \citep{kim2025train,peng2025p2}, dilated and low-discrepancy orders \citep{luxembourg2025dus,besnier2025halton}, entropy bounds and independence tests \citep{benhamu2025eb,azangulov2025punt}, and learned unmasking strategies \citep{jazbec2025unmasking,fox2026orders,asano2026where}. Other work analyzes when a parallel reveal is safe \citep{zhang2026order,bansal2026randomwalks}. DEMASK \citep{ringel2026demask} and mean-field parallel decoding \citep{zoabi2026meanfield} estimate the pairwise dependence between masked positions to choose the reveal set. We instead use this dependence to draw the revealed tokens jointly, and show that our method complements these approaches.

\section{Conclusion}

We presented \method{}, which models the conditional dependence between the tokens revealed in a denoising step with a first-order Markov chain over the revealed positions and implements its transitions with a lightweight module integrated into a pre-trained masked diffusion model. 
\method{} lowers the 4-step generative perplexity of MDLM by a factor of 1.7 to 3.8 on text, restores the sample entropy of distilled students, and raises the validity of 4-step molecule samples by up to $1.5\times$. Because \method{} applies directly to the logits, hidden states, and input embedding of a denoiser, it can be integrated naturally into any pre-trained distilled and from-scratch masked diffusion model.

\newpage

\section*{Acknowledgments}

The work used the Jetstream2 Supercomputer at Indiana University through allocation CIS251200 from the Advanced Cyberinfrastructure Coordination Ecosystem: Services \& Support (ACCESS) program~\citep{boerner2023access}, which is supported by National Science Foundation grants \#2138259, \#2138286, \#2138307, \#2137603, and \#2138296.

\bibliography{references}
\bibliographystyle{tmlr}

\newpage
\appendix

\startcontents[appendix]
\printcontents[appendix]{}{0}{\setcounter{tocdepth}{2}\section*{Appendix}}
\vspace{1em}

\newpage

\section{Theory and proofs}
\label{app:proofs}

This section provides the formal definitions and the proofs of Section~\ref{sec:theory}. The notations are listed in Table~\ref{tab:notation}.

\begin{table}[h!]
\centering
\small
\caption{Notation used in the paper.}
\label{tab:notation}
\begin{tabular}{@{}l p{0.66\textwidth}@{}}
\toprule
symbol & meaning \\
\midrule
$\Sset$, $V$ & token vocabulary and its size, excluding $\Mask$ \\
$\Mask$ & absorbing mask symbol \\
$L$ & sequence length \\
$\bx = (x^{1}, \dots, x^{L})$ & a clean sequence, $x^{i}$ its $i$-th token, $\bx^{S}$ its tokens at the positions $S$ \\
$\bz_t$ & partially masked state at time $t$ \\
$s < t$ & end and start times of a reverse step, with $t = 0$ clean and $t = 1$ fully masked \\
$\alpha_t$ & probability that a position is unmasked at time $t$, decreasing from $\alpha_0 = 1$ to $\alpha_1 = 0$ \\
$K$, $t_k$, $\Delta_k$ & number of sampling steps, the schedule $1 = t_0 > \dots > t_K = 0$, and $\Delta_k = \alpha_{t_{k+1}} - \alpha_{t_k}$ \\
NFE & number of function evaluations, network calls of a sampler \\
$\mathcal{M}(\xt)$ & positions still masked at $\xt$ \\
$S$, $n = |S|$, $r_{s,t}$ & reveal set of a step, its size, and the reveal probability $(\alpha_s - \alpha_t)/(1-\alpha_t)$ of a masked position \\
$\pi_{s,t}(S \mid \xt)$ & reveal distribution of Eq.~\ref{eq:reveal} \\
$S_k$, $R_k$ & reveal set of step $k$ and the positions revealed before step $k$ \\
$p_K$ & distribution of the samples of a $K$-step sampler \\
$q = q(\bx^{S} \mid \xt)$ & the data's conditional over the reveal set, $q_i$ and $q_{ij}$ its marginals on one and two positions \\
$P_i(\cdot \mid \xt)$, $\ell_i$, $h_i$ & clean-token posterior of the pre-trained denoiser at position $i$, its logits, and its hidden state \\
$j_1 < \dots < j_n$, $C(S)$ & positions of $S$ in index order and the set of their consecutive pairs \\
$u_i = W_u h_i$, $d_p$ & pair factor at position $i$ and its dimension \\
$E$, $A_{ij}$, $\psi_{ij}$ & token embedding, potential matrix of the pair $(i, j)$, and the potential $E_a^{\top} A_{ij} E_b$ \\
$\mathrm{PMI}_{ij}(a, b)$ & pointwise mutual information of the tokens $a$ and $b$ at the pair $(i, j)$, Remark~\ref{rem:pmi} \\
$\Fam_{\mathrm{prod}}, \Fam_{\mathrm{chain}}$ & product and chain kernels, Definition~\ref{def:families} \\
$\TC(q)$, $\rho(q)$, $\varepsilon_\theta(q)$ & total correlation, error of the best chain kernel, and transition error of \method{}, Proposition~\ref{prop:chowliu} \\
$\mathcal{L}_K$ & $K$-step bound of Proposition~\ref{prop:kbound} \\
$\HH$, $\KL$, $\I$ & entropy, Kullback-Leibler divergence, mutual information, all in nats \\
\bottomrule
\end{tabular}
\end{table}

\subsection{Conventions and kernel families}
\label{app:conventions}

Throughout this section, the vocabulary $\Sset$ is finite with $V = |\Sset| \ge 2$, and we measure entropies, divergences, and mutual information in nats. A distribution over a reveal set $S$ is a distribution on $\Sset^{S}$. For such a distribution $q$, we write $q_i$ and $q_{ij}$ for its marginals on one and two positions. The conditional $q(x^{j} \mid x^{i}) = q_{ij}(x^{i}, x^{j}) / q_i(x^{i})$ is defined where $q_i(x^{i}) > 0$, and we complete it by the uniform distribution elsewhere. We use $0 \log 0 = 0$ and set $\KL(\mu \,\|\, \nu) = +\infty$ when $\nu(\bx) = 0 < \mu(\bx)$ for some $\bx$, so that the identities below hold in $[0, +\infty]$. We also rely on three standard relations \citep{cover1991elements}.
\begin{itemize}
\item[(F1)] \emph{Gibbs' inequality.} $\KL(\mu \,\|\, \nu) \ge 0$, with equality if and only if $\mu = \nu$.
\item[(F2)] \emph{Chain rule.} If a random vector is split into blocks $Y_1, \dots, Y_m$, then $\HH(Y_1, \dots, Y_m) = \sum_{a=1}^{m} \HH(Y_a \mid Y_1, \dots, Y_{a-1})$.
\item[(F3)] \emph{Cross-entropy.} If $Y$ has the distribution $\mu$ and $\nu$ is another distribution of $Y$, then $\E_{\mu}[-\log \nu(Y)] = \HH(\mu) + \KL(\mu \,\|\, \nu)$.
\end{itemize}

\begin{definition}[Total correlation]
\label{def:tc}
For a distribution $q$ on $\Sset^{S}$, $\TC(q) = \KL\big(q \,\|\, \prod_{i \in S} q_i\big) = \sum_{i \in S} \HH(q_i) - \HH(q)$.
\end{definition}

\begin{lemma}[Error of a product kernel]
\label{lem:product}
For a distribution $q$ on $\Sset^{S}$ and each product $p = \prod_{i \in S} p_i$,
\begin{equation}
\KL(q \,\|\, p) \;=\; \TC(q) + \sum_{i \in S} \KL\big(q_i \,\|\, p_i\big).
\label{eq:prodsplit}
\end{equation}
Hence, the best product kernel is the product $\prod_{i \in S} q_i$ of the marginals of $q$, and its error is $\TC(q)$, as stated in Eq.~\ref{eq:tc}.
\end{lemma}

\emph{Proof.} We write $\log q(\bx) - \sum_{i} \log p_i(x^{i})$ as the sum of $\log q(\bx) - \sum_i \log q_i(x^{i})$ and $\sum_i \log (q_i(x^{i}) / p_i(x^{i}))$ and take the expectation of each part under $q$. The first part has the expectation $\TC(q)$ by Definition~\ref{def:tc}, and the $i$-th term of the second part depends only on the marginal $q_i$ and has the expectation $\KL(q_i \,\|\, p_i)$. By (F1), each of these divergences is nonnegative and vanishes if and only if $p_i = q_i$. \qed

\begin{definition}[Families of content kernels]
\label{def:families}
Let $S$ be a reveal set with positions $j_1 < \dots < j_n$.
\begin{itemize}
\item[(i)] $\Fam_{\mathrm{prod}}(S)$ is the set of product distributions $\prod_{i \in S} p_i$.
\item[(ii)] $\Fam_{\mathrm{chain}}(S)$ is the set of distributions $p(\bx) = p_{j_1}(x^{j_1}) \prod_{m=2}^{n} p_{j_m}(x^{j_m} \mid x^{j_{m-1}})$ with an arbitrary first factor and arbitrary transitions, i.e., the first-order Markov chains over $S$ in index order.
\end{itemize}
\end{definition}

The kernel of \method{} in Eq.~\ref{eq:kernel} is a chain kernel whose first factor is $P_{j_1}$, and a chain kernel whose transitions ignore the predecessor is a product kernel. 

\subsection{Proof of the $K$-step bound}
\label{proof:kbound}

\begin{snugshade}\noindent \textbf{Proposition~\ref{prop:kbound}} ($K$-step bound). \itshape Let a $K$-step sampler draw, at each step $k = 0, \dots, K-1$, a reveal set $S_k$ from the reveal distribution of Eq.~\ref{eq:reveal} and the tokens of $S_k$ from a content kernel $p^{\theta}(\cdot \mid \bz_{t_k}, S_k)$, and let $p_K$ be the distribution of its samples. For a clean sequence $\bx$, mask each position $i$ at an independent random time $U_i$ with $\Pr(U_i \le t) = 1 - \alpha_t$, let $S_k$ contain the positions with $t_{k+1} < U_i \le t_k$, and let $\bz_{t_k}$ show the tokens of $R_k = S_0 \cup \dots \cup S_{k-1}$. Then, for each content kernel $p^{\theta}$ and sequence $\bx$,
\[
-\log p_K(\bx) \;\le\; \mathcal{L}_K(\theta; \bx) \;=\; \E\Big[\sum_{k=0}^{K-1} -\log p^{\theta}\big(\bx^{S_k} \,\big|\, \bz_{t_k}, S_k\big)\Big],
\]
where the expectation is over the reveal sets of the forward process. Moreover, in expectation over the data,
\[
\E_{\bx \sim q}\big[\mathcal{L}_K(\theta; \bx)\big] \;=\; \HH(\bX) + \sum_{k=0}^{K-1} \E\Big[\KL\big(q_k \,\big\|\, p^{\theta}(\cdot \mid \bz_{t_k}, S_k)\big)\Big], \qquad q_k = q\big(\bx^{S_k} \,\big|\, \bz_{t_k}\big).
\]
\end{snugshade}

Each position is visible at time $t$ with probability $\Pr(U_i > t) = \alpha_t$, hence $\bz_t$ follows the forward process of Eq.~\ref{eq:forward} at each $t$. The position $i$ is revealed at step $\tau_i = k$ when $t_{k+1} < U_i \le t_k$, and $S_k = \{i : \tau_i = k\}$. Let $\Delta_k = \alpha_{t_{k+1}} - \alpha_{t_k}$ and $r_k = \Delta_k / (1 - \alpha_{t_k})$, with $\alpha_{t_0} = 0$ and $\alpha_{t_K} = 1$. Then $\Pr(\tau_i = k) = \Delta_k$, and the reveal steps are independent of each other and of $\bx$.

\begin{lemma}[Reveal probabilities]
\label{lem:reveal}
For each sequence $\bx$ and partition $(S_0, \dots, S_{K-1})$ of the positions, with $\bz_{t_k}$ showing the tokens of $\bx$ at $R_k$, the reveal distribution of Eq.~\ref{eq:reveal} assigns the probability to the sequence of reveal sets
\begin{equation}
\prod_{k=0}^{K-1} \pi_{t_{k+1}, t_k}\big(S_k \mid \bz_{t_k}\big) \;=\; \prod_{k=0}^{K-1} \Delta_k^{|S_k|},
\label{eq:revealprob}
\end{equation}
which is the probability that the forward process reveals the positions in these sets. In particular, the reveal sets do not depend on the tokens of $\bx$.
\end{lemma}

\emph{Proof.} By Eq.~\ref{eq:reveal}, step $k$ contributes the factor $r_k$ for each position of $S_k$ and the factor $1 - r_k$ for each masked position. We group these factors by position. A position revealed at step $k$ stays masked at the steps $k' < k$, and since $1 - r_{k'} = (1 - \alpha_{t_{k'+1}}) / (1 - \alpha_{t_{k'}})$, its factors telescope to
\begin{equation}
r_k \prod_{k' < k} \big(1 - r_{k'}\big) \;=\; \frac{\Delta_k}{1 - \alpha_{t_k}} \cdot \frac{1 - \alpha_{t_k}}{1 - \alpha_{t_0}} \;=\; \Delta_k.
\end{equation}
Multiplying over the positions yields Eq.~\ref{eq:revealprob}. Since the reveal steps $\tau_i$ are independent with $\Pr(\tau_i = k) = \Delta_k$, the right side is also the probability that the forward process produces the sets $S_k = \{i : \tau_i = k\}$. \qed

\emph{Step 1 (the likelihood is an average over reveal sets).} A run of the sampler that ends in $\bx$ is determined by its reveal sets, since the tokens that it draws at step $k$ must be $\bx^{S_k}$ and its states then show $\bx^{R_k}$. The probability of the run is the product over the steps of the reveal probability and of the content kernel. Summing over the partitions and applying Lemma~\ref{lem:reveal},
\begin{equation}
p_K(\bx) \;=\; \sum_{S_0, \dots, S_{K-1}} \prod_{k=0}^{K-1} \pi_{t_{k+1}, t_k}\big(S_k \mid \bz_{t_k}\big)\, p^{\theta}\big(\bx^{S_k} \,\big|\, \bz_{t_k}, S_k\big) \;=\; \E\bigg[\prod_{k=0}^{K-1} p^{\theta}\big(\bx^{S_k} \,\big|\, \bz_{t_k}, S_k\big)\bigg],
\label{eq:pkavg}
\end{equation}
where the expectation is over the forward reveal sets. A step with $S_k = \emptyset$ contributes the factor $1$.

\emph{Step 2 (Jensen's inequality).} Since the logarithm is concave, Eq.~\ref{eq:pkavg} yields $\log p_K(\bx) \ge \E[\sum_k \log p^{\theta}(\bx^{S_k} \mid \bz_{t_k}, S_k)] = -\mathcal{L}_K(\theta; \bx)$, which is Eq.~\ref{eq:kbound}.

\emph{Step 3 (the tokens of a step follow $q_k$).} Now let $\bx \sim q$. By Lemma~\ref{lem:reveal}, the reveal sets are independent of $\bx$. Hence, given the reveal sets and the revealed tokens $\bx^{R_k}$, the tokens $\bx^{S_k}$ follow the data's conditional $q_k$. By (F3), the term of step $k$ then has the conditional expectation
\begin{equation}
\E\Big[-\log p^{\theta}\big(\bx^{S_k} \,\big|\, \bz_{t_k}, S_k\big) \,\Big|\, S_0, \dots, S_{K-1}, \bx^{R_k}\Big] \;=\; \HH(q_k) + \KL\big(q_k \,\big\|\, p^{\theta}(\cdot \mid \bz_{t_k}, S_k)\big).
\end{equation}

\emph{Step 4 (the entropy terms).} By the same argument, $-\log q(\bx^{S_k} \mid \bx^{R_k})$ has the conditional expectation $\HH(q_k)$. For each partition, the chain rule of probability yields $\sum_k -\log q(\bx^{S_k} \mid \bx^{R_k}) = -\log q(\bx)$, and therefore $\sum_k \E[\HH(q_k)] = \HH(\bX)$. Summing Step 3 over the steps yields Eq.~\ref{eq:kdecomp}. \qed

\begin{corollary}[Divergence of the sampler]
\label{cor:pkdiv}
For each content kernel $p^{\theta}$, the samples of the $K$-step sampler satisfy
\begin{equation}
\KL\big(q \,\big\|\, p_K\big) \;\le\; \sum_{k=0}^{K-1} \E\Big[\KL\big(q_k \,\big\|\, p^{\theta}(\cdot \mid \bz_{t_k}, S_k)\big)\Big].
\label{eq:pkdiv}
\end{equation}
In particular, if the kernel equals the data's conditional $q_k$ at each step, then $p_K = q$.
\end{corollary}

\emph{Proof.} By (F3), $\KL(q \,\|\, p_K) = \E_{\bx \sim q}[-\log p_K(\bx)] - \HH(\bX)$. Eq.~\ref{eq:kbound} bounds the first term by $\E_{\bx \sim q}[\mathcal{L}_K(\theta; \bx)]$, and subtracting $\HH(\bX)$ from Eq.~\ref{eq:kdecomp} yields Eq.~\ref{eq:pkdiv}. When the kernel equals $q_k$ at each step, the right side of Eq.~\ref{eq:pkdiv} vanishes, and (F1) yields $p_K = q$. \qed

The inequality of Eq.~\ref{eq:kbound} is strict in general, since $p_K$ averages over the orders in which a sequence can be revealed, while $\mathcal{L}_K$ averages their log-probabilities.

\begin{remark}[Relation to the continuous-time bound]
\label{rem:ctelbo}
For a product kernel $\prod_i P_i$, each masked position of $\bz_{t_k}$ belongs to $S_k$ with probability $r_k$, so the term of step $k$ in $\mathcal{L}_K$ is the expectation of $r_k \sum_{i \in \mathcal{M}(\bz_{t_k})} -\log P_i(x^{i} \mid \bz_{t_k})$. This is the discrete-time diffusion loss of MDLM with $K$ steps \citep[Eq.~8]{sahoo2024mdlm}, whose limit as $K \to \infty$ is Eq.~\ref{eq:nelbo} \citep[Suppl.~B.2]{sahoo2024mdlm}.

The excess of $\mathcal{L}_K$ over the entropy of the data also vanishes in this limit. With exact one-position conditionals, Eq.~\ref{eq:kdecomp} and Lemma~\ref{lem:product} yield $\E_{\bx \sim q}[\mathcal{L}_K] = \HH(\bX) + \sum_k \E[\TC(q_k)]$ for the product kernel. A distribution over $n$ positions has a total correlation of at most $(n - 1)^{+} \log V \le \binom{n}{2} \log V$, and $|S_k|$ is binomial with parameters $L$ and $\Delta_k$ by Lemma~\ref{lem:reveal}. Hence,
\[
\sum_{k=0}^{K-1} \E\big[\TC(q_k)\big] \;\le\; \log V \sum_{k=0}^{K-1} \E\Big[\binom{|S_k|}{2}\Big] \;=\; \binom{L}{2} \log V \sum_{k=0}^{K-1} \Delta_k^{2} \;\le\; \binom{L}{2} \log V\, \max_k \Delta_k,
\]
which is of order $1/K$ for an equally spaced schedule. The bound therefore approaches $\HH(\bX)$, which is the value of the continuous-time bound under exact conditionals \citep{ou2024your}, and so does the bound of the best chain kernel, whose error is at most $\TC(q_k)$ at each step.
\end{remark}

\begin{remark}[Training objective]
\label{rem:objective}
The first factor of Eq.~\ref{eq:kernel} does not depend on $\theta$, so at each step the loss of Eq.~\ref{eq:loss} equals $\E[-\log p^{\theta}(\bx^{S} \mid \xt, S)]$ up to a constant. When $\nu$ is uniform over the $K$ steps of a schedule, Lemma~\ref{lem:reveal} shows that the states and reveal sets of the loss are those of the $K$-step bound, and $\mathcal{L}(\theta) = (\E_{\bx \sim q}[\mathcal{L}_K(\theta; \bx)] - c)/K$ for a constant $c$. In our experiments, $\nu$ also draws steps of random size, and the loss then mixes the step terms of several schedules. The optimal transitions stay the same, since by Proposition~\ref{prop:chowliu} the loss at each step equals a constant plus the transition error $\varepsilon_\theta(q)$, which the data's pairwise conditionals minimize for any $\nu$. The choice of $\nu$ only decides at which step sizes a module of finite capacity is most accurate.
\end{remark}

\subsection{Proof of the chain error}
\label{proof:chowliu}

\begin{snugshade}\noindent \textbf{Proposition~\ref{prop:chowliu}} (Error of the chain). \itshape Fix a state $\bz_t$ and a reveal set $S$, let $q = q(\bx^{S} \mid \bz_t)$, and number the positions $j_1 < \dots < j_n$ of $S$ as $1, \dots, n$, so that $C(S) = \{(m-1, m) : 2 \le m \le n\}$. For each chain kernel $p(\bx) = p_1(x^{1}) \prod_{m=2}^{n} p_m(x^{m} \mid x^{m-1})$ over $S$,
\[
\KL\big(q \,\big\|\, p\big) \;=\; \rho(q) + \KL\big(q_1 \,\big\|\, p_1\big) + \varepsilon(q),
\]
where the residual of the chain and the transition error are
\begin{align}
\rho(q) \;&=\; \TC(q) - \sum_{m=2}^{n} \I_q\big(X^{m-1}; X^{m}\big) \;=\; \sum_{m=3}^{n} \I_q\big(X^{m}; X^{1}, \dots, X^{m-2} \,\big|\, X^{m-1}\big), \label{eq:residual} \\
\varepsilon(q) \;&=\; \sum_{m=2}^{n} \E_{a \sim q_{m-1}}\Big[\KL\big(q(X^{m} = \cdot \mid X^{m-1} = a) \,\big\|\, p_m(\cdot \mid a)\big)\Big] \;\ge\; 0. \nonumber
\end{align}
The residual vanishes if and only if $q$ is a first-order Markov chain over $S$ in index order, and the transition error vanishes if and only if each transition equals the data's pairwise conditional wherever $q_{m-1}(a) > 0$. The best chain kernel is therefore $q^{\mathrm{c}}(\bx) = q_1(x^{1}) \prod_{m=2}^{n} q(x^{m} \mid x^{m-1})$, with the error $\rho(q)$. For the kernel $p^{\theta}$ of \method{}, with $p_1 = P_{j_1}$ and the transitions of Eq.~\ref{eq:potential}, the gain over the pre-trained product kernel is
\[
\KL\Big(q \,\Big\|\, \prod_{i \in S} P_i\Big) - \KL\big(q \,\big\|\, p^{\theta}\big) \;=\; \sum_{m=2}^{n} \I_q\big(X^{m-1}; X^{m}\big) + \sum_{m=2}^{n} \KL\big(q_m \,\big\|\, P_{j_m}\big) - \varepsilon_\theta(q),
\]
where $\varepsilon_\theta(q)$ is the transition error of $p^{\theta}$. This is Eq.~\ref{eq:gain} of the main text.
\end{snugshade}

We prove the decomposition for an arbitrary chain kernel, of which \method{} is the case $p_1 = P_{j_1}$.

\emph{Proof.} Let $q^{\mathrm{c}}(\bx) = q_1(x^{1}) \prod_{m=2}^{n} q(x^{m} \mid x^{m-1})$ be the chain kernel that keeps the first marginal and the pairwise conditionals of $q$. We show that $\rho(q) = \KL(q \,\|\, q^{\mathrm{c}})$ and then compare $q^{\mathrm{c}}$ with an arbitrary chain kernel.

\emph{Step 1 (first form of $\rho$).} Let $q(\bx) > 0$, which makes all marginals of $q$ at $\bx$ positive. Writing each conditional as $q_{m-1, m}(x^{m-1}, x^{m}) / q_{m-1}(x^{m-1})$ and multiplying and dividing by $\prod_{m \ge 2} q_m(x^{m})$ yields
\begin{equation}
q^{\mathrm{c}}(\bx) \;=\; \prod_{m=1}^{n} q_m(x^{m}) \prod_{m=2}^{n} \frac{q_{m-1, m}(x^{m-1}, x^{m})}{q_{m-1}(x^{m-1})\, q_m(x^{m})}.
\end{equation}
In particular, $q^{\mathrm{c}}(\bx) > 0$ wherever $q(\bx) > 0$. We take the logarithm and its expectation under $q$. Each term $\log q_m(X^{m})$ has the expectation $-\HH(q_m)$, and each ratio has the expectation $\I_q(X^{m-1}; X^{m})$. Therefore, $\KL(q \,\|\, q^{\mathrm{c}}) = \sum_m \HH(q_m) - \HH(q) - \sum_{m \ge 2} \I_q(X^{m-1}; X^{m})$, which is the first form of Eq.~\ref{eq:residual} by Definition~\ref{def:tc}.

\emph{Step 2 (second form of $\rho$).} Since each factor of $q^{\mathrm{c}}$ is a conditional of $q$ itself, $-\E_q[\log q^{\mathrm{c}}(\bX)] = \HH(X^{1}) + \sum_{m=2}^{n} \HH(X^{m} \mid X^{m-1})$, and the chain rule (F2) writes $\HH(q) = \HH(X^{1}) + \sum_{m=2}^{n} \HH(X^{m} \mid X^{1}, \dots, X^{m-1})$. Subtracting the second expression from the first, we obtain
\begin{equation}
\KL(q \,\|\, q^{\mathrm{c}}) \;=\; \sum_{m=2}^{n} \Big[\HH\big(X^{m} \,\big|\, X^{m-1}\big) - \HH\big(X^{m} \,\big|\, X^{1}, \dots, X^{m-1}\big)\Big] \;=\; \sum_{m=3}^{n} \I_q\big(X^{m}; X^{1}, \dots, X^{m-2} \,\big|\, X^{m-1}\big),
\end{equation}
where the term $m = 2$ vanishes. All terms vanish if and only if $q$ is a first-order Markov chain over $S$ in index order.

\emph{Step 3 (an arbitrary chain kernel).} The ratio $q^{\mathrm{c}} / p$ is a product of the factor $q_1(x^{1}) / p_1(x^{1})$ and of one factor $q(x^{m} \mid x^{m-1}) / p_m(x^{m} \mid x^{m-1})$ for each $m \ge 2$, and the expectation of the logarithm of each factor under $q$ depends only on the pair marginal $q_{m-1, m}$. Hence,
\begin{equation}
\begin{aligned}
\KL(q \,\|\, p) - \KL(q \,\|\, q^{\mathrm{c}}) \;&=\; \E_{q}\Big[\log \frac{q^{\mathrm{c}}(\bX)}{p(\bX)}\Big] \\
&=\; \KL\big(q_1 \,\|\, p_1\big) + \sum_{m=2}^{n} \E_{a \sim q_{m-1}}\Big[\KL\big(q(X^{m} = \cdot \mid X^{m-1} = a) \,\big\|\, p_m(\cdot \mid a)\big)\Big],
\end{aligned}
\end{equation}
which together with Step 1 proves the first claim. By (F1), the sum is nonnegative and vanishes if and only if each transition equals the pairwise conditional of $q$ wherever $q_{m-1}(a) > 0$. In particular, the best chain kernel is $q^{\mathrm{c}}$, and its error is $\rho(q)$.

\emph{Step 4 (the gain).} The product kernel $\prod_i P_i$ has the error $\TC(q) + \sum_{i \in S} \KL(q_i \,\|\, P_i)$ by Lemma~\ref{lem:product}. Subtracting the error of \method{} from Step 3 and inserting the first form of Eq.~\ref{eq:residual} yields Eq.~\ref{eq:gain}. \qed

Note that the gain of Eq.~\ref{eq:gain}, summed over the steps of the bound, can be evaluated without knowing $q$. On a sequence $\bx$, the difference between the $K$-step bounds of the product kernel and of \method{} is the expectation of $\sum_k \sum_{m \ge 2} \log\big(P^{\theta}_{j_m \mid j_{m-1}}(x^{j_m} \mid x^{j_{m-1}}, \bz_{t_k}) / P_{j_m}(x^{j_m} \mid \bz_{t_k})\big)$ over the reveal sets.

\begin{remark}[Optimal potential]
\label{rem:pmi}
Fix a consecutive pair $(i, j) \in C(S)$. By Proposition~\ref{prop:chowliu}, the best transition into $j$ is the data's conditional $q(b \mid a) = q_j(b) \exp\big(\mathrm{PMI}_{ij}(a, b)\big)$, where $\mathrm{PMI}_{ij}(a, b) = \log q_{ij}(a, b) - \log q_i(a) - \log q_j(b)$ is the pointwise mutual information of the tokens $a$ and $b$. This target depends on the reveal set only through the pair $(i, j)$, which is why the potential of Eq.~\ref{eq:potential} is computed from the two positions alone. The transition of Eq.~\ref{eq:potential} is $\softmax_b(\ell_j(b) + \psi_{ij}(a, b))$ with $\softmax(\ell_j) = P_j$, and it equals $q(b \mid a)$ when
\begin{equation}
\psi_{ij}(a, b) \;=\; \mathrm{PMI}_{ij}(a, b) + \log \frac{q_j(b)}{P_j(b)},
\label{eq:pmi}
\end{equation}
up to a function of $a$ alone, which the softmax removes. With exact pre-trained marginals, the optimal potential is therefore the pointwise mutual information of the pair. Otherwise, it also carries a correction of the marginal at $j$, which is the second source of gain in Eq.~\ref{eq:gain}. The module outputs the potential as the $V \times V$ matrix $E A_{ij} E^{\top}$ of rank at most $d_p$ and fits Eq.~\ref{eq:pmi} within this set. Since the fit is measured by the divergence of the resulting transitions rather than by a matrix norm, the rank of Eq.~\ref{eq:pmi} alone does not determine its error.
\end{remark}

% =============================================================================
\section{Experimental details}
\label{app:protocol}

\subsection{Datasets}

We follow the experimental setup of \citet{sahoo2024mdlm} for the WikiText-103 and LM1B datasets. The GPT-2 byte-pair tokenizer is used in WikiText-103, which has a vocabulary of $V = 50{,}257$ and a context length of $L = 128$ tokens. LM1B uses the uncased BERT WordPiece tokenizer with $V = 30{,}522$ and a context length of $L = 128$ tokens. For QM9 and ZINC-250k, we use the preprocessing, SMILES tokenizers, and split settings of \citet{park2025pairflow}. QM9 has $V = 40$, $L = 32$, 127,190 training molecules, and 6,695 held-out molecules, while ZINC-250k has $V = 72$, $L = 74$, 224,568 training molecules, and 24,887 held-out molecules.

\subsection{Architecture and training}
\label{app:arch}
% =============================================================================

We adopt the diffusion transformer (DiT) implementation of MDLM \citep{sahoo2024mdlm,peebles2023dit} with 12 blocks of width 768, an adaptive layer-norm conditioning vector of dimension 128, dropout 0.1, and untied embeddings. The \method{} module takes the hidden state after the final layer norm of the denoiser, which is the input of the output layer. Its pair projection maps this state linearly to $d_p$ dimensions, and its token embedding $E$ projects the pre-trained input embedding linearly to the same width. The gap embedding is a learned table of 16 dimensions over signed gaps clipped to $[-32, 32]$, and $g_\theta$ is a two-layer network of hidden width 256 that returns the $d_p^{2}$ entries of $A_{ij}$. \method{} roughly adds 4.5M parameters for rank $d_p = 128$, compared with the 170M parameters in the base MDLM denoiser. The training settings are listed in Table~\ref{tab:hparams}.

\begin{table}[h!]
\centering
\footnotesize
\caption{\textbf{Hyperparameters} of the pre-trained denoisers and \method{} module.}
\label{tab:hparams}
\begin{tabular}{@{}l p{0.36\textwidth} p{0.42\textwidth}@{}}
\toprule
setting & denoiser & \method{} module \\
\midrule
objective & MDLM continuous-time bound, log-linear schedule & Eq.~\ref{eq:loss}, $K = 4$, $\eta = 1/2$, random step sizes in $[1/32, 1]$ \\
denoiser & trained & frozen, averaged weights, no gradient \\
steps & 200k on text, 50k on QM9, 100k on ZINC-250k & 25k at $d_p = 128$, 50k at $d_p = 64$ \\
batch & 512 & 512 \\
optimizer & AdamW, $3 \times 10^{-4}$, 2,500 warmup steps, no weight decay & AdamW on the module alone, $10^{-3}$, 500 warmup steps, cosine decay, no weight decay \\
precision & bf16 autocast, weight average at 0.9999 & bf16 denoiser, fp32 module, weight average at 0.999 \\
potentials & none & $d_p \in \{32, 64, 128, 256\}$, $g_\theta$ of hidden width 256, output layer at zero \\
\bottomrule
\end{tabular}
\end{table}

\section{Additional results}
\label{app:addexp}

\subsection{Text quality and diversity}
\label{app:textquality}

\begin{figure}[h!]
\centering
\includegraphics[width=\textwidth]{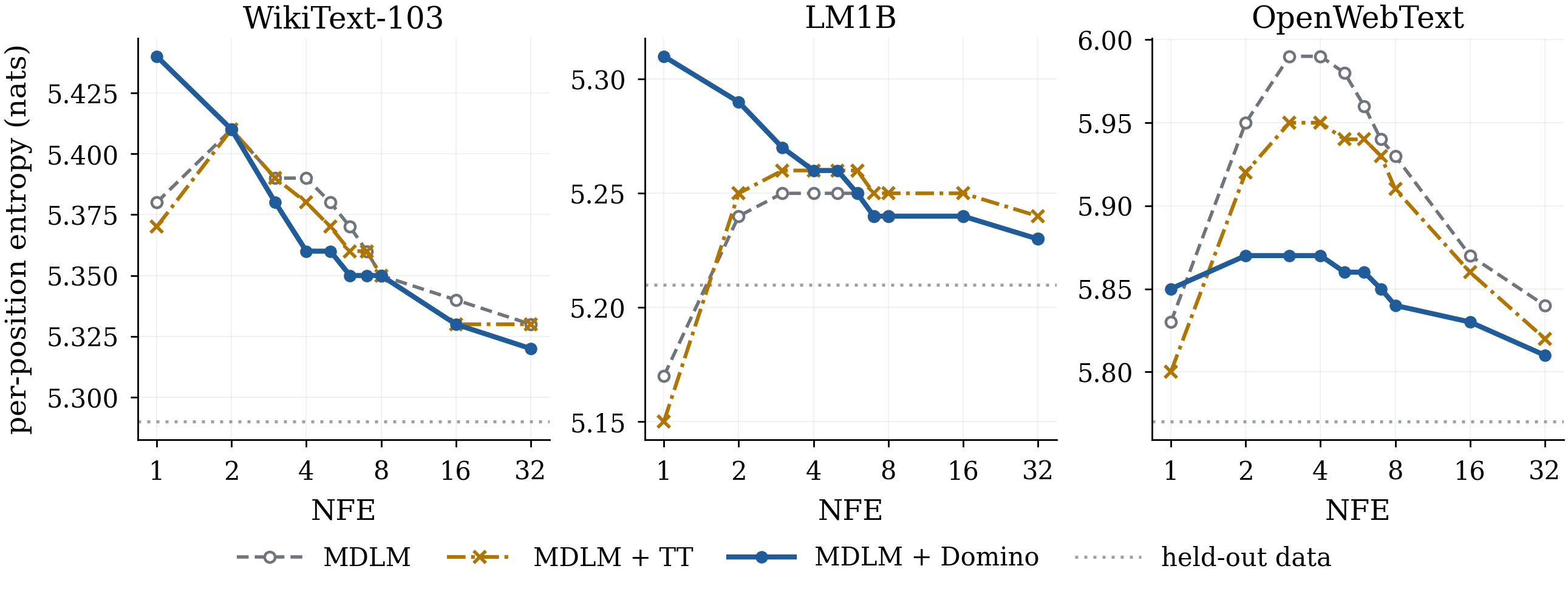}
\caption{Comparison of the per-position entropy across NFEs. }
\label{fig:entropy}
\end{figure}

\begin{figure}[h!]
\centering
\includegraphics[width=\textwidth]{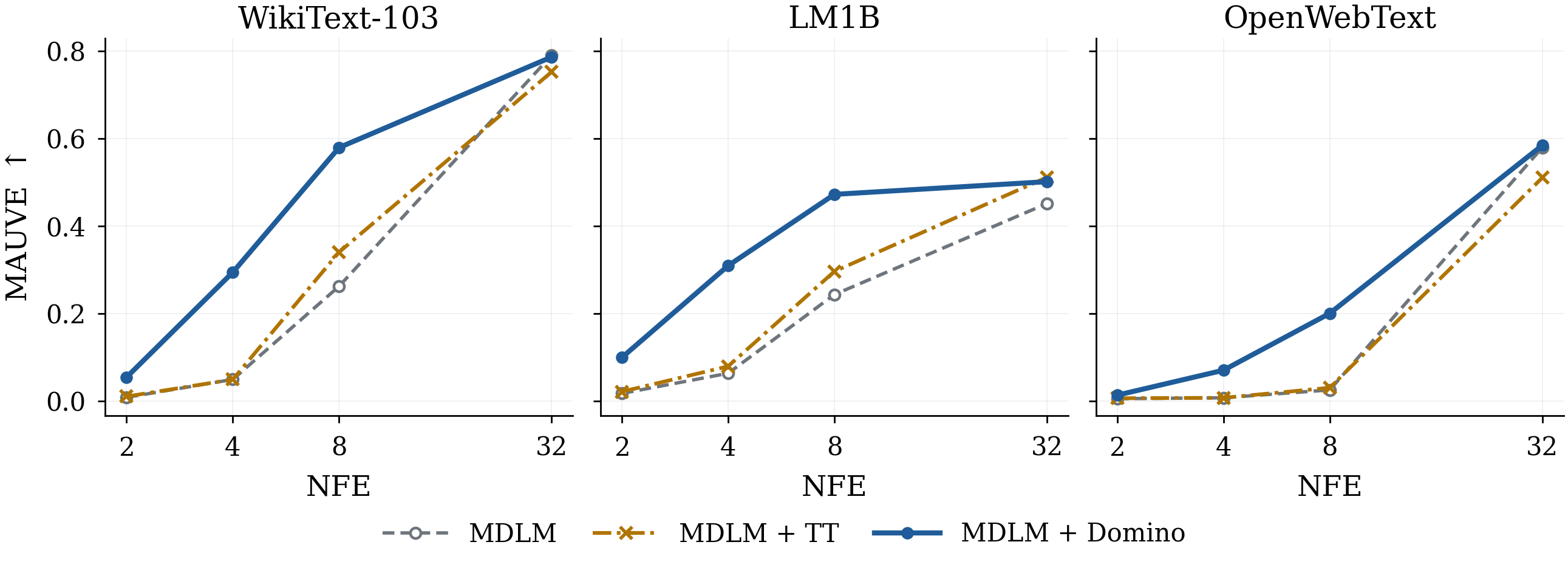}
\caption{Comparison of MAUVE against held-out text ($\uparrow$) across NFEs.}
\label{fig:quality}
\end{figure}

The entropies of the per-position marginal at each NFE are reported in Table~\ref{tab:entropy} and Figure~\ref{fig:entropy}. On WikiText-103 and LM1B, \method{} keeps the entropy of MDLM and SEDD within 0.05 nats from 3 NFEs onwards, and it moves the entropy toward the held-out text on each SDTT student.
Similarly, MAUVE and the diversity statistics are reported in Table~\ref{tab:textquality}. The share of distinct bigrams counts the repetition across the pooled samples, where a value near that of held-out text is better than a higher one, because incoherent text repeats fewer word pairs than natural text. As shown in Figure~\ref{fig:quality}, \method{} raises MAUVE at 2 and 4 NFEs on each base model and corpus, for example, from 0.049 to 0.294 at 4 NFEs for MDLM on WikiText-103. The gap narrows as the number of steps grows, except for SEDD on LM1B, where \method{} raises MAUVE further. The distilled DUO student has the highest MAUVE from 8 NFEs onwards on OpenWebText, while SDTT + \method{} at a temperature of 0.9 leads at 2 and 4 NFEs. \method{} keeps its lead over MDLM at a temperature of 0.95, where the two kernels have nearly the same per-position entropy.

\begin{table}[h!]
\centering
\scriptsize
\setlength{\tabcolsep}{2pt}
\caption{\textbf{Entropy of the per-position marginal} across NFEs. Shaded rows are ours (\method{}), and $\tau$ is the sampling temperature, which is 1 elsewhere. Within each dataset, bold marks the value closest to held-out data and underline marks the second closest. }
\label{tab:entropy}
\resizebox{\textwidth}{!}{%
\begin{tabular}{@{}l *{7}{r} >{\hspace{4pt}}r*{6}{r} >{\hspace{4pt}}r*{6}{r}@{}}
\toprule
& \multicolumn{7}{c}{\textbf{WikiText-103} ($L = 128$)} & \multicolumn{7}{c}{\textbf{LM1B} ($L = 128$)} & \multicolumn{7}{c}{\textbf{OpenWebText} ($L = 1024$)} \\
\cmidrule(lr){2-8}\cmidrule(lr){9-15}\cmidrule(l){16-22}
& \multicolumn{7}{c}{$\HH$ at NFE} & \multicolumn{7}{c}{$\HH$ at NFE} & \multicolumn{7}{c}{$\HH$ at NFE} \\
model & $1$ & $2$ & $3$ & $4$ & $8$ & $16$ & $32$ & $1$ & $2$ & $3$ & $4$ & $8$ & $16$ & $32$ & $1$ & $2$ & $3$ & $4$ & $8$ & $16$ & $32$ \\
\midrule
MDLM & 5.38 & 5.41 & 5.39 & 5.39 & 5.35 & 5.34 & 5.33 & \textbf{5.17} & \underline{5.24} & 5.25 & \underline{5.25} & \underline{5.24} & \underline{5.24} & \underline{5.23} & 5.83 & 5.95 & 5.99 & 5.99 & 5.93 & 5.87 & 5.84 \\
\quad + TTD & 5.37 & 5.41 & 5.39 & 5.38 & 5.35 & 5.33 & 5.33 & \underline{5.15} & 5.25 & 5.26 & 5.26 & 5.25 & 5.25 & 5.24 & 5.80 & 5.92 & 5.95 & 5.95 & 5.91 & 5.86 & 5.82 \\
\rowcolor{gray!12}
\quad + \method{} & 5.44 & 5.41 & 5.38 & \underline{5.36} & 5.35 & 5.33 & \underline{5.32} & 5.31 & 5.29 & 5.27 & 5.26 & \underline{5.24} & \underline{5.24} & \underline{5.23} & 5.85 & \underline{5.87} & \underline{5.87} & \underline{5.87} & \underline{5.84} & 5.83 & 5.81 \\
\midrule
SEDD & 5.44 & 5.45 & 5.43 & 5.42 & 5.42 & 5.40 & 5.40 & 5.42  & 5.41  & 5.38  & 5.39  & 5.36  & 5.36  & 5.35  & 5.80 & 5.96 & 6.00 & 6.00 & 5.92 & 5.86 & 5.84 \\
\quad + TTD & \underline{5.36} & 5.42 & 5.42 & 5.41 & 5.39 & 5.37 & 5.35 & 5.35  & 5.42  & 5.41  & 5.40  & 5.38  & 5.38  & 5.37  & 5.79 & 5.91 & 5.94 & 5.95 & 5.90 & 5.85 & 5.83 \\
\rowcolor{gray!12}
\quad + \method{} & 5.43 & 5.43 & 5.42 & 5.41 & 5.39 & 5.39 & 5.39 & 5.41  & 5.39  & 5.38  & 5.38  & 5.37  & 5.37  & 5.37  & \textbf{5.77} & 5.91 & 5.94 & 5.94 & 5.91 & 5.87 & 5.86 \\
\midrule
SDTT & 4.79 & 4.95 & 5.02 & 5.04 & 5.08 & 5.07 & 5.08 & 4.81 & 4.93 & 4.93 & 4.95 & 4.95 & 4.93 & 4.91 & 5.26 & 5.36 & 5.40 & 5.43 & 5.49 & 5.52 & 5.52 \\
\quad + TTD & 5.39 & 5.40 & 5.38 & 5.37 & 5.35 & 5.33 & \underline{5.32} & 5.06 & \textbf{5.18} & \underline{5.20} & \textbf{5.20} & \textbf{5.20} & \textbf{5.20} & \textbf{5.19} & 5.73 & \underline{5.87} & 5.89 & 5.89 & 5.85 & \underline{5.81} & \underline{5.79} \\
\rowcolor{gray!12}
\quad + \method{} & 5.41 & \underline{5.36} & \textbf{5.33} & \textbf{5.31} & \textbf{5.29} & \textbf{5.26} & 5.24 & 5.28 & \underline{5.24} & \textbf{5.21} & \textbf{5.20} & 5.16 & 5.14 & 5.10 & 5.80 & \textbf{5.84} & \textbf{5.83} & \textbf{5.82} & \textbf{5.79} & \textbf{5.77} & \textbf{5.76} \\
\rowcolor{gray!12}
\quad + \method{}, $\tau{=}0.9$ & 5.14 & 5.14 & 5.14 & 5.14 & 5.13 & 5.12 & 5.10 & 5.01 & 5.00 & 4.99 & 4.98 & 4.96 & 4.95 & 4.90 & 5.43 & 5.50 & 5.51 & 5.52 & 5.53 & 5.53 & 5.52 \\
\midrule
DUO & \textbf{5.33} & \textbf{5.22} & \underline{5.20} & 5.21 & \underline{5.24} & \underline{5.25} & \textbf{5.28} & 4.86 & 5.01 & 5.02 & 5.04 & 5.09 & 5.13 & 5.15 & \underline{5.78} & 5.64 & 5.58 & 5.59 & 5.63 & 5.66 & 5.69 \\
DUO, distilled & \multicolumn{7}{c}{--} & \multicolumn{7}{c}{--} & 5.99 & 5.55 & 5.50 & 5.53 & 5.60 & 5.64 & 5.69 \\
\midrule
\textit{held-out data} & \multicolumn{7}{c}{5.29} & \multicolumn{7}{c}{5.21} & \multicolumn{7}{c}{5.77} \\
\bottomrule
\end{tabular}}
\end{table}

\begin{table}[h!]
\centering
\scriptsize
\setlength{\tabcolsep}{2pt}
\caption{\textbf{Quality and diversity of text samples.} MAUVE across NFEs, and the per-sample token entropy (ent.) and share of distinct bigrams (dist-2) at 4 NFEs, where closer to held-out text is better. Shaded rows are ours, and $\tau$ is the sampling temperature, which is 1 elsewhere. Within each dataset, bold and underline mark the best and second-best MAUVE in each column. }
\label{tab:textquality}
\resizebox{\textwidth}{!}{%
\begin{tabular}{@{}l rrrr>{\hspace{3pt}}rr >{\hspace{4pt}}rrrr>{\hspace{3pt}}rr >{\hspace{4pt}}rrrr>{\hspace{3pt}}rr@{}}
\toprule
& \multicolumn{6}{c}{\textbf{WikiText-103} ($L = 128$)} & \multicolumn{6}{c}{\textbf{LM1B} ($L = 128$)} & \multicolumn{6}{c}{\textbf{OpenWebText} ($L = 1024$)} \\
\cmidrule(lr){2-7}\cmidrule(lr){8-13}\cmidrule(l){14-19}
& \multicolumn{4}{c}{MAUVE $\uparrow$ at NFE} & \multicolumn{2}{c}{at 4 NFEs} & \multicolumn{4}{c}{MAUVE $\uparrow$ at NFE} & \multicolumn{2}{c}{at 4 NFEs} & \multicolumn{4}{c}{MAUVE $\uparrow$ at NFE} & \multicolumn{2}{c}{at 4 NFEs} \\
\cmidrule(lr){2-5}\cmidrule(lr){6-7}\cmidrule(lr){8-11}\cmidrule(lr){12-13}\cmidrule(lr){14-17}\cmidrule(lr){18-19}
model & $2$ & $4$ & $8$ & $32$ & ent. & dist-2 & $2$ & $4$ & $8$ & $32$ & ent. & dist-2 & $2$ & $4$ & $8$ & $32$ & ent. & dist-2 \\
\midrule
MDLM & 0.008 & 0.049 & 0.262 & 0.790 & 4.44 & 0.75 & 0.017 & 0.063 & 0.243 & 0.451 & 4.45 & 0.67 & 0.005 & 0.007 & 0.025 & 0.579 & 6.00 & 0.66 \\
\quad + TTD & 0.010 & 0.049 & 0.340 & 0.752 & 4.43 & 0.75 & 0.021 & 0.079 & 0.296 & \textbf{0.511} & 4.44 & 0.67 & 0.006 & 0.007 & 0.030 & 0.510 & 5.96 & 0.65 \\
\rowcolor{gray!12}
\quad + \method{} & 0.054 & 0.294 & 0.579 & 0.786 & 4.42 & 0.69 & 0.099 & 0.309 & \underline{0.472} & \underline{0.501} & 4.43 & 0.63 & 0.013 & \underline{0.070} & 0.200 & 0.584 & 5.84 & 0.54 \\
\midrule
MDLM, $\tau{=}0.95$ & 0.008 & 0.042 & 0.348 & 0.759 & 4.36 & 0.70 & \multicolumn{6}{c}{--} & \multicolumn{6}{c}{--} \\
\rowcolor{gray!12}
\quad + \method{}, $\tau{=}0.95$ & 0.065 & 0.288 & \underline{0.640} & 0.807 & 4.35 & 0.65 & \multicolumn{6}{c}{--} & \multicolumn{6}{c}{--} \\
\midrule
SEDD & 0.021 & 0.118 & 0.379 & 0.725 & 4.43 & 0.73 & 0.058  & 0.333  & 0.560  & 0.684  & 4.44  & 0.69  & 0.005 & 0.007 & 0.035 & 0.516 & 6.00 & 0.66 \\
\quad + TTD & 0.008 & 0.067 & 0.267 & 0.723 & 4.43 & 0.74 & 0.017  & 0.113  & 0.616  & 0.865  & 4.45  & 0.72  & 0.006 & 0.009 & 0.034 & 0.437 & 5.95 & 0.63 \\
\rowcolor{gray!12}
\quad + \method{} & 0.057 & 0.245 & 0.524 & 0.767 & 4.43 & 0.69 & 0.152  & 0.489  & 0.840  & 0.914  & 4.45  & 0.66  & 0.007 & 0.023 & 0.080 & 0.590 & 5.92 & 0.60 \\
\midrule
SDTT & 0.008 & 0.081 & 0.420 & 0.804 & 4.21 & 0.57 & 0.016 & 0.091 & 0.252 & 0.404 & 4.25 & 0.55 & 0.005 & 0.009 & 0.174 & 0.764 & 5.42 & 0.41 \\
\quad + TTD & 0.010 & 0.046 & 0.256 & 0.683 & 4.43 & 0.74 & 0.019 & 0.073 & 0.300 & 0.481 & 4.41 & 0.64 & 0.006 & 0.007 & 0.032 & 0.390 & 5.90 & 0.62 \\
\rowcolor{gray!12}
\quad + \method{} & \underline{0.072} & \underline{0.296} & 0.619 & 0.793 & 4.38 & 0.66 & \underline{0.138} & \textbf{0.361} & \textbf{0.496} & 0.444 & 4.38 & 0.59 & \underline{0.019} & 0.066 & 0.393 & 0.801 & 5.79 & 0.52 \\
\rowcolor{gray!12}
\quad + \method{}, $\tau{=}0.9$ & \textbf{0.095} & \textbf{0.457} & \textbf{0.691} & \underline{0.833} & 4.26 & 0.57 & \textbf{0.160} & \underline{0.344} & 0.401 & 0.420 & 4.25 & 0.50 & \textbf{0.023} & \textbf{0.171} & \underline{0.478} & \underline{0.813} & 5.48 & 0.39 \\
\midrule
DUO & 0.009 & 0.140 & 0.599 & \textbf{0.839} & 4.33 & 0.64 & 0.016 & 0.095 & 0.322 & 0.403 & 4.34 & 0.61 & 0.005 & 0.015 & 0.214 & 0.747 & 5.58 & 0.48 \\
DUO, distilled & \multicolumn{6}{c}{--} & \multicolumn{6}{c}{--} & 0.006 & 0.059 & \textbf{0.537} & \textbf{0.894} & 5.51 & 0.43 \\
\midrule
\textit{held-out data} & \multicolumn{4}{c}{0.999} & 4.26 & 0.53 & \multicolumn{4}{c}{0.973} & 4.43 & 0.55 & \multicolumn{4}{c}{0.991} & 5.45 & 0.41 \\
\bottomrule
\end{tabular}}
\end{table}

\subsection{Stability and diversity of molecule samples}
\label{app:molextra}

\begin{figure}[h!]
\centering
\includegraphics[width=\textwidth]{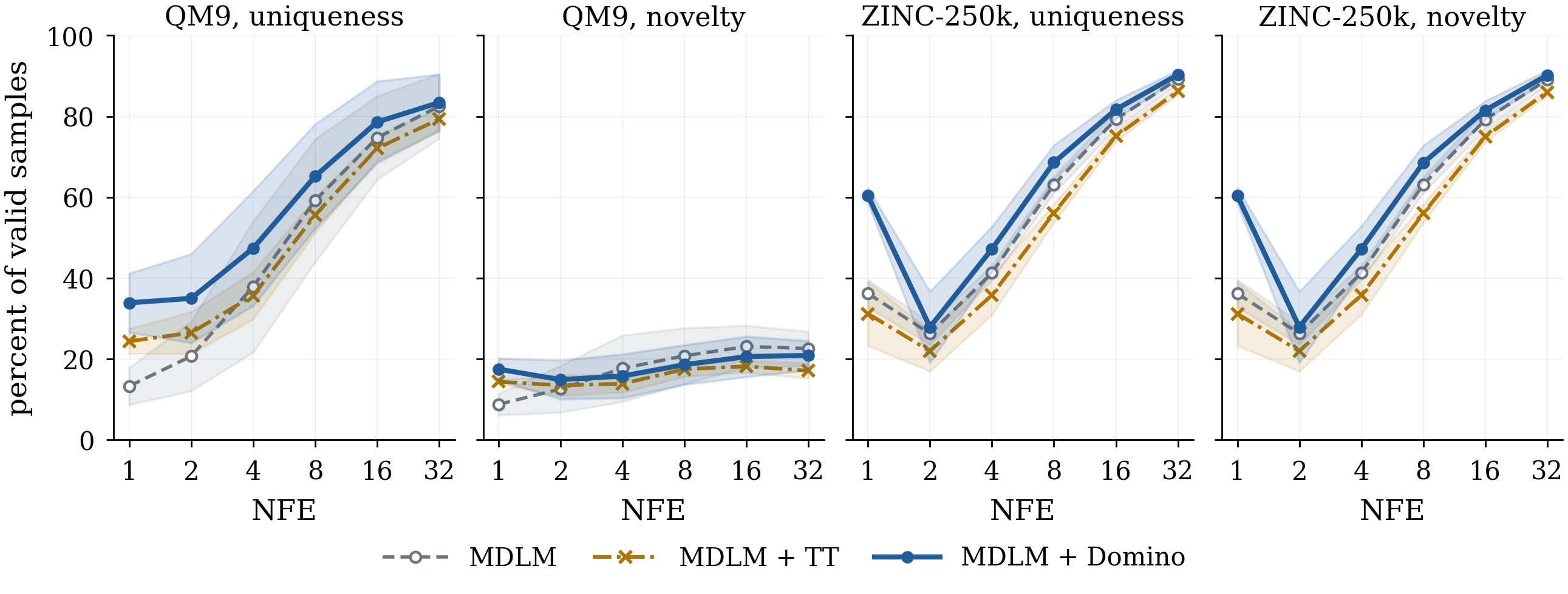}
\caption{\textbf{Diversity of valid molecules} ($\uparrow$) across NFEs on the QM9 and ZINC-250k datasets. }
\label{fig:molun}
\end{figure}

We report the uniqueness and novelty of the valid molecules at each number of NFEs in Table~\ref{tab:molun} and Figure~\ref{fig:molun}. We note in our experiments that a few-step sampler queries the model mostly at heavy masking, where the training loss puts little weight, and two training runs with the same loss can therefore differ widely at few steps. On QM9, for example, two SEDD runs with nearly equal validation losses have 16.0 and 35.5$\%$ validity at 4 NFEs. To account for this variability, we train the molecule models three times and report means and standard deviations. The 4-step validity of MDLM on QM9 ranges from 13.2 to 33.8$\%$ over these runs, while \method{} ranges from 37.0 to 38.7 percent. Moreover, \method{} raises the uniqueness of MDLM at 4 NFEs by 6 to 10 points and leaves that of SEDD nearly unchanged. However, it does not restore the novelty of an SDTT student, likely because it keeps the marginals that the student has learned.

\begin{table}[h!]
\centering\scriptsize
\setlength{\tabcolsep}{2.6pt}
\caption{\textbf{Uniqueness and novelty of valid molecules} ($\uparrow$, \%) across NFEs on 4,096 samples. A dash marks fewer than 0.1\% valid molecules, and the metrics are averaged over three training runs with the standard deviation as a subscript.}
\label{tab:molun}
\resizebox{\textwidth}{!}{%
\begin{tabular}{@{}l rrrrrr rrrrrr@{}}
\toprule
& \multicolumn{6}{c}{uniqueness (\%) at NFE} & \multicolumn{6}{c}{novelty (\%) at NFE} \\
\cmidrule(lr){2-7}\cmidrule(lr){8-13}
model & $1$ & $2$ & $4$ & $8$ & $16$ & $32$ & $1$ & $2$ & $4$ & $8$ & $16$ & $32$ \\
\midrule
\multicolumn{13}{@{}l}{\textit{\textbf{QM9}}} \\
MDLM & 13.3\pms{4.6} & 20.7\pms{8.6} & 37.9\pms{16.2} & 59.2\pms{15.2} & 74.7\pms{10.3} & 82.5\pms{8.0} & 8.8\pms{2.7} & 12.6\pms{5.8} & 17.7\pms{8.2} & 20.8\pms{6.9} & 23.1\pms{5.2} & 22.6\pms{4.2} \\
MDLM + TTD& 24.4\pms{3.1} & 26.5\pms{5.2} & 35.7\pms{5.8} & 55.6\pms{4.3} & 72.1\pms{3.1} & 79.3\pms{2.7} & 14.4\pms{0.5} & 13.5\pms{2.6} & 13.9\pms{2.1} & 17.5\pms{1.8} & 18.2\pms{1.5} & 17.1\pms{1.9} \\
\rowcolor{gray!12}
MDLM + \method{} (ours) & \textbf{33.9}\pms{7.3} & \underline{35.0}\pms{11.0} & 47.4\pms{14.2} & 65.2\pms{12.9} & 78.6\pms{10.1} & 83.4\pms{7.0} & \underline{17.5}\pms{2.7} & 14.9\pms{4.8} & 15.8\pms{5.4} & 18.6\pms{4.9} & 20.6\pms{5.0} & 20.9\pms{3.6} \\
SEDD & \underline{27.8}\pms{4.7} & 28.8\pms{6.1} & \underline{51.2}\pms{15.6} & \underline{71.4}\pms{15.1} & \underline{83.6}\pms{9.8} & \underline{89.6}\pms{5.8} & \textbf{18.9}\pms{4.8} & \underline{15.9}\pms{2.3} & \textbf{23.6}\pms{7.1} & \textbf{28.4}\pms{8.1} & \textbf{28.2}\pms{4.8} & \textbf{25.4}\pms{3.0} \\
SEDD + TTD& 24.4\pms{0.3} & 26.2\pms{0.6} & 33.3\pms{0.7} & 52.7\pms{0.2} & 69.7\pms{0.6} & 76.4\pms{0.6} & 14.9\pms{0.3} & 13.3\pms{0.1} & 13.9\pms{0.3} & 16.3\pms{0.3} & 16.7\pms{0.1} & 15.1\pms{0.2} \\
\rowcolor{gray!12}
SEDD + \method{} (ours) & 21.7\pms{11.1} & \textbf{39.3}\pms{8.6} & \textbf{54.9}\pms{10.4} & \textbf{74.4}\pms{10.6} & \textbf{85.6}\pms{7.3} & \textbf{90.0}\pms{5.5} & 12.8\pms{5.6} & \textbf{19.6}\pms{4.4} & \underline{22.6}\pms{4.7} & \underline{28.0}\pms{6.5} & \underline{27.1}\pms{4.3} & \underline{25.1}\pms{3.3} \\
SDTT & -- & 7.2\pms{0.7} & 15.0\pms{0.7} & 35.5\pms{0.9} & 54.4\pms{0.7} & 65.9\pms{1.0} & -- & 3.1\pms{0.5} & 4.3\pms{0.3} & 5.6\pms{0.4} & 5.5\pms{0.3} & 5.0\pms{0.7} \\
SDTT + TTD& 27.1\pms{0.1} & 28.9\pms{1.3} & 34.5\pms{1.3} & 54.0\pms{1.1} & 70.6\pms{0.9} & 77.9\pms{0.6} & 14.5\pms{0.8} & 14.2\pms{0.3} & 12.9\pms{0.6} & 15.2\pms{0.5} & 16.0\pms{0.3} & 15.3\pms{0.4} \\
\rowcolor{gray!12}
SDTT + \method{} (ours) & -- & 7.9\pms{0.2} & 16.5\pms{0.5} & 37.0\pms{0.4} & 56.2\pms{0.8} & 65.5\pms{0.5} & -- & 3.2\pms{0.4} & 3.2\pms{0.2} & 5.7\pms{0.2} & 6.2\pms{0.3} & 5.3\pms{0.5} \\
\rowcolor{gray!12}
SDTT + \method{}, $\tau = 0.9$ (ours) & -- & 7.6\pms{0.3} & 16.3\pms{0.5} & 37.0\pms{0.5} & 55.7\pms{0.7} & 65.0\pms{0.9} & -- & 2.7\pms{0.3} & 2.9\pms{0.2} & 4.8\pms{0.2} & 5.0\pms{0.2} & 4.3\pms{0.5} \\
\midrule
\multicolumn{13}{@{}l}{\textit{\textbf{ZINC-250k}}} \\
MDLM & 36.2\pms{3.4} & 26.3\pms{2.7} & 41.3\pms{1.8} & 63.1\pms{2.0} & 79.3\pms{2.3} & 89.2\pms{1.3} & 36.2\pms{3.4} & 26.3\pms{2.7} & 41.3\pms{1.8} & 63.1\pms{2.0} & 79.2\pms{2.3} & 89.1\pms{1.3} \\
MDLM + TTD& 31.2\pms{7.9} & 22.0\pms{5.0} & 35.8\pms{4.8} & 56.1\pms{2.3} & 75.1\pms{1.3} & 86.2\pms{0.8} & 31.2\pms{7.9} & 22.0\pms{5.0} & 35.8\pms{4.8} & 56.0\pms{2.3} & 75.0\pms{1.3} & 86.0\pms{0.8} \\
\rowcolor{gray!12}
MDLM + \method{} (ours) & \textbf{60.4}\pms{1.9} & \underline{27.9}\pms{8.8} & 47.2\pms{5.6} & \underline{68.6}\pms{4.3} & \underline{81.7}\pms{2.3} & 90.3\pms{1.1} & \textbf{60.4}\pms{1.9} & \underline{27.9}\pms{8.8} & 47.2\pms{5.7} & \underline{68.5}\pms{4.3} & \underline{81.5}\pms{2.3} & 90.2\pms{1.2} \\
SEDD & 41.2\pms{1.4} & 26.8\pms{2.7} & \textbf{52.5}\pms{1.8} & \textbf{72.2}\pms{0.9} & \textbf{84.8}\pms{1.2} & \textbf{91.6}\pms{1.2} & 41.2\pms{1.4} & 26.8\pms{2.7} & \textbf{52.5}\pms{1.8} & \textbf{72.1}\pms{0.9} & \textbf{84.7}\pms{1.2} & \textbf{91.5}\pms{1.2} \\
SEDD + TTD& 39.4\pms{7.2} & 25.9\pms{3.9} & 37.7\pms{2.3} & 59.0\pms{1.2} & 78.0\pms{0.4} & 88.4\pms{0.4} & 39.4\pms{7.2} & 25.9\pms{3.9} & 37.7\pms{2.3} & 58.9\pms{1.2} & 77.9\pms{0.4} & 88.2\pms{0.4} \\
\rowcolor{gray!12}
SEDD + \method{} (ours) & \underline{41.8}\pms{16.0} & \textbf{38.7}\pms{7.5} & \underline{51.1}\pms{6.6} & 67.2\pms{6.2} & 81.6\pms{4.0} & \underline{90.9}\pms{1.7} & \underline{41.8}\pms{16.0} & \textbf{38.7}\pms{7.5} & \underline{51.0}\pms{6.6} & 67.2\pms{6.1} & \underline{81.5}\pms{3.9} & \underline{90.8}\pms{1.7} \\
SDTT & -- & 13.2\pms{3.2} & 20.2\pms{2.0} & 40.1\pms{2.5} & 64.1\pms{1.7} & 78.6\pms{1.7} & -- & 13.2\pms{3.2} & 20.2\pms{2.0} & 40.0\pms{2.5} & 63.9\pms{1.8} & 78.4\pms{1.7} \\
SDTT + TTD& 34.4\pms{9.2} & 24.7\pms{5.7} & 38.1\pms{4.7} & 58.6\pms{3.7} & 77.9\pms{2.9} & 88.8\pms{1.6} & 34.4\pms{9.2} & 24.7\pms{5.7} & 38.1\pms{4.7} & 58.6\pms{3.8} & 77.9\pms{2.9} & 88.7\pms{1.7} \\
\rowcolor{gray!12}
SDTT + \method{} (ours) & -- & 8.7\pms{2.1} & 21.9\pms{2.7} & 43.2\pms{2.8} & 63.6\pms{2.4} & 79.3\pms{1.5} & -- & 8.7\pms{2.1} & 21.9\pms{2.7} & 43.1\pms{2.8} & 63.5\pms{2.5} & 79.0\pms{1.4} \\
\rowcolor{gray!12}
SDTT + \method{}, $\tau = 0.9$ (ours) & -- & 7.7\pms{1.7} & 21.0\pms{2.5} & 41.9\pms{2.5} & 62.9\pms{2.1} & 78.5\pms{1.4} & -- & 7.7\pms{1.7} & 21.0\pms{2.5} & 41.9\pms{2.5} & 62.8\pms{2.1} & 78.3\pms{1.3} \\
\bottomrule
\end{tabular}}
\end{table}

\begin{table}[h!]
\centering
\scriptsize
\setlength{\tabcolsep}{3pt}
\caption{\textbf{Quantitative analysis of the invalid molecules at 32 NFEs.} Share of samples (\%) that are valid, empty, leave a ring-closure label open, have unbalanced parentheses (paren.), or break another rule of SMILES syntax or chemistry (valence, aromaticity). }
\label{tab:molfail}
\begin{tabular}{@{}l rrrrrr rrrrrr@{}}
\toprule
& \multicolumn{6}{c}{QM9} & \multicolumn{6}{c}{ZINC-250k} \\
\cmidrule(lr){2-7}\cmidrule(lr){8-13}
model & valid & empty & open ring & paren. & other & atoms & valid & empty & open ring & paren. & other & atoms \\
\midrule
SEDD & 66.1 & 1.9 & 17.7 & 7.0 & 7.3 & 8.2 & 50.2 & 1.7 & 15.7 & 19.5 & 12.8 & 16.7 \\
SEDD + TTD& 78.2 & 8.1 & 6.1 & 2.8 & 4.8 & 7.0 & 64.0 & 5.2 & 10.3 & 11.1 & 9.5 & 14.0 \\
\rowcolor{gray!12}
SEDD + \method{} (ours) & 70.5 & 2.8 & 14.5 & 5.8 & 6.4 & 8.2 & 49.1 & 3.8 & 16.2 & 19.6 & 11.4 & 17.3 \\
SDTT & 81.1 & 15.5 & 1.9 & 0.6 & 1.0 & 6.7 & 85.3 & 8.4 & 1.9 & 2.2 & 2.2 & 11.3 \\
SDTT + TTD& 73.6 & 9.8 & 7.8 & 4.2 & 4.6 & 7.1 & 51.1 & 2.3 & 15.6 & 19.7 & 11.5 & 15.8 \\
\rowcolor{gray!12}
SDTT + \method{} (ours) & 82.6 & 15.1 & 1.4 & 0.4 & 0.5 & 6.6 & 83.7 & 8.6 & 2.5 & 2.2 & 3.0 & 11.6 \\
\textit{held-out data} & -- & -- & -- & -- & -- & 8.8 & -- & -- & -- & -- & -- & 23.2 \\
\bottomrule
\end{tabular}
\end{table}

We further classify the samples at 32 NFEs by the first rule of SMILES syntax or chemistry that they break in Table~\ref{tab:molfail}. TTD leaves fewer rings open and fewer parentheses unbalanced than \method{} on SEDD, but it also produces more empty strings and smaller molecules, with 7 heavy atoms against 8.2 on QM9 and 14 against 17.3 on ZINC-250k, where held-out molecules have 8.8 and 23.2 atoms. We also compare the validity of strings of equal length because a short string has fewer chances to leave a ring open. On ZINC-250k, the two kernels then have nearly the same validity and share of open rings in each range of length. For example, 88.9 and 88.7$\%$ validity for strings of at most 20 characters. Hence, the higher validity of TTD mainly comes from its shorter molecules, which also raise its FCD. On QM9, TTD remains 3 to 5 points more valid at equal length and leaves fewer rings open, which suggests that the state it carries through the whole sequence helps close rings whose two labels are far apart, while the chain of \method{} links only neighboring revealed positions. However, the pattern reverses on the SDTT students. TTD generates longer molecules towards those of the data, to 15.8 heavy atoms against 11.6 for \method{} on ZINC-250k. It then leaves rings open far more often at each length, for example, 9.9 against 1.3$\%$ of strings of at most 20 characters, which accounts for its lower validity.

\section{Supplementary analyses}
\label{app:analysis}

% =============================================================================
\subsection{Unmasking approaches}
\label{app:rules}

We replace the random masking in \method{} with several unmasking approaches that decide the positions to be revealed in a denoising step.
These include confidence strategies that first reveal the positions whose predictions have the highest top-1 probability, the largest gap between the two most probable tokens, or the lowest entropy \citep{kim2025train}. The dilated unmasking scheduler (DUS; \citealp{luxembourg2025dus}) spreads the positions of a step over the sequence, so that positions revealed together lie far apart. The entropy-bounded sampler (EB-Sampler; \citealp{benhamu2025eb}) reveals low-entropy positions until their summed entropy reaches a bound. Finally, the strategy of MaskGIT first draws a token at each masked position and then reveals the positions whose drawn tokens are most probable.

The generative perplexity and entropy of both kernels under each unmasking approach are reported in Table~\ref{tab:rules}. \method{} lowers the perplexity at 2 and 4 NFEs on both corpora under each strategy, and the per-position entropy changes by at most 0.02 nats. The gain is smallest under DUS, whose reveal sets contain positions that are far apart and nearly independent at 4 NFEs. Under the strategy of MaskGIT on WikiText-103, the product kernel has the lower perplexity from 8 NFEs onwards. However, its entropy at 4 NFEs is already 3.76 nats against 4.69 for \method{}, which indicates narrower samples. These results show that a careful choice of the reveal set reduces the dependence within a step.

\begin{table}[h!]
\centering
\footnotesize
\setlength{\tabcolsep}{3pt}
\caption{\textbf{Unmasking strategies with and without \method{}.} Generative perplexity ($\downarrow$) across NFEs, while the per-position entropy is shown at 4 NFEs. }
\label{tab:rules}
\begin{tabular}{@{}ll rrrr r rrrr r@{}}
\toprule
& & \multicolumn{5}{c}{WikiText-103} & \multicolumn{5}{c}{LM1B} \\
\cmidrule(lr){3-7}\cmidrule(lr){8-12}
strategy & kernel & $2$ & $4$ & $8$ & $16$ & $H$@4 & $2$ & $4$ & $8$ & $16$ & $H$@4 \\
\midrule
random (ancestral) & MDLM & 3361 & 1330 & 560 & 307 & 5.38 & 1621 & 927 & 570 & 394 & 5.25 \\
 & \cellcolor{gray!12}\method{} & \textbf{1148} & \textbf{586} & \textbf{339} & \textbf{230} & 5.37 & \textbf{767} & \textbf{539} & \textbf{412} & \textbf{335} & 5.26 \\
dilated (DUS) & MDLM & 1723 & 468 & 255 & 191 & 5.39 & 1053 & 474 & \textbf{331} & \textbf{298} & 5.26 \\
 & \cellcolor{gray!12}\method{} & \textbf{968} & \textbf{426} & \textbf{247} & 191 & 5.39 & \textbf{694} & \textbf{441} & 339 & 311 & 5.27 \\
top-1 confidence & MDLM & 1742 & 2811 & 2420 & 1685 & 5.38 & 1632 & 662 & 845 & 723 & 5.21 \\
 & \cellcolor{gray!12}\method{} & \textbf{993} & \textbf{1131} & \textbf{803} & \textbf{529} & 5.36 & \textbf{818} & \textbf{478} & \textbf{503} & \textbf{430} & 5.23 \\
top-2 margin & MDLM & 7051 & 5378 & 3189 & 1724 & 5.36 & 1768 & 652 & 784 & 652 & 5.22 \\
 & \cellcolor{gray!12}\method{} & \textbf{2212} & \textbf{1543} & \textbf{911} & \textbf{521} & 5.34 & \textbf{868} & \textbf{472} & \textbf{484} & \textbf{412} & 5.23 \\
lowest entropy & MDLM & 1723 & 3070 & 3127 & 2114 & 5.32 & 1618 & 663 & 891 & 794 & 5.21 \\
 & \cellcolor{gray!12}\method{} & \textbf{968} & \textbf{1266} & \textbf{1037} & \textbf{642} & 5.30 & \textbf{824} & \textbf{474} & \textbf{513} & \textbf{442} & 5.22 \\
drawn-token confidence & MDLM & 695 & 96 & \textbf{19} & \textbf{9} & 3.76 & 568 & 189 & 95 & 71 & 2.77 \\
 & \cellcolor{gray!12}\method{} & \textbf{426} & \textbf{84} & 31 & 26 & 4.69 & \textbf{356} & \textbf{120} & \textbf{52} & \textbf{34} & 4.56 \\
EB-Sampler & MDLM & 1831 & 3213 & 3232 & 2444 & 5.32 & 1688 & 795 & 1002 & 957 & 5.18 \\
 & \cellcolor{gray!12}\method{} & \textbf{994} & \textbf{1292} & \textbf{1059} & \textbf{745} & 5.30 & \textbf{824} & \textbf{531} & \textbf{552} & \textbf{490} & 5.19 \\
\bottomrule
\end{tabular}
\end{table}

\subsection{Subject-verb agreement: a quantitative analysis}
\label{app:consistency}

We present a quantitative analysis to test whether \method{} repairs the failure demonstrated in Figure~\ref{fig:concept} on real text, where two tokens revealed in the same step must be grammatically consistent. We use the six subject-verb agreement subsets of BLiMP \citep{warstadt2020blimp}, at 1,000 items each. Each item is a minimal pair of sentences that differ only in the form of the subject or of the verb. Four subsets vary a regular or an irregular plural, and in the other two, a relational noun or a relative clause stands between the subject and the verb as a distractor. We locate the word that does not vary with the part-of-speech tagger of the Natural Language Toolkit (NLTK), taking the first noun before a varying verb and the first finite verb after a varying noun. We keep only the items whose subject, verb, and alternative form are single tokens, which leaves 2,502 items on WikiText-103 and OpenWebText.

We then place each sentence at position 32 of a held-out window, reveal all other tokens of the window, and mask the subject and the verb. A kernel then assigns a joint probability to the two masked tokens in one network call, evaluated at the time value at which two positions remain masked. For each item, the agreement share $A$ is the joint probability of the grammatical pair divided by the sum of the joint probabilities of the grammatical and the ungrammatical pair, and chance corresponds to $A = 1/2$. The product kernel scores the verb without seeing the subject, and therefore stays near chance. In the chain of \method{}, the subject precedes the verb, and the transition into the verb depends on the subject. As a reference, we reveal the subject in a first call and score the verb in a second call of the pre-trained model. This reference costs two calls and shows how well the pre-trained model makes the verb consistent once it sees the subject.

As shown in Table~\ref{tab:consistency}, MDLM stays near chance in one call, while \method{} raises the mean agreement from 46.0 to 53.3$\%$ on WikiText-103, from 46.6 to 55.5$\%$ on LM1B, and from 47.7 to 60.8$\%$ on OpenWebText, about 40$\%$ of the way to the two-call reference on the first two corpora and 58$\%$ on OpenWebText. The gain concentrates on the four subsets without a distractor. On the two distractor subsets, even the two-call reference stays near chance, which points to the pre-trained denoiser as the limit there.

\begin{table}[h!]
\centering
\footnotesize
\setlength{\tabcolsep}{5pt}
\caption{\textbf{Subject-verb agreement.} Mean agreement $A$ and the fraction of items with $A > 1/2$ (\%, $\uparrow$) when both positions are revealed in one call, and for the reference that reveals the subject in a first call and scores the verb in a second.}
\label{tab:consistency}
\begin{tabular}{@{}l rr rr rr@{}}
\toprule
 & \multicolumn{2}{c}{WikiText-103} & \multicolumn{2}{c}{LM1B} & \multicolumn{2}{c}{OpenWebText} \\
\cmidrule(lr){2-3}\cmidrule(lr){4-5}\cmidrule(lr){6-7}
& mean $A$ & $A > 1/2$ & mean $A$ & $A > 1/2$ & mean $A$ & $A > 1/2$ \\
\midrule
MDLM (one call) & 46.0 & 43.9 & 46.6 & 45.0 & 47.7 & 45.7 \\
\rowcolor{gray!12}
\method{} chain (one call) & \textbf{53.3} & \textbf{53.3} & \textbf{55.5} & \textbf{56.3} & \textbf{60.8} & \textbf{62.1} \\
\midrule
two calls (reference) & 65.0 & 66.2 & 70.2 & 73.0 & 70.1 & 73.5 \\
\bottomrule
\end{tabular}
\end{table}

% =============================================================================
\section{Limitations}
\label{sec:limits}
% =============================================================================

A first-order Markov chain removes the information shared by consecutive revealed positions, and it links distant positions only through the positions between them. Future work could explore modeling the residual $\rho(q)$ of Eq.~\ref{eq:steperror}, which is the information that a token shares with its earlier non-adjacent predecessors given its immediate predecessor (App.~\ref{proof:chowliu}), and it includes higher-order constraints such as parity.

% =============================================================================
\raggedbottom
\section{Example generations}
\label{app:samples}
% =============================================================================

We generate pairs of samples with MDLM and \method{} under the same reveal schedule. We fix the positions to be unmasked in advance and then draw the tokens of those positions. 
These samples are shown as a card, gray for MDLM and teal for \method{}, including the generative perplexity and entropy of the token histogram of the sample. A token is shown in \exbad{red} when GPT-2 Large assigns it a lower probability than it assigns to nine in ten tokens of held-out text from the same corpus, while the count of these tokens is shown in the title. Since the OpenWebText samples have a context length of 1,024 tokens, we only show the first 128 tokens and the generative perplexity of the whole sample.

\begin{figure}[h!]
\centering
\begin{tcbraster}[raster columns=3, raster equal height=rows, raster column skip=3pt, raster row skip=3pt]
\begin{mdlmcard}{MDLM, 1 NFE}{PPL 2186, H 4.23}
iny \exbad{ep} \exbad{,} \exbad{resume} \exbad{child} the have complete , the . to s .y \exbad{\P{}} from \P{} . \exbad{saying} such \exbad{victim}\exbad{onal} clear kernx to i \exbad{bangladesh} \$ \exbad{district} \exbad{unpleasant} \exbad{mu} \exbad{dead} \exbad{particularly} as \exbad{upset} , from \exbad{normal} \exbad{assembled} with \exbad{operations} \exbad{study} . \exmore{}
\end{mdlmcard}
\begin{mdlmcard}{MDLM, 2 NFEs}{PPL 1254, H 4.28}
iny \exbad{ep}mics ? of the \exbad{have}\exbad{nl} . \exbad{politics} . to the . ! \exbad{\P{}} . \P{} . \exbad{saying} ? \exbad{victim}\exbad{onal} language kernx to i \exbad{bangladesh} \$ 1 ) but \exbad{dead} \exbad{particularly} as evidence , from \exbad{institutions} \exbad{assembled} with a study squad \exmore{}
\end{mdlmcard}
\begin{mdlmcard}{MDLM, 4 NFEs}{PPL 1080, H 4.49}
py ep ,\exbad{mal}m \exbad{technology} have \exbad{shocked} thailand to end years of war . \exbad{\P{}} . \P{} . \exbad{saying} the victim of . \exbad{n}x ' i \exbad{bangladesh} \$ 10rta \exbad{hence} dead \exbad{particularly} as \exbad{sticks} are \exbad{au} \exbad{remains} \exbad{assembled} as an underground group \exmore{}
\end{mdlmcard}
\begin{srhcard}{\method{}, 1 NFE}{PPL 1153, H 4.52}
in the account . \exbad{\P{}} \exbad{give} information \exbad{world} series of his \exbad{currently} \exbad{leaving} a library professor \exbad{jane} nina said she delivered more \exbad{aggressive} opening a \exbad{cutting} costs and love against the arches ramon perez \exbad{drive} , '' he accepted straight year , person . many other \exmore{}
\end{srhcard}
\begin{srhcard}{\method{}, 2 NFEs}{PPL 628, H 4.42}
in the account group had \exbad{97}\exbad{0}m . \exbad{operators} that give information is how the world \exbad{cup} and \exbad{energy}\exbad{ction} - currently leaving prices \exbad{severely}\exbad{ailed} by \exbad{freeing} \exbad{15} , 000 \exbad{combatants} , and \exbad{sen} \exbad{hillary} led bill clinton \exbad{steward} said cutting more in afghanistan \exmore{}
\end{srhcard}
\begin{srhcard}{\method{}, 4 NFEs}{PPL 341, H 4.33}
\P{} in the end , \exbad{obama} , who was behind \exbad{creative} questioning , was a \exbad{brief}\exbad{ed} , \exbad{televised} hearing \exbad{formally} calling world trade the easiest and \exbad{seeking} out his \exbad{currently} struggling country from the troubles . \exbad{man}cini , the professor of economics \exbad{person} \exbad{thoughts} \exmore{}
\end{srhcard}
\end{tcbraster}

\caption{\textbf{Examples generated by MDLM (gray, top) and MDLM + \method{} (teal, bottom) on LM1B at 1, 2, and 4 steps.} Red tokens represent the top tenth of GPT-2 Large surprise.}
\label{fig:samples}
\end{figure}

\subsection{Trajectory of one sample}

We also show the trajectory of a sample through its 4 denoising steps as follows, with MDLM on the left and \method{} on the right. The rows show the sample after each step, with the positions not yet revealed drawn as \exmask{} squares, and red tokens represent the surprising tokens.

\begin{tcbraster}[raster columns=2, raster equal height=rows, raster column skip=4pt, raster row skip=4pt]
\begin{mdlmcard}{MDLM, after step 1 of 4}{32 of 128 positions revealed}
\exmask{}y ep \exmask{}\exmask{}\exmask{} \exmask{} have \exmask{} \exmask{} \exmask{} \exmask{} \exmask{} \exmask{} \exmask{} \exmask{} \exbad{\P{}} \exmask{} \P{} . \exbad{saying} \exmask{} victim \exmask{} \exmask{} \exmask{}x \exmask{} i \exbad{bangladesh} \$ \exmask{}\exmask{} \exmask{} dead \exbad{particularly} as \exmask{} \exmask{} \exmask{} \exmask{} \exbad{assembled} \exmask{} \exmask{} \exmask{} \exmask{} of \exmask{} \exmask{} carrying all \exmask{} \exmask{} \exmask{} of \exmask{} \exmask{}\exmask{} \exmask{} \exmask{} \exmask{} \exmask{} \exmask{} \exmask{} \exmask{}\exmask{} \exbad{professor} \exmask{} \exmask{} \exmask{} \exmask{} \exmask{}\exmask{}\exmask{} \exmask{} . \exmask{} \exmask{} \exmask{} \exmask{} \exmask{} \exmask{} \exmask{} \exmask{} \exmask{} . . \exmask{} \exmask{} \exmask{} \exmask{} \exmask{} \exmask{} \exmask{} party \exmask{} \exmask{} \exmask{}\exmask{}\exmask{}\exmask{} \exmask{}\exbad{we} \exbad{ape}\exbad{gle}\exmask{} \exmask{} \exmask{}\exbad{en}\exmask{} \exmask{} between \exmask{} \exmask{} \exmask{} uk \exmask{} \exmask{} \exmask{} \exmask{} \exmask{} \exmask{} \exmask{} \exmask{} through \exmask{} \exmask{} \exmask{}
\end{mdlmcard}
\begin{srhcard}{\method{}, after step 1 of 4}{32 of 128 positions revealed}
\exmask{} in the \exmask{} \exmask{} \exmask{} \exmask{} who \exmask{} \exmask{} \exmask{} \exmask{} \exmask{} \exmask{} \exmask{} \exmask{}\exbad{ed} \exmask{} \exbad{televised} hearing \exbad{formally} \exmask{} world \exmask{} \exmask{} \exmask{} and \exmask{} out his \exbad{currently} \exmask{} \exmask{} \exmask{} the troubles . \exmask{}\exmask{} \exmask{} \exmask{} professor \exmask{} \exmask{} \exmask{} \exmask{} on \exmask{} \exmask{} been said \exmask{} \exmask{} \exmask{} - \exmask{}\exmask{} \exmask{} \exmask{}\exmask{} \exmask{} \exmask{} \exmask{} \exmask{} \exmask{} \exmask{} \exbad{sen} \exmask{} \exmask{} \exmask{} \exmask{} \exmask{} \exmask{} \exmask{} \exmask{} , \exmask{} \exmask{} \exmask{} \exmask{} \exmask{} \exmask{} \exmask{} \exmask{} \exmask{} to have \exmask{} \exmask{} \exmask{} \exmask{} \exmask{} \exmask{} \exmask{} said \exmask{} \exmask{} \exmask{} \exmask{} \exmask{} \exmask{} \exmask{} cutting costs and \exmask{} \exmask{} \exmask{} capacity \exmask{} \exmask{} against \exmask{} \exmask{} \exmask{} \exbad{\P{}} \exmask{} \exmask{} \exmask{} \exmask{} \exmask{} \exmask{} \exmask{} \exmask{} from \exmask{} \exmask{} \exmask{}
\end{srhcard}
\begin{mdlmcard}{MDLM, after step 2 of 4}{64 of 128 positions revealed}
py ep \exmask{}\exmask{}\exmask{} \exbad{technology} have \exmask{} \exmask{} \exmask{} end years \exmask{} \exmask{} \exmask{} \exbad{\P{}} \exmask{} \P{} . \exbad{saying} \exmask{} victim of \exmask{} \exbad{n}x ' i \exbad{bangladesh} \$ \exmask{}\exmask{} \exmask{} dead \exbad{particularly} as \exmask{} are \exbad{au} \exmask{} \exbad{assembled} as \exmask{} underground \exmask{} of \exmask{} are carrying all \exmask{} square \exmask{} of \exmask{} \exmask{}\exmask{} county \exmask{} \exmask{} \exmask{} \exmask{} \exmask{} nm\exmask{} \exbad{professor} \exmask{} \exmask{} \exmask{} happened \exmask{}\exbad{shi}\exmask{} \exmask{} . \exmask{} how , \exmask{} \exmask{} type \exmask{} \exmask{} youth . . \exmask{} \exmask{} \exmask{} for \exmask{} the tea party \exmask{} \exbad{sinn} great\exbad{rip}\exmask{}\exmask{} \exmask{}\exbad{we} \exbad{ape}\exbad{gle}\exmask{} \exmask{} \exmask{}\exbad{en}\exmask{} \exmask{} between \exbad{rules} \exmask{} \exmask{} uk \exmask{} \exmask{} \exmask{} to \exmask{} \exmask{} sri \exmask{} through \exmask{} heads \exbad{\P{}}
\end{mdlmcard}
\begin{srhcard}{\method{}, after step 2 of 4}{64 of 128 positions revealed}
\P{} in the \exmask{} \exmask{} \exmask{} , who \exmask{} \exmask{} \exmask{} questioning , \exmask{} \exmask{} \exmask{}\exbad{ed} \exmask{} \exbad{televised} hearing \exbad{formally} \exmask{} world trade \exmask{} easiest and \exbad{seeking} out his \exbad{currently} \exmask{} \exmask{} \exmask{} the troubles . \exmask{}cini , \exmask{} professor of \exmask{} \exbad{person} \exmask{} on \exmask{} have been said \exmask{} many \exmask{} - \exmask{}\exmask{} \exmask{} hind\exmask{} \exmask{} \exmask{} \exmask{} \exmask{} are \exmask{} \exbad{sen} \exmask{} \exmask{} \exmask{} the \exmask{} - \exmask{} \exmask{} , \exmask{} he is \exmask{} \exmask{} that \exmask{} \exmask{} appear to have \exmask{} \exmask{} \exmask{} . \exmask{} . cheney said \exmask{} - working with \exmask{} \exmask{} \exmask{} cutting costs and \exmask{} \exmask{} \exmask{} capacity \exmask{} \exmask{} against \exbad{mounting} \exmask{} \exmask{} \exbad{\P{}} \exmask{} \exmask{} \exmask{} flood \exmask{} \exmask{} country \exmask{} from \exmask{} , marc
\end{srhcard}
\begin{mdlmcard}{MDLM, after step 3 of 4}{96 of 128 positions revealed}
py ep \exmask{}\exmask{}\exmask{} \exbad{technology} have \exmask{} thailand \exmask{} end years of \exmask{} . \exbad{\P{}} . \P{} . \exbad{saying} the victim of . \exbad{n}x ' i \exbad{bangladesh} \$ 10\exmask{} \exmask{} dead \exbad{particularly} as \exbad{sticks} are \exbad{au} \exmask{} \exbad{assembled} as an underground \exmask{} of \exmask{} are carrying all 75 square \exmask{} of a \exbad{across}\exmask{} county of \exbad{memorial} \exmask{} \exmask{} \exmask{} nmc \exbad{professor} who \exmask{} \exmask{} happened 65\exbad{shi}\exmask{} \exmask{} . . how , \exbad{consider} new type of chinese youth . . is \exmask{} \exmask{} for \exmask{} the tea party \exmask{} \exbad{sinn} great\exbad{rip}s\exbad{put} \exmask{}\exbad{we} \exbad{ape}\exbad{gle}e \exmask{} \exmask{}\exbad{en}uous \exmask{} between \exbad{rules} \exmask{} \exmask{} uk society are \exmask{} to cope \exmask{} sri lanka through parallel heads \exbad{\P{}}
\end{mdlmcard}
\begin{srhcard}{\method{}, after step 3 of 4}{96 of 128 positions revealed}
\P{} in the \exmask{} \exmask{} \exmask{} , who \exmask{} behind \exmask{} questioning , was \exmask{} \exbad{brief}\exbad{ed} , \exbad{televised} hearing \exbad{formally} calling world trade the easiest and \exbad{seeking} out his \exbad{currently} struggling \exmask{} \exmask{} the troubles . \exbad{man}cini , \exmask{} professor of economics \exbad{person} \exmask{} on \exmask{} have been said that many \exmask{} - -\exbad{ane} \exmask{} hindsight by \exmask{} \exmask{} \exmask{} are being \exbad{sen} . \exmask{} \exmask{} the post - \exmask{} \exmask{} , and he is \exbad{letter} adding that other proposals appear to have turned \exmask{} \exmask{} . \exmask{} . cheney said \exmask{} - working with them to \exmask{} cutting costs and increasing \exmask{} \exmask{} capacity for \exmask{} against \exbad{mounting} \exmask{} \exmask{} \exbad{\P{}} \exbad{halt} a \exmask{} flood in \exmask{} country away from home , marc
\end{srhcard}
\begin{mdlmcard}{MDLM, after step 4 of 4}{gen-PPL 1080, entropy 4.49, 32 in red}
py ep ,\exbad{mal}m \exbad{technology} have \exbad{shocked} thailand to end years of war . \exbad{\P{}} . \P{} . \exbad{saying} the victim of . \exbad{n}x ' i \exbad{bangladesh} \$ 10rta \exbad{hence} dead \exbad{particularly} as \exbad{sticks} are \exbad{au} \exbad{remains} \exbad{assembled} as an underground group of people are carrying all 75 square miles of a \exbad{across}\exbad{la} county of \exbad{memorial} cemetery a this nmc \exbad{professor} who said s happened 65\exbad{shi}c . . . how , \exbad{consider} new type of chinese youth . . is just \exbad{recipe} for \exbad{filling} the tea party that \exbad{sinn} great\exbad{rip}s\exbad{put} pan\exbad{we} \exbad{ape}\exbad{gle}e down \exbad{ing}\exbad{en}uous \exbad{differences} between \exbad{rules} that and uk society are used to cope with sri lanka through parallel heads \exbad{\P{}}
\end{mdlmcard}
\begin{srhcard}{\method{}, after step 4 of 4}{gen-PPL 341, entropy 4.33, 22 in red}
\P{} in the end , \exbad{obama} , who was behind \exbad{creative} questioning , was a \exbad{brief}\exbad{ed} , \exbad{televised} hearing \exbad{formally} calling world trade the easiest and \exbad{seeking} out his \exbad{currently} struggling country from the troubles . \exbad{man}cini , the professor of economics \exbad{person} \exbad{thoughts} on what have been said that many ideas - -\exbad{ane} with hindsight by \exbad{philosophy} - - are being \exbad{sen} . obama during the post - war era , and he is \exbad{letter} adding that other proposals appear to have turned to \exbad{ms} . mr . cheney said - - working with them to include cutting costs and increasing the \exbad{seating} capacity for \exbad{responding} against \exbad{mounting} economic \exbad{hunger} \exbad{\P{}} \exbad{halt} a housing flood in a country away from home , marc
\end{srhcard}
\end{tcbraster}

\subsection{Paired samples at 1 and 4 steps}

\paragraph{WikiText-103, 1 NFE.}

\begin{tcbraster}[raster columns=2, raster equal height=rows, raster column skip=4pt]
\begin{mdlmcard}{MDLM, pair 1}{gen-PPL 9115, entropy 4.54, 67 in red}
by 23;.\exbad{et} \exbad{It} \exbad{\P{}} \exbad{budget} \exbad{Line} was, a \exbad{through}\exbad{This} \exbad{arrives} \exbad{compulsory} \exbad{Beijing} \exbad{ought}'s- a, \exbad{Bowie} \exbad{25}'s \exbad{become} \exbad{Allied} into (\exbad{ction} \exbad{Pokémon} it was \exbad{==}\exbad{ue} \exbad{member}, not \exbad{nose} \exbad{song} all 8's \exbad{groups}, the \exbad{locks} \exbad{reason} \exbad{satisfied}), L \exbad{major}. \exbad{Simpsons} \exbad{urban} be \exbad{which} other \exbad{furious} the with \exbad{against}, \exbad{mud} on G\exbad{ested} the when two he \exbad{pop} the in \P{} \exbad{the}\exbad{)} that \exbad{hits} \exbad{2015}\exbad{isk} \exbad{guns}.''\exbad{ia} \exbad{Cart}\exbad{anic} \exbad{causing} \exbad{rest} \exbad{John} \exbad{scenes} of \exbad{also}\exbad{ous} \exbad{Congress} \exbad{Tal} \exbad{public} \exbad{responsible} were \exbad{constraints}\exbad{bach} a school \exbad{occ} than they\exbad{wood} from non\exbad{560} \exbad{Express} \exbad{inside}\exbad{Pl}: \exbad{tattoo} was \exbad{plan} a \exbad{bur} he I new his\exbad{acting}, \exbad{included} \P{}
\end{mdlmcard}
\begin{srhcard}{\method{}, pair 1}{gen-PPL 3324, entropy 4.43, 59 in red}
by the second, \exbad{Pier} \exbad{\P{}} \exbad{Wayne} \exbad{Sa}ik \exbad{the} Queen was also \exbad{resulted} in his own \exbad{depth} \exbad{their} presence of the first world \exbad{really} \exbad{technical} information on \exbad{After} \exbad{starting} \exbad{innings} \exbad{1983}, \exbad{parallel} \exbad{called} ''\exbad{Stars} All the \exbad{California} area by his command \exbad{self}-\exbad{23} - 14, \exbad{allowed} being the \exbad{90} \P{} \exbad{\P{}} \exbad{=}\exbad{202} months \exbad{resulting} \exbad{line} at number of a \exbad{lion}'s \exbad{adj}\exbad{ibia}, a \exbad{hub} \exbad{deposits} of \exbad{Core}\exbad{ev} \exbad{increasingly} \exbad{hostile} \exbad{monopoly}\exbad{able} or \exbad{Sk}and\exbad{ics} \exbad{===} \exbad{\P{}} \exbad{debut}. \exbad{\P{}} \exbad{In} his book \exbad{was} unable to \exbad{their} degree during the \exbad{turf} \exbad{grows} at \exbad{record} in the city and Ge\exbad{vol}\exbad{sis} is located in \exbad{2001} to her brother \exbad{Weeks}\exbad{'}\exbad{d} \exbad{==} \P{} \exbad{The} \exbad{RPGs}
\end{srhcard}
\end{tcbraster}

\paragraph{WikiText-103, 4 NFEs.}

\begin{tcbraster}[raster columns=2, raster equal height=rows, raster column skip=4pt]
\begin{mdlmcard}{MDLM, pair 1}{gen-PPL 2322, entropy 4.54, 53 in red}
\P{} \exbad{23}; \exbad{tied} \exbad{Zero} \exbad{missing} Zero \exbad{budget}ing was \exbad{represented} by \exbad{notorious}\exbad{Not}ch \exbad{Professor} \exbad{Beijing} 7's- a\exbad{foot} \exbad{Bowie} (as an \exbad{Allied} \exbad{viruses} (\exbad{ction} \exbad{Pokémon}, \exbad{Bill} Zue \exbad{member}, in) representative ), 8\exbad{45} \exbad{dark} \exbad{privileges} \& \exbad{locks} (\exbad{FBI}), L6, the \exbad{urban} sewer, \exbad{approximately} \exbad{k}\exbad{azel}\exbad{aj}r, an angel infested, Dr \exbad{investigator} Jeff Goldblum, the \exbad{petrol} \exbad{payment} \exbad{13}59\exbad{650} to the \exbad{cruiser} \exbad{Acad}ia \exbad{Cart}oons for 18 \exbad{Nixon}'s and\exbad{danger}ous medical \exbad{exclusive} \exbad{Ichigo} with the \exbad{tap} into a school \exbad{occ}\exbad{cal}ation \exbad{pipe} from the Orient Express, and a tattooed \exbad{business}\exbad{man} and hacker \exbad{TOP}\exbad{402} \exbad{En}\exbad{acting} \exbad{Sydney} \exbad{Kov} \exbad{\P{}}
\end{mdlmcard}
\begin{srhcard}{\method{}, pair 1}{gen-PPL 436, entropy 4.45, 23 in red}
\P{} \exbad{by} the New York \exbad{Aviation}, which they had earlier called to a \exbad{Marina} 3, Tower \exbad{Pier} \P{} \exbad{Wayne} \exbad{Ye}\exbad{oke} \exbad{presided} as a passenger when the Queen was \exbad{departing} \exbad{without} \exbad{boat}. After the match, \exbad{Ben} \exbad{Ver}de's and his partner \exbad{Franco} Nespor \exbad{their} final decision \exbad{selecting} an ideal player for the New \exbad{Indies}. PoTTDscored his highest Test century, with 19.67, and \exbad{remarked} that ''I are being red-staged together the next morning. Our tour being really easy because the world has put our \exbad{technical} information on the players.'' After an ODI A Test \exbad{cancelled} \exbad{innings} started, the resulting international defeats coming during \exbad{1983}. P \exbad{\P{}}
\end{srhcard}
\end{tcbraster}

\paragraph{OpenWebText, 1 NFE.}

\begin{tcbraster}[raster columns=2, raster equal height=rows, raster column skip=4pt]
\begin{mdlmcard}{MDLM, pair 2}{gen-PPL 4056, first 128 of 1024 tokens}
the\exbad{ue} \exbad{one}, \exbad{may} the \exbad{reported}, \exbad{don} \exbad{for} \exbad{based} \exbad{Labor} of \exbad{job}\exbad{IS} \exbad{brought}\exbad{On}\exbad{You}\exbad{'} \exbad{like}\exbad{'t} \exbad{if} \exbad{Thrones} \exbad{V} \exbad{as} \exbad{because}\exbad{-} \exbad{This} \exbad{Bowl} \exbad{an}'\exbad{After}\exbad{167} \exbad{rink} \exbad{ultimately} \exbad{main}, of \exbad{at} \exbad{Ltd} \exbad{market} \exbad{felt}, \exbad{sensit} \exbad{its}\exbad{cow} \exbad{Computer} and\exbad{s} \exbad{Football} \exbad{Russian} \exbad{could} \exbad{direct} \exbad{Subscribe}\exbad{acc} \exbad{live} at and\exbad{'t},\exbad{uru} \exbad{explain}S. \exbad{seem} the \exbad{labour} the \exbad{eurozone}\exbad{onde} \exbad{word} \exbad{normally} \exbad{Lucifer} \exbad{South} \exbad{basically} \exbad{my} \exbad{cervical} in \exbad{war}'an, \exbad{cheerful} \exbad{14} \exbad{pret}\exbad{OU}\exbad{ot} not \exbad{then} \exbad{has} \exbad{declined}\exbad{t} \exbad{overall} \exbad{shown} \exbad{like} that \exbad{prison} a \exbad{couple}' the, \exbad{good} \exbad{at} so she \exbad{argument} \exbad{front} \exbad{there} she\exbad{uts} is the. the for \exbad{clashes} \exbad{crap}\exbad{mm}, the\exbad{bell} \exbad{Volume} \exbad{very} ' \exbad{Ferguson} \exbad{social} \exbad{Netherlands} \exmore{}
\end{mdlmcard}
\begin{srhcard}{\method{}, pair 2}{gen-PPL 1119, first 128 of 1024 tokens}
a \exbad{people} and a \exbad{p}\exbad{ons}s \exbad{a} \exbad{compliment}ed e\exbad{'}\exbad{I} \exbad{hadn} \exbad{had} been\exbad{min}\exbad{id}\exbad{\textbar{}}\exbad{AL}\exbad{ITY},\exbad{'}s \exbad{known} as \exbad{simply} \exbad{characterized} by the \exbad{character} in \exbad{2010}.\exbad{Using} their \exbad{left} \exbad{it} not \exbad{evidence} that they're from a \exbad{comprehensive} \exbad{story}\exbad{include} \exbad{http}://www. \exbad{A}\exbad{TS} \exbad{in} the \exbad{Red}, the \exbad{international} \exbad{government} \exbad{was} that \exbad{single} \exbad{table} and \exbad{foreign} \exbad{province} \exbad{towards} \exbad{minorities}, \exbad{gloss}\exbad{ate}, where you choose you \exbad{couldn}'t \exbad{had} \exbad{decided} to the \exbad{May} 21, \exbad{on} \exbad{serious} time and \exbad{AFP}/\exbad{list}\exbad{ended} \exbad{veteran} \exbad{look} about why you \exbad{love} to make these \exbad{large} as a \exbad{no} not having \exbad{compared} \exbad{designer} \exbad{B}\exbad{IG}\exbad{'}s, there are \exbad{aligned} with \exbad{subtly} of a \exbad{year}.\exbad{And} according \exmore{}
\end{srhcard}
\end{tcbraster}

\paragraph{OpenWebText, 4 NFEs.}

\begin{tcbraster}[raster columns=2, raster equal height=rows, raster column skip=4pt]
\begin{mdlmcard}{MDLM, pair 3}{gen-PPL 1810, first 128 of 1024 tokens}
\P{} \exbad{a} \exbad{their} \exbad{home}defines \exbad{book} a, \exbad{bond} to \exbad{for} you \exbad{good} \exbad{women} \exbad{c}\exbad{uties} \exbad{n}\exbad{other} \exbad{baby}. \exbad{eating}\exbad{mm} \exbad{imaging} or \exbad{son}'s \exbad{lunch}\exbad{''}a \exbad{human}\exbad{to} \exbad{forced} some \exbad{sn} \exbad{is} \exbad{known} as:\exbad{dem}\exbad{erg}\exbad{boot} or:\exbad{etic}\exbad{olitics} of \exbad{activists}. The \exbad{cycle} ends because you \exbad{permanently} \exbad{dim} \exbad{room} to meet the demands\exbad{-} the \exbad{kin} are \exbad{conform} to\exbad{gn}\exbad{root} \exbad{pass}\exbad{shirts}.\exbad{'}The \exbad{friends} said he \exbad{want} him \exbad{through} \exbad{record} in \exbad{getting} over.The \exbad{the} \exbad{stand}-\exbad{1} a high of \exbad{39} and\exbad{'}\exbad{William} did \exbad{music}\exbad{!} \exbad{pressure}. It\exbad{'}s \exbad{41}$\%$ but the \exbad{quickest} \exbad{comes} \exbad{despite} \exbad{Of} late to the \exbad{Son} \exbad{JavaScript} \exbad{hit} \exbad{''}\exbad{Sk}irt\exbad{aking} \exbad{expert} \exbad{J}\exbad{un}\exbad{co}\exbad{ows} \exmore{}
\end{mdlmcard}
\begin{srhcard}{\method{}, pair 3}{gen-PPL 541, first 128 of 1024 tokens}
\P{} \exbad{this} \exbad{issue}How \exbad{our} \exbad{article} was \exbad{missing} that \exbad{every} \exbad{prom}\exbad{í}\exbad{\textbackslash{}}\exbad{ring} \exbad{reference}, I would \exbad{treat} a \exbad{lack} of \exbad{trem}ors and \exbad{tone} lossIn terms \exbad{that} makes sense\exbad{Find} your \exbad{European} \exbad{Ros}er to \exbad{you} and \exbad{recruit} that is a \exbad{niche} product by \exbad{i}H\exbad{appiness}\exbad{?,} \exbad{prescription} \exbad{Drug}\exbad{Le}aving the \exbad{high} price of \exbad{FC} \exbad{4} \exbad{PM} in his \exbad{blood} \exbad{your} sidekick \exbad{tackles} \exbad{have} been all \exbad{through} the \exbad{regular} season. \exbad{-} \exbad{his} \exbad{Xavier}\exbad{B}\exbad{AN}\exbad{IFA}\exbad{PA} - \exbad{K}OM\exbad{'}s \exbad{story} \exbad{ranked} \exbad{Views}\exbad{[}\exbad{From}: \exbad{Mes}\exbad{ric} \exbad{Bal}\exbad{fort}]\exbad{Rest}ore \exbad{battery} \exbad{packs} - \exbad{Second} game \exbad{Baltimore} \exbad{Is}k\exbad{orea} \exbad{Twitter} User\exbad{\textgreater{}}The \exbad{teams} \exbad{celebrities} and fans at \exbad{get} \exbad{special} \exbad{soccer} field \exbad{service} in one \exbad{mid} \exmore{}
\end{srhcard}
\end{tcbraster}

\subsection{QM9 and ZINC-250k samples}

\begin{figure}[h!]
\centering
\includegraphics[width=\textwidth]{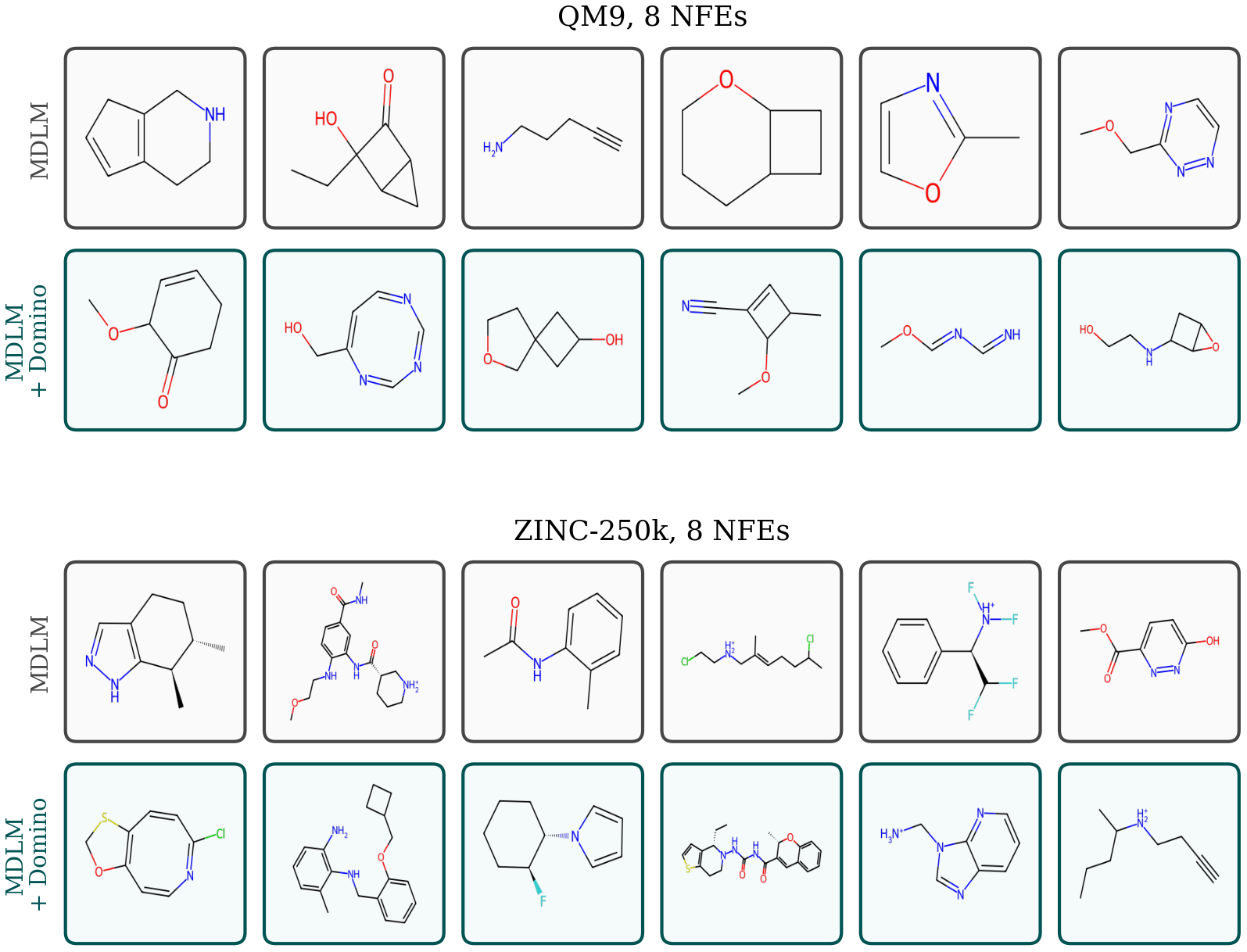}
\caption{\textbf{Valid molecules at 8 NFEs.} The first six valid samples above a size cutoff from MDLM (gray) and MDLM + \method{} (teal) on QM9 (top) and ZINC-250k (bottom).}
\label{fig:molsamples}
\end{figure}

Figure~\ref{fig:molsamples} shows example molecules generated by the pre-trained MDLM and \method{} on QM9 and ZINC-250k at 8 NFEs.

\subsection{ImageNet-256 samples}

\begin{figure}[h!]
\centering
\includegraphics[width=\textwidth]{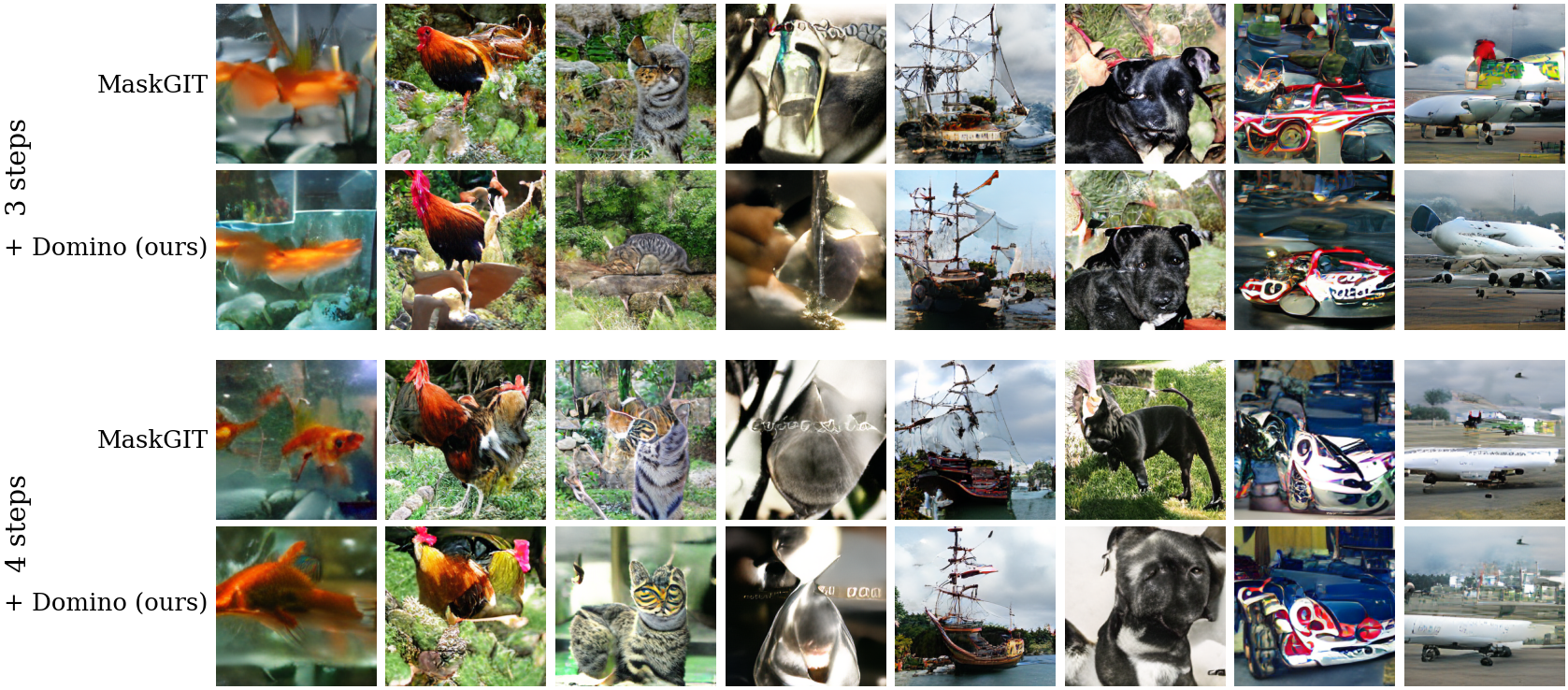}
\caption{Example images generated by MaskGIT and MaskGIT + \method{} on ImageNet-256 at 3 and 4 steps with the same class shown in columns. }
\label{fig:imsamples}
\end{figure}

Figure~\ref{fig:imsamples} shows uncurated samples of MaskGIT and \method{} on ImageNet-256 at 3 and 4 steps, under MaskGIT's own sampler at a guidance weight of 3 and temperature 1. 

\end{document}